\documentclass[%
	paper=A4,					
	twoside=false,				
	openright,					
	parskip=half,				
	chapterprefix=true,			
	11pt,						
	headings=normal,			
	bibliography=totoc,			
	listof=totoc,				
	titlepage=on,				
	captions=tableabove,		
	draft=false,				
]{scrreprt}%

\usepackage{CJKutf8}
\usepackage[					
	figuresep=colon,%
	sansserif=false,%
	hangfigurecaption=false,%
	hangsection=true,%
	hangsubsection=true,%
	colorize=bw,%
	colortheme=bluemagenta,%
	wrapfooter=true,%
  bibsys=bibtex,%
  bibfile=main,%
	bibstyle=authoryear,%
]{cleanthesis}

\usepackage{titlesec}
\makeatletter
\g@addto@macro{\UrlBreaks}{\UrlOrds}
\makeatother

\usepackage{lgrind}
\usepackage{graphicx}
\usepackage{setspace}

\usepackage{glossaries}
\makeglossaries

\usepackage{cmap}
\usepackage[T1]{fontenc}
\usepackage{comment}
\usepackage{amsmath,amsthm,amssymb,scrextend}
\usepackage{bm}
\usepackage{bbm}
\usepackage[ruled]{algorithm2e}
\usepackage{svg}
\usepackage{booktabs}
\usepackage{array}
\usepackage{breakcites}
\usepackage{tikz}

 \usepackage{subcaption}
 \usepackage{multirow}
 \usepackage{xhfill}
 \usepackage{cleveref}
 \usepackage{float}
\usepackage{titlesec}
\usepackage{mathtools}
\usepackage{epigraph}
  \usepackage{etoolbox}

\makeatletter
\newcommand{\removelatexerror}{\let\@latex@error\@gobble}
\makeatother

\newenvironment{derivation}{\begin{center}\begin{math}\begin{array}{>{\displaystyle}rc>{\displaystyle}l}}{\end{array}\end{math}\end{center}}
 \newcommand{\bx}{\mathbf{x}}
 \newcommand{\by}{\mathbf{y}}
   \newcommand{\yhat}{\hat{y}}
  
 \newcommand{\bh}{\mathbf{h}}
  
 \newcommand{\bo}{\mathbf{o}}
  \newcommand{\botilde}{\mathbf{\tilde{o}}}
  \newcommand{\bohat}{\mathbf{\hat{o}}}
	\newcommand{\bkappa}{\mathbf{\varkappa}}
 \newcommand{\bW}{\mathbf{W}}
  \newcommand{\bw}{\mathbf{w}}
 \newcommand{\bb}{\mathbf{b}}
 \newcommand{\bZ}{\mathbf{Z}}
  
 \newcommand{\btheta}{{\bm{\theta}}}
 \newcommand{\bdelta}{{\bm{\delta}}}
 \newcommand{\bomega}{{\bm{\omega}}}
 \newcommand{\bphi}{{\bm{\phi}}}
  \newcommand{\bmu}{{\bm{\mu}}}
   \newcommand{\bSigma}{{\bm{\Sigma}}}
   \newcommand{\bGamma}{{\bm{\Gamma}}}

\newcommand{\wnl}{\\[0.35cm]}
\newcommand{\indicator}[1]{\mathbbm{1}(#1)}
\renewcommand{\qed}{\hfill$\blacksquare$}
\newcommand{\pder}[2]{\displaystyle \frac{\partial #1}{\partial #2}}

\renewenvironment{proof}{\begin{addmargin}[1em]{0em}\begin{newproof}}{\end{newproof}\end{addmargin}\qed}
\newcommand{\proofend}{\begin{flushright}$\Box$\end{flushright}}
\makeatletter
\newcommand{\vast}{\bBigg@{4}}
\newcommand{\Vast}{\bBigg@{5}}
\makeatother

\newtheorem{theorem}{Theorem}
\newtheorem{definition}{Definition}
\newtheorem{lemma}{Lemma}

\newcommand{\surpcol}[1]{\textcolor{orange!70!black}{#1}}
\newcommand{\baecol}[1]{\textcolor{green!40!black}{#1}}
\newcommand{\mcdcol}[1]{\textcolor{BlueViolet}{#1}}
\newcommand{\varcol}[1]{\textcolor{blue!60!black}{#1}}
\newcommand{\updiff}[1]{\textcolor{green!30!black}{$\uparrow #1$}}
\newcommand{\samediff}[1]{\textcolor{gray!50!black}{$\rightarrow \pm #1$}}
\newcommand{\downdiff}[1]{\textcolor{red!60!black}{$\downarrow #1$}}

\DeclareMathOperator*{\argmax}{arg\,max}
\DeclareMathOperator*{\argmin}{arg\,min}

\begin{document}

\begin{titlepage}

\newcommand{\HRule}{\rule{\linewidth}{0.5mm}} 
\center 


\includegraphics[width=\linewidth]{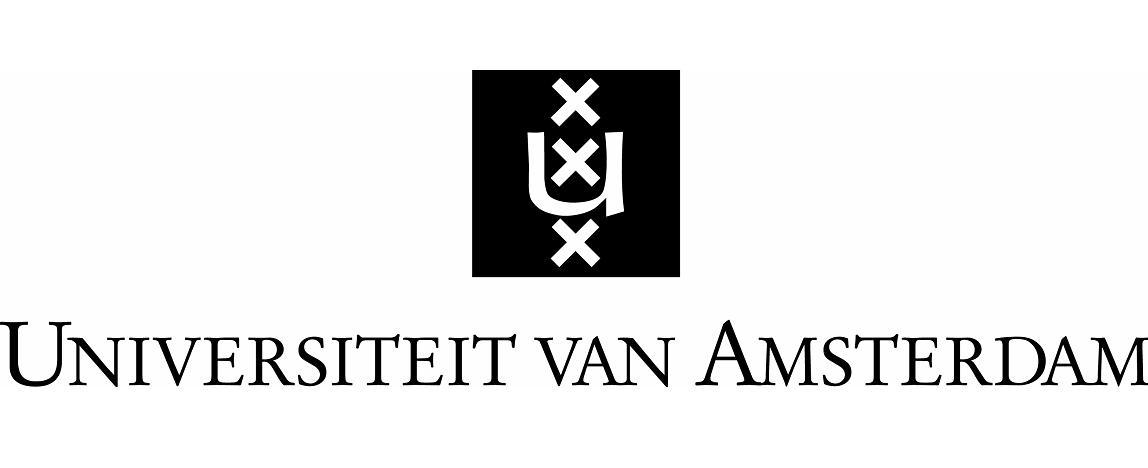}\\[0.4cm]
\textsc{\Large MSc Artificial Intelligence}\\[0.2cm]
\textsc{\Large Master Thesis}\\[0.4cm]


\HRule \\[0.25cm]
{\huge \bfseries Recoding latent sentence representations\\[0.25cm]{\large Dynamic gradient-based activation modification in RNNs}
}\\[0.25cm] 
\HRule \\[0.2cm]


by\\[0.2cm]
\textsc{\Large Dennis Ulmer}\\[0.2cm] 
11733187\\[0.3cm]


{\Large August 9, 2019}\\[0.3cm] 

36 ECTS\\ %
January 2019 - August 2019\\[0.3cm]%

\begin{minipage}[t]{0.4\textwidth}
\begin{flushleft} \large
\emph{Supervisors:} \\
\textsc{Dieuwke Hupkes}\\
Dr. \textsc{Elia Bruni}
\end{flushleft}
\end{minipage}
~
\begin{minipage}[t]{0.4\textwidth}
\begin{flushright} \large
\emph{Assessor:} \\
Dr. \textsc{Willem Zuidema}\\
\end{flushright}
\end{minipage}\\[1cm]


\includegraphics[width=2.5cm]{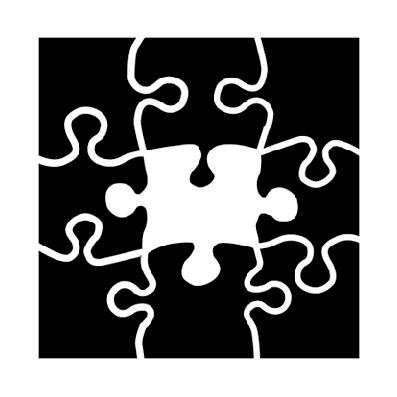}\\ 
\textsc{\large Institute for Logic, Language and Computation\\University of Amsterdam}\\[1.0cm] %


\vfill 

\end{titlepage}

\pagestyle{plain}
\newpage\null\thispagestyle{empty}\newpage
%
%
%
\chapter*{Abstract}
\thispagestyle{empty}

In Recurrent Neural Networks (RNNs), encoding information in a suboptimal or erroneous way can impact the quality of representations based on later elements in the sequence and subsequently lead to wrong predictions and a worse model performance. In humans, challenging cases like garden path sentences (an instance of this being the infamous \emph{The horse raced past the barn fell}) can lead their language understanding astray. However, they are still able to correct their representation accordingly and recover when new information is encountered. Inspired by this, I propose an augmentation to standard RNNs in form of a gradient-based correction mechanism: This way I hope to enable such models to dynamically adapt their inner representation of a sentence, adding a way to correct deviations as soon as they occur. This could therefore lead to more robust models using more flexible representations, even during inference time.\\

The mechanism explored in this work is inspired by work of \cite{giulianelli2018under}, who demonstrated that activations in a LSTM can be corrected in order to recover corrupted information (\emph{interventions}). This is achieved by the usage of “Diagnostic Classifiers” (DG) \citep{hupkes2018visualisation}, linear models that can help to identify the kind of information recurrent neural networks encode in their hidden representations and correct them using their gradients. In this thesis, I present a generalized framework based on this idea to dynamically adapt hidden activations based on local error signals (\emph{recoding}) and implement it in form of a novel mechanism. I explore signals which are either based directly on the task's objective or on the model's confidence, leveraging recent techniques from Bayesian Deep Learning \citep{gal2016dropout, gal2016theoretically, pearce2018uncertainty}.\\

Crucially, these updates take place on an \emph{activation} rather than a parameter level and are performed during training and testing alike. I conduct different experiments in the context of language modeling,
where the impact of using such a mechanism is examined in detail.\\

To this end, I look at modifications based on different kinds of time-dependent error signals and how they influence the model performance. Furthermore, this work contains a study of the model's confidence in its predictions during training and for challenging test samples and the effect of the manipulation thereof. Lastly, I also study the difference in behavior of these novel models compared to a standard LSTM baseline and investigate error cases in detail to identify points of future research. I show that while the proposed approach comes with promising theoretical guarantees and an appealing intuition, it is only able to produce minor improvements over the baseline due to challenges in its practical application and the efficacy of the tested model variants.

\newpage\null\thispagestyle{empty}\newpage
\chapter*{Acknowledgements}
\thispagestyle{empty}

\epigraph{\hfill \begin{CJK}{UTF8}{gbsn}
 \textbf{春风化雨}
\end{CJK} - Spring wind and rain}{\vspace{0.4cm}\emph{Chinese idiom about the merits of higher education}}

\begin{flushright}
  Amsterdam, August 9, 2019
\end{flushright}

Nobody in this world succeeds alone. Giraffe calves are able to walk around on
their own within an hour after their birth, while the whole process takes human babies
around a year. And even then they require intense parental care throughout
infancy, and, in modern society, often depend on others way into their
adulthood.\\
In the end, the essence of our own personality is distilled from the sum of all
of our experiences, which includes all the interactions we have with others.
This forms a giant, chaotic and beautiful network of beings, constantly in flux.
I therefore owe myself to the ones around me, a selection of which I want to
thank at this point.\\

First and foremost, I want to thank my supervisors Dieuwke and Elia. Not only
have they fulfilled their duties as such with excellence by providing guidance and
constant feedback, they were also always available when I had questions or expressed doubts.
They furthermore gave me the opportunity to participate in research projects and
encouraged me to attend conferences and publish findings, essentially supporting
my first steps into academia. I also want to thank Willem Zuidema for taking
his time in assessing my thesis and taking part in my defense.\\

Secondly, I would like express my eternal gratitude to my family for all the
ways that they have supported my throughout my entire life in a myriad of ways.
My grandma Ingrid, my father Siegfried and my sister Anna; thank you for
believing in me and supporting me and providing me with the opportunity to develop
follow my interests. I want to especially thank my mother Michaela for
constantly trying to enkindle my curiosity when growing up, always nudging me into the right direction during my teenage years and
offering guidance to this very day.\\

Next, I would like to thank all my friends who contributed to this work and the
past two years of my life in a manifold of ways: Thank you Anna, Christina, Raj and
Santhosh for all those beautiful memories that I will be fondly remembering for
the rest of my life as soon as my mind reminisces about my time here in Amsterdam.
Futhermore, thank you Caro for holding me accountable and being a pain in the
neck of my inertia, and living through the highs and lows of the last two years.
Thank you Putri for getting me addicted to coffee, something I had resisted to my
whole life up until this point and for providing company in a costantly emptying Science Park.
Also, thank you Mirthe for applying mathematical rigour to my derivations and to
Atilla for being a super-human proof-reader.\\

I could extend this list forever but I have to leave some pages
for my actual thesis, therefore let me state the following: Although you might
have stayed unmentioned throughout this text, your contribution has not stayed
unnoticed. And for the case that I have not made my appreciation to you
explicit:
Thank you. I am grateful beyond measure to be surrounded by some many great
people that have supported me in times of doubt and that I know that I can
always turn to and the many more that have enriched my life in smaller ways.\\

\clearpage

\newpage\null\thispagestyle{empty}\newpage
\newglossaryentry{RNN}{name=RNN, description={Recurrent Neural Network}}
\newglossaryentry{HMM}{name=HMM, description={Hidden Markov Model}}
\newglossaryentry{BAE}{name=BAE, description={Bayesian Anchored Ensembling}}
\newglossaryentry{LSTM}{name=LSTM, description={Long-Short Term Memory}}
\newglossaryentry{MAP}{name=MAP, description={Maximum a posteriori}}
\newglossaryentry{MAP}{name=DNN, description={Deep Neural Network}}
\newglossaryentry{ReLU}{name=ReLU, description={Rectified Linear Unit}}
\newglossaryentry{MSE}{name=MSE, description={Minimum Squared Error (loss)}}
\newglossaryentry{BPTT}{name=BPTT, description={Backpropagation Through Time}}
\newglossaryentry{ELBO}{name=ELBO, description={Evidence Lower Bound}}
\newglossaryentry{NN}{name=NN, description={Neural Network}}
\newglossaryentry{DNN}{name=DNN, description={Deep Neural Network}}
\newglossaryentry{DC}{name=DC, description={Diagnostic Classifier}}
\newglossaryentry{MLP}{name=MLP, description={Multi-Layer Perceptron}}
\newglossaryentry{PTB}{name=PTB, description={Penn Treebank}}
\newglossaryentry{MCD}{name=MCD, description={Monte Carlo Dropout}}
\newglossaryentry{MLE}{name=MLE, description={Maximum Likelihood Estimate}}

\printglossary[title={List of Abbreviations}]

\tableofcontents
\newpage
\listoftables
\listoffigures

\chapter*{Note on Notation}

In this short section I introduce some conventions about the mathematical notations in this thesis.
This effort aims to make the use of letters and symbols more consistent and thereby avoid any confusion.

\textbf{Statistics} A random variable $X: \Omega \mapsto E$ is a function that maps a set of possible outcomes $\Omega$ to measurable space $E$. In this work, the measurable space $E$ always corresponds to the
set of real numbers $\mathbb{R}$. Parallel to the conventions in the Machine Learning research community, random variables are denoted using an uppercase letter, while realizations of the random variable from $E$ are denoted with lowercase letters $X = x, x \in E$. Furthermore, the probability of this realization is denoted using a lowercase $p$ like in $p(x)$, disregarding whether the underlying distribution is a probability mass function (discrete case) or probability density function (continuous case). In cases where a variable has multiple distributions (e.g. in the context of variational inference), another lowercase letter like $q(x)$ is used. The same holds when using conditional probabilites like $p(y|x)$. In case the lowercase realization is bolded, the distribution is multivariate, e.g. $p(\bx)$. $\mathbb{E}[\cdot], \text{Var}[\cdot]$ denote the expected value and variance of a random variable, while $\mathbb{H}[\cdot]$ denotes the entropy of its corresponding distribution. If no subscript is used for the expected value, we are evaluating the expected value with respect to the same probability distribution; in other cases this is made explicit like in $\mathbb{E}_{q(x)}[p(x)] = \int q(x)p(x)dx$. \wnl

\textbf{Linear Algebra} Matrices are being denoted by using bold uppercase letters, e.g. $\mathbf{A}$ or $\bGamma$, including the identity matrix, $\mathbf{I}$. Vectors
however are written using lowercase bold letter like $\bb$. Scalars are simply written as lowercase letters like $y$. Vector norms are written using $||\ldots||$, using the subscript $(\infty)$ for the $l_\infty$ norm. Without subscripts, this notations corresponds to the euclidean norm or inner product $||\bx|| = (\bx^T\bx)^{1/2} = \langle \bx, \bx\rangle$.\wnl

\textbf{Parameters} Throughout this work, sets of (learnable) model parameters are denoted using bolded greek lowercase letters, where $\btheta$ denotes the set of all model parameters and $\bomega$ the set of all model weight matrices (without biases). In a slight abuse of notation, these two symbols are also used in contexts where they symbolize a \emph{vector} of (all) parameters. In case a function is parameterized by a set of parameters, this is indicated by adding them as a subscript, e.g. $f_\btheta(\cdot)$.

\textbf{Indices, Sets \& Sequences} When using indexed elements, the lowercase letter is used for some element of an indexed sequence while the uppercase letter denotes the highest index, like $1 \ldots, k, \ldots K$. The only exception to this is the index $i$, where $N$ is used to write the total number of elements indexed by $i$. The index $t$ is exclusively used to denote time. Sets of elements are written using a shorthand $\{x\}_1^K = \{x_1, \ldots, x_k, \ldots, x_K\}$. In case the elements are ordered, a sequence is written similarly using $\langle x \rangle_1^K = \langle x_1, \ldots, x_k, \ldots, x_K\rangle$. Both sets and sequences are written using caligraphic letters like $\mathcal{D}$ or $\mathcal{V}$. In case of sets or sequences are used for probalities or function arguments, the brackets are omitted completely for readability, as in $p(x_1^T) = p(x_1, \ldots, x_t, \ldots, x_T)$

\textbf{Superscripts} $\cdot^*$ is used to denote the best realization of some variable or parameter. $\cdot^\prime$ is used to signifies some new, modified or updated version. $\hat{\cdot}$ denotes an estimate, while $\tilde{\cdot}$ is used for approximations or preliminary values. And index in parenthesis like $\cdot^{(k)}$ indicates the membership to some group or set.

\textbf{Functions} Here I simply list a short description of the most commonly used functions in this work:
\begin{itemize}
  \item $\exp(\cdot)$: Exponential function with base $e$
  \item $\log(\cdot), \log_2(\cdot)$: Natural and base $2$ logarithm
  \item $\mathcal{N}(\cdot|\cdot, \cdot)$: Uni- or multivariate normal distribution
  \item $\text{Bernoulli}(\cdot)$: Bernoulli distribution
  \item $\indicator{\cdot}$: Indicator function
  \item $\text{KL}[\cdot||\cdot]$: Kullback-Leibler divergence
  \item $\mathcal{L}(\cdot)$: Some loss function used for neural network training
  \item $\text{ELBO}[\cdot]$: Evidence lower bound of a probability distribution
\end{itemize}

\pagestyle{maincontentstyle}
\chapter[Introduction]{Introduction}\label{chapter:intro}

\epigraph{``Language is perhaps our greatest accomplishment as a species. Once a people have established a language, they have a series of agreements on how to label, characterize, and categorize the world around them [...]. These agreements then serve as the foundation for all other agreements in society. Rosseau's social contract is not the first contractual foundation of human society [...]. Language is."}{\vspace{0.4cm}\emph{Don't sleep, there are snakes - David Everett}}

Language forms the basis of most human communication. It enables abstract reasoning, the expression of thought and the transfer of knowledge from one individual to another. It also seems to be a constant across all human languages that their expressivity is infinite, i.e.\ that there is an infinite amount of possible sentences in every human language \citep{hauser2002faculty}.\footnote{Although this also remains a controversial notion considering e.g.\ the Pirah\~a language, which seems to lack recursion \citep{everett20042005}. This is further discussed in \cite{evans2009myth}, but not the intention of this work.} Furthermore, natural language is often also ambiguous and messy \citep{sapir1921introduction} and relies on the recipient to derive the right meaning from context \citep{chomsky1956three, grice1975logic}, a fact that distinguishes it from unambiguous artificial or context-free languages like predicate logic or some programming languages \citep{ginsburg1966mathematical}.
It might therefore seem surprising that humans - given the constraints in time and memory they face when processing language - are able to do successfully. How they do so is still far from being fully understood, which is why it might be helpful to study instances which challenge human language processing.\\

As an example, take the following sentence: ``The old man the boat'' is an instance of a phenomenon called \emph{garden-path sentences}.\footnote{The name is derived from the idiom ``(to) be led down the garden path'', as in being mislead or deceived.} These sentences are syntactically challenging, because their beginning suggests a parse tree that turns out to be wrong once they finish reading the sentence. In the end, it forces them to reconstruct the sentence's parse. In this particular instance, the sentence ``The old man the boat'' seems to lack its main verb, until it is realized by the reader that ``The old man'' is \emph{not} a noun phrase consisting of \emph{determiner - adjective - noun} but actually a verb phrase based on ``(to) man'' as the main verb and ``The old'' as the sentence's subject. Humans are prone to these ambiguities \citep{christiansen2016now, tessler2019incremental}, but display two interesting abilities when handling them:

\begin{enumerate}
    \item They process inputs based on some prior beliefs about the frequency and prevalence of a sentence's syntactic patterns and semantics.
    \item They are able to adapt and correct their inner representation based on newly encountered, incrementally processed information.
\end{enumerate}

Insights like these acquired in linguistics can enable progress in the field of \emph{computational linguistics}, which tries to solve language-based problems with computers. Since the surge of electronic computing in the 1940s, applying these powerful machines to difficult problems like automatic translation has been an active area of research \citep{abney2007semisupervised}. While early works in natural language processing and other subfields of artificial intelligence have focused on developing intricate symbolic approaches (for an overview see e.g.\ \cite{buchanan2005very, jurafsky2000speech}), the need for statistical solutions was realized over time.
Early trailblazers of this trend can be found in the work of \cite{church1989stochastic} and \cite{derose1988grammatical}, who apply so-called Hidden Markov Models (\glspl{HMM}) to the problem of Part-of-Speech tagging, i.e.\ assigning categories to words in a sentence. Nowadays, the field remains deeply intertwined with a new generation of statistical models: Artificial Neural Networks, also called Deep Neural Networks (\glspl{DNN}) or Multi-Layer Perceptrons (\glspl{MLP}). These connectonist models, trying to (very loosely) imitate the organization of neurons in the human brain, are by no means a recent idea.
With the first artificial neuron being trained as early as the 1950s \citep{rosenblatt1958perceptron}, effective algorithms to train several layers of artificial cells connected in parallel were developed by \cite{rumelhart1988learning}. However, after some fundamental criticisms \citep{minsky1969perceptrons}, this line of research fell mostly dormant. Although another wave of related research gained attraction in the 1980s and 1990s, it failed to meet expectations; only due to the effort of some few researches combined with improved hardware and the availability of bigger data sets, Deep Learning was able to produce several landmark improvements \citep{bengio2015deep}, which sent a ripple through adjacent fields, including computational linguistics.
Indicative for this is e.g.\ the work of \cite{bengio2003neural}, developing a neural network-based Language Model, which was able to outperform seasoned $n$-gram approaches by a big margin and also introduced the concept of \emph{word embeddings}, i.e.\ the representation of words as learned, real-valued vectors.

In recent years, Recurrent Neural Networks (\glspl{RNN}) have developed into a popular choice in Deep Learning to model sequential data, especially language. Although fully attention-based models without recurrency \citep{vaswani2017attention, radford2018improving, radford2019language} have lately become a considerable alternative, \gls{RNN}-based models still obtain state-of-the-art scores on tasks like Language Modeling \citep{gong2018frage}, Sentiment Analysis \citep{howard2018universal} and more. However, they still lack the two aforementioned capabilities present in humans when it comes to processing language: They do not take the uncertainty about their predictions and representations into account and they are not able to correct themselves based on new inputs.\\

The aim of this thesis is to augment a \gls{RNN} architecture to address both of these points: Firstly by developing a mechanism that corrects the model's internal representation of the input at every time step, similar to a human's correction when reading a garden path sentence. Secondly, this correction will be based on several methods, some of which like \emph{Monte Carlo Dropout} (MC Dropout) or \emph{Bayesian Anchored Ensembling} (\gls{BAE}) were developed in a line of research called \emph{Bayesian Deep Learning}, which is concerned with incorporating bayesian statistics into Deep Learning models. These methods help to quantify the degree of uncertainty in a model when it is making a prediction and can be used in combination with the correction mechanism to directly influence the amount of confidence the model has about its internal representations.

\section{Research Questions}

Building on the ideas of uncertainty and self-correction, this thesis is concerned with realizing them in a mathematical framework and followingly adapt the implementation of a \gls{RNN} language model with a mechanism that performs the layed-out adaptations of hidden representations.
From this intention, the subsequent research questions follow:\\

\textbf{How to realize corrections?} Gradient-based methods are a backbone of the success of Deep Learning since the inception of the first artificial neurons in the 1950s \citep{bengio2015deep} and its variants in their manifoldness still remain an important tool in practice today.\footnote{E.g.\ refer to \cite{ruder2016overview} for an overview over the most frequently used gradient-based optimizers.} However, these methods are usually only applied to the model \emph{parameters} after processing a batch of data. Frequent local adaptions of activations have - to the best of my knowledge - not been attempted or examined in the literature, with the single exception of the paper this work draws inspiration from \citep{giulianelli2018under}.\\

\textbf{On what basis should corrections be performed?} Another research question that follows as a logical consequence of the previous one focuses on the issue of the mechanism's modality: The described idea requires to take the gradient of some ``useful'' quantity with respect to the network's latent representations. This quantity can e.g.\ spring from additional data as in \cite{giulianelli2018under}, which results in a partly supervised approach. This supervision requires additional information about the task, which is not always available or suffers from all the problems that come with manually labeling data on scale: The need for expertise (which might not even produce handcrafted features that are helpful for models; deep neural networks often find solutions that seem unexpected or even surprising to humans) as well as a time- and resource-consuming process. This work thus examines different unsupervised approaches based on a word's perplexity and uncertainty measures from the works of \cite{gal2016dropout, gal2016theoretically, pearce2018uncertainty}.\\

\textbf{What is the impact on training and inference?} In the work of \cite{giulianelli2018under}, the idea has only been applied to a fully-trained model during test time. The question still remains how using such an extended model would behave during training as adapting activations in such a way will have a so far unstudied effect on the learning dynamics of a model. Furthermore, it is the aim of this thesis to also shed some light on the way that this procedure influences the model's behavior during inference compared to a baseline, which is going to be evaluated on a batch and word-by-word scale.

\section{Contributions}

In this thesis I make the following contributions:

\begin{itemize}
    \item I formulate a novel and flexible framework for local gradient-based adaptions called \emph{recoding}, including the proof of some theoretical guarantees.
    \item I implement this new framework in the context of a recurrent neural language model.\footnote{The code is available online only under \url{https://github.com/Kaleidophon/tenacious-toucan}.}
    \item I examine potential sources of unsupervised ``error signals'' that trigger corrections of the model's internal representations.
    \item Finally, I provide an in-depth analysis of the effect of recoding on the model during training and inference as well as a validation of theoretical results and an error analysis.
\end{itemize}

\section{Thesis structure}

The thesis is structured as follows: After this introduction, other relevant works about the linguistic abilities of \glspl{RNN}, Bayesian Deep Learning and more dynamic deep networks are being summarized in chapter \ref{chapter:related-work}. Afterwards, I provide some relevant background knowledge for the approaches outlined in this thesis in chapter \ref{chapter:background}, including a brief introduction to Deep Learning, Language Modeling with Recurrent Neural Networks and Bayesian Deep Learning.
A formalization of the core idea described above is given in chapter \ref{chapter:recoding-framework}, where I develop several variants of its implementations based on different error signals. These are subsequently evaluated in an experimental setting in chapter \ref{chapter:experiments}, where the impact of the choice of hyperparameters concerning recoding and importance of different model components is studied. A closer look at the behavior of the models on a word level is taken in chapter \ref{chapter:qualitative}. A full discussion of all experimental results is then given in chapter \ref{chapter:discussion}. Lastly, all findings and implications of this work are summarized in chapter \ref{chapter:conclusion}, giving an outlook onto future research. Additional supplementary material concerning relevant lemmata, extensive derivations and additional figures as well as more details about the experimental setup are given in appendices \ref{appendix:derivations} and \ref{appendix:additional}.

\chapter[Related Work]{Related Work}\label{chapter:related-work}

\epigraph{``The world (which some call the Library) is made up of an unknown, or perhaps unlimited, number of hexagonal galleries, each with a vast central
ventilation shaft surrounded by a low railing. From any given hexagon, the higher and lower galleries can be seen stretching away interminably."}{\vspace{0.4cm}\emph{The Library of Babel - Jorge Luis Borges}}

In this chapter I review some relevant works for this thesis regarding the linguistic abilities of recurrent neural network models as well as the state-of-the-art approaches for Language Modeling. A small discussion is dedicated to the role of some techniques for ``interpretable'' artificial intelligence and the influences from the area of Bayesian Deep Learning that are the foundation for the framework presented in chapter \ref{chapter:recoding-framework}. Lastly, I give an overview over a diverse set of ideas concerning more dynamic models and representations.\\

\section{Recurrent Neural Networks \& Language}\label{sec:related-work-rnn-language}

Recurrent Neural Networks \citep{rumelhart1988learning} have long been established as a go-to architecture when modelling sequential data like handwriting \citep{fernandez2009novel} or speech \citep{sak2014long}. One particularly popular variant of \glspl{RNN} is the \emph{Long-Short Term Memory Network} (\gls{LSTM}) \citep{hochreiter1997long}, where the network learns to ``forget'' previous information and add new to a memory cell, which is used in several gates to create the new hidden representation of the input and functions
as a sort of long-term memory. This has some special implications when applying recurrent networks to language-related tasks: It has been shown that these networks are able to learn some degree of hierarchical structure and exploit linguistic attributes \citep{liu2018lstms}, e.g.\ in the form of numerosity \citep{linzen2016assessing, gulordava2018colorless}, nested recursion \citep{bernardy2018can}, negative polarity items \citep{jumelet2018language}, or by learning entire artificial context-free languages \citep{gers2001lstm}.\\

The most basic of language-based tasks lies in modeling the conditional probabilities of words in a language, called \emph{Language Modeling}. Current state-of-the-art models follow two architectural paradigms: They either use a \gls{LSTM} architecture, for instance in the case of \cite{gong2018frage} or use another popular attention-centric approach which recently was able to produce impressive results. Attention, a mechanism to complement a model in a way that is able to focus on specific inputs was first introduced in the context of a recurrent encoder-decoder architecture \citep{bahdanau2014neural}.
In the work of \cite{vaswani2017attention}, this was extended to a fully (self-)attention-based architecture, which comprises no recurrent connections whatsoever. Instead, the model uses several attention mechanisms between every layer and makes additional smaller adjustments to the architecture to accommodate this structural shift. This new model type was further extended in \cite{dai2019transformerxl} and most recently in \cite{radford2019language}.\\

However, due to the sheer amount of parameters and general complexity of models, the way they arrive at their predictions is often not clear. In order to allow the deployment of models in real-life scenarios with possible implications for the security and well-being of others, it is thus of paramount importance to understand the decision process and learned input representations of models \citep{brundage2018malicious}.
One such tool to provide these insights into the inner workings of a neural network can be linear classifiers which are trained on the network's activations to predict some feature of interest \citep{hupkes2018visualisation, dalvi2019one, conneau2018you}, called Diagnostic Classifiers (\gls{DC}).
\cite{giulianelli2018under} use this exact technique to extend the work of \cite{gulordava2018colorless} and show when information about the number of the sentence's subject in \gls{LSTM} activations is being corrupted. They furthermore adjust the activations of the network based on the error of the linear classifier and are thus able to avoid the loss of information in some cases.\\

\section{Model Uncertainty}

Standard Deep Learning approaches do not consider the uncertainty of model predictions. This is especially critical in cases where deployed models are delegated important and possibly life-affecting decisions, such as in self-driving cars or medicine. Furthermore, incorporating the Bayesian perspective into Deep Learning reduces the danger of making overly confident predictions \citep{blundell2015weight} and allows for the consideration of prior beliefs to bias a model toward a desired solution. Therefore, a line of research emerged in recent years trying to harmonize established Deep Learning approaches with ideas from Bayesian inference. However, ideas that try sample weights from learned distributions \citep{blundell2015weight} are often unfeasible as they do not scale to the profound models and humongous datasets utilized in practice.\\

An influential work managed to exploit a regularization technique called Dropout \citep{srivastava2014dropout}, which deactives neurons randomly during training in order to avoid overfitting the training data, to model the approximate variational distribution over the model weights \citep{gal2016dropout}, which in turn can be used to estimate the predictive uncertainty of a model. These results have also been extended to recurrent models \citep{gal2016theoretically} and successfully applied in cases like time series prediction in a ride-sharing app \citep{zhu2017deep}.
Another promising approach to quantify uncertainty is exploiting model ensembles, as shown in \cite{lakshminarayanan2017simple}, which forgoes the variational approach. Technically, this approach cannot be considered Bayesian and was therefore extended using a technique called \emph{Randomized Anchored \gls{MAP} Sampling}, where uncertainty estimates can be procuded by employing an ensemble of Deep Neural Networks regularized using the same anchor noise distribution \citep{pearce2018uncertainty}.

\section{Dynamic Models \& Representations}

Lastly, there have also been some efforts to make models more adaptive to new data. Traditionally, models parameters are adjusted during a designated training phase. Afterwards, when a model is tested and then potentially deployed in a real-world enviromment, these parameters stay fixed. This carries several disadvantages: If the distribution of the data the model is confronted with differs from the training distribution or changes over time, the model performance might drop. Therefore, different lines of research are concerned with finding a solution that either enable continuous learning (like observed in humans or animals) or more flexible representations.\\

Some works directly attack the problem of differing data distributions and the transfer of knowledge across different domains \citep{kirkpatrick2017overcoming, achille2018life}.
Further ideas about continuous learning have been explored under the ``Life-long learning'' paradigm e.g.\ in Reinforcement Learning \citep{nagabandi2018learning} or robotics \citep{koenig2017robot}.
In Natural Language Processing, one such idea consists of equipping a network with a self-managed external memory, like in the works of \cite{sukhbaatar2015end} and \cite{kaiser2017learning}. This way allows for a way to adjust hidden representations based on relevant information learned during training.
In contrast, the work of \cite{krause2018dynamic} has helped models to achieve state-of-the-art results in Language Modeling by making slight adjustments to the models' parameters when processing unseen data. It should be noted that these approaches directly target model parameters, while this work is concerned with adjusting model \emph{activations}, which, to the best of my knowledge, has not been attempted in the way that it is being presented here.\\

The next chapter goes more into the technical details of related works.

\chapter[Background]{Background}\label{chapter:background}

\epigraph{``As well as any human beings could, they knew what lay behind the cold, clicking, flashing face -- miles and miles of face -- of that giant computer. They had at least a vague notion of the general plan of relays and circuits that had long since grown past the point where any single human could possibly have a firm grasp of the whole.
Multivac was self-adjusting and self-correcting. It had to be, for nothing human could adjust and correct it quickly enough or even adequately enough."}{\vspace{0.4cm}\emph{The Last Question - Isaac Asimov}}

In this chapter, I supply the reader with some background knowledge relevant to the content of this work, from the basic concepts of Deep Learning in section \ref{sec:deep-learning} and their extension to language modeling in section \ref{sec:rnlm}. In section \ref{sec:bayesian-deep-learning} I try to convey the intuition of Bayesian Deep Learning, which helps to quantify the uncertainty in a network's prediction. I also outline approaches derived from it used in this thesis as possible signals for activation corrections which approximate this very quantity.

\section{Deep Learning}\label{sec:deep-learning}

Deep Learning has achieved impressive results on several problems in recent years, be it high-quality machine translation \citep{edunov2018understanding}, lip reading \citep{van2016wavenet}, protein-folding \citep{evans2018novo} and beating human opponents in complex games like Go \citep{silver2016mastering}, just to name a few examples among many. On a theoretical level, Deep Neural Networks (\gls{DNN}) have been shown to be able to approximate any continuous function under certain conditions \citep{hornik1989multilayer, csaji2001approximation}, which helps to explain their flexibility and success on vastly different domains.\\

These \glspl{DNN} consist of an arbitrary number of individual layers, each stacked upon one another. The layers are feeding off increasingly complex \emph{features} - internal, learned representations of the model realized in real-valued vectors - that are the output produced by the preceeding layer. A single layer in one of these networks can be described as an affine transformation of an input $\bx$ with a non-linear \emph{activation function} $\sigma$ applied to it

\begin{equation}
    \bx^{(l+1)} = \sigma(\bW^{(l)}\bx^{(l)} + \bb^{(l)})
\end{equation}

where $\bx^{(l)}$ describes the input of the $l$-th layer of the network and $\bW^{(l)}$ and $\bb^{(l)}$ are the weight and bias of the current layer, which are called learned parameters of the model. Using a single weight vector $\bw$, this formulation corresponds to the original Perceptron by \cite{rosenblatt1958perceptron}, a simple mathemathical model for a neuron cell which weighs each incoming input with a single factor in order to obtain a prediction. Using a weight \emph{matrix} $\bW$ therefore signifies using multiple artificial neurons in parallel. The non-linear activation function $\sigma(\cdot)$ plays a pivotal role: Stacking up linear transformation results in just another linear transformation in the end, but by inserting non-linearities we can achieve the ability to approximate arbitrary continuous functions like mentioned above. Popular choices of activation functions include the sigmoid function, tangens hyperbolicus (tanh), the rectified linear unit (\gls{ReLU}) as well as the Softplus function, \gls{ReLU}'s continuous counterpart:

\begin{equation}
  \text{sigmoid}(x) = \frac{e^x}{e^x + 1}
\end{equation}
\begin{equation}
  \text{tanh}(x) = \frac{2}{1 + e^{-2x}} - 1
\end{equation}
\begin{equation}
  \text{relu}(x) = \max(0, x)
\end{equation}
\begin{equation}\label{eq:softplus}
  \text{softplus}(x) = \log(1 + e^x)
\end{equation}

The network itself is typically improved by first computing and then trying to minimize the value of an objective function or \emph{loss}. This loss is usually defined as the deviation of the model's prediction from the ground truth, where minimizing the loss implies improving the training objective (a good prediction). A popular choice for this loss is the minimum squared error (\gls{MSE}) loss, where we compute the average squared distance between the models prediction $\yhat_i$ for a datapoint $\bx_i$ and the actual true prediction $y_i$ using the $l_2$-norm:

\begin{equation}
  \mathcal{L}_{\text{MSE}} = \frac{1}{N}\sum_{i=1}^N ||y_i - \yhat_i||^2
\end{equation}

However in practice, many other loss functions are used depending on the task, architecture and objective at hand.
This loss can then be used by the Backpropagation algorithm \citep{rumelhart1988learning}, which recursively computes the gradient $\nabla_\btheta \mathcal{L}_{\text{MSE}}$, a vector of all partial derivatives of the loss with respect to every single of the model's parameters $\btheta$. It is then used to move the same parameters into a direction which minimizes the loss, commonly using the gradient descent update rule:

\begin{equation}
    \btheta^\prime = \btheta - \eta \nabla_\btheta \mathcal{L}_{\text{MSE}}
\end{equation}

\begin{figure}[h]
  \centering
  \begin{tabular}{cc}
    \includegraphics[width=0.45\textwidth]{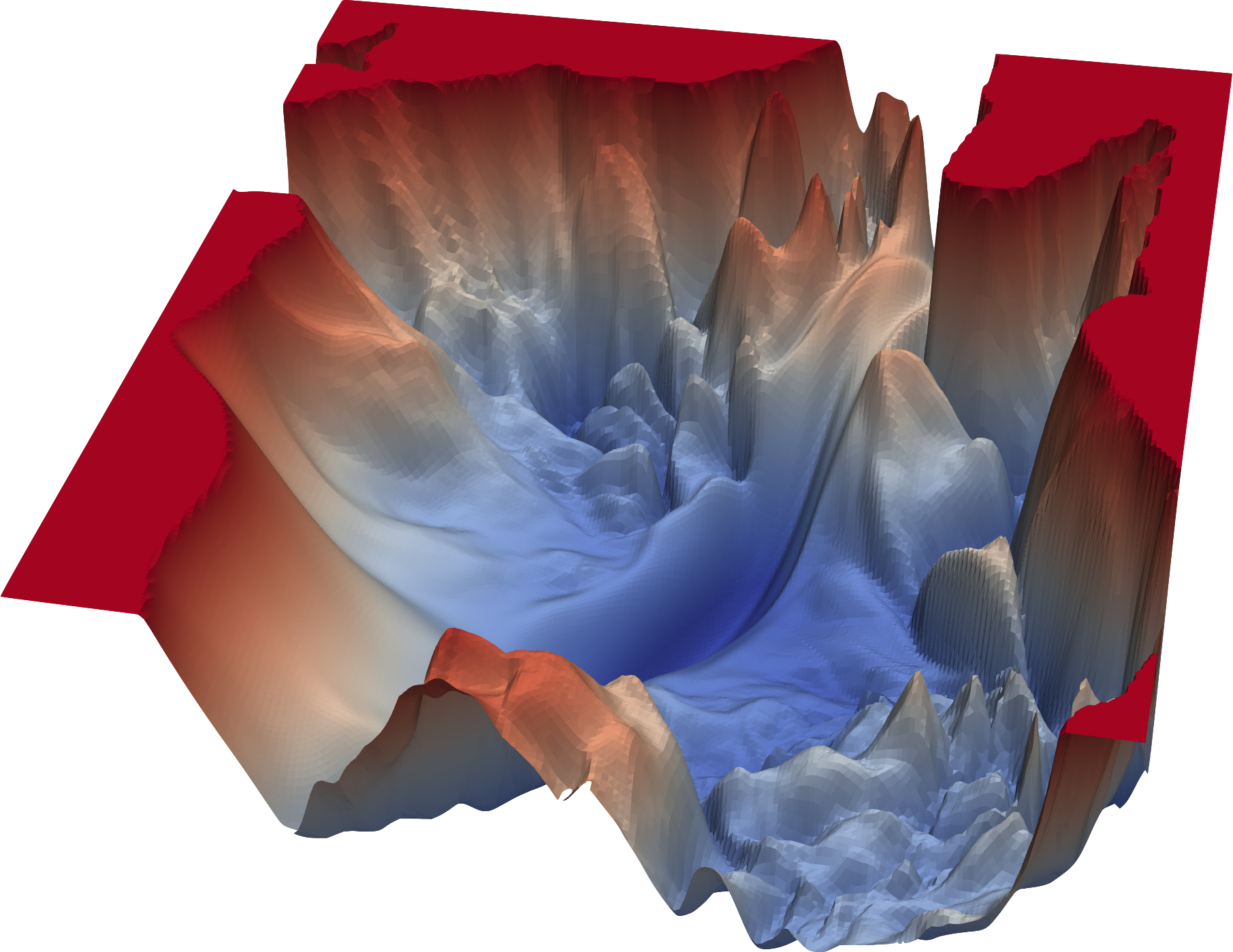} & \includegraphics[width=0.45\textwidth]{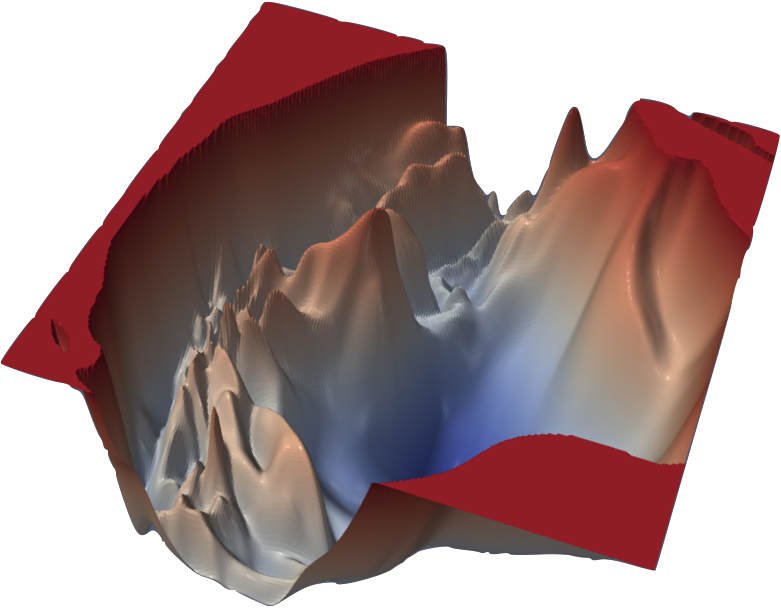}
  \end{tabular}
  \caption[Visualized loss surfaces of two deep neural networks]{Visualized loss surfaces of two deep neural networks. The ``altitude'' of the surfaces is determined by the value of the loss function the parameters of the models with the corresponding $x$- and $y$-values produce. This figure only depicts the projection of the loss surface into 3D space, as neural networks contain many more than just two parameters. Images taken from \cite{li2018visualizing}. Best viewed in color.}\label{fig:loss-surfaces}
\end{figure}

where $\eta$ is a positive learning rate with $\eta \in \mathbb{R}^+$. This effectively means that we evaluate the quality of the network parameters at every step by looking at the prediction they produced. The gradient points into the direction of steepest ascend. Consequently, the antigradient ($-\nabla_\btheta \mathcal{L}_{\text{MSE}}$) points us into the direction of the steepest \emph{descend}. Updating the parameters into this direction means that we seek to produce a set of new parameters $\btheta^\prime$ which theoretically decreases the value of the loss function during the next step.\\

It is possible for convex loss functions to find their global minimum, i.e.\ the set of parameters that produces a loss that is lower than all other possible losses, in a single step (cp. \cite{bishop2006pattern} \S 3). However, as illustrated in figure \ref{fig:loss-surfaces}, neural networks are known to have highly non-convex loss functions - highly dimensional, rugged loss surfaces with many local minima and an increasing number of saddle points as the dimensionality of the parameter space (the number of parameters in the model) increases \citep{dauphin2014identifying}. We therefore only take a step of length $\eta$ into the direction of the antigradient. Using a suitable learning rate $\eta$, this is guaranteed to converge to a local minimum \citep{nesterov2018lectures}. Although this local mininum is rarely the global minimum, the found solution often suffices to solve the task at hand in a satisfactory manner.

\section{Recurrent Neural Language Models}\label{sec:rnlm}

In this section I provide some insight into the motivation for and inner workings of Recurrent Neural Networks (\ref{sec:rnn}), an extension to standard \glspl{MLP} especially suited for sequential data. Also, I outline their application to Language Modeling (\ref{sec:lm}), where we try to model the probability distribution that human language is ``generated'' from and which will be the main objective of this thesis.

\subsection{Recurrent Neural Networks}\label{sec:rnn}

\emph{Recurrent Neural Networks} (\glspl{RNN}) are a special type of Deep Neural Networks which comprises recurrent connections, meaning that they produce new representations based on their prior representations.
Due to this recurrent structure, they present an inductive bias in their architecture which makes them especially suitable for modeling sequences of data \citep{battaglia2018relational}.\\

Formally, a sequence of input symbols like words is mapped to a sequence of continuous input representations $\mathcal{X} = \langle\bx_1, \ldots, \bx_T\rangle$ with $\bx_t \in \mathbb{R}^M$ called \emph{embeddings}, real-valued vector representations for every possible symbol that is used as an input to the model. When dealing with language, every word in the vocabulary of a corpus is mapped to a \emph{word embedding}. These representations are initialized randomly and then tuned alongside the model's parameters during training. Furthermore, a recurrent function $g(\cdot)$ parameterized by $\btheta$ is used to compute hidden representations $\bh_t$ based on the current input representation $\bx_t$ and the last hidden representation $\bh_{t-1}$ with $\bh_t \in \mathbb{R}^N$:

\begin{equation}\label{eq:recurrency}
    \bh_t = g_{\btheta}(\bx_t, \bh_{t-1})
\end{equation}

Considering that the hidden activations $\bh_t$ depend on their predecessors thus allows the \gls{RNN} to capture information across multiple processing steps. This recurrent function $g_\btheta(\cdot)$ can be realized in many different ways. In its simplest, most commonly used form, it is stated as\\

\begin{equation}\label{eq:vanilla-rnn}
    \bh_t = \text{tanh}(\bW_{hx}\bx_t + \bb_{hx} + \bW_{hh}\bh_{t-1} + \bb_{hh})
\end{equation}

with $\btheta = \{\bW_{hx}, \bW_{hh}, \bb_{hx}, \bb_{hh}\}$ as the learnable parameters. In a classification setting, where we try to predict the most likely output category, we can produce a categorical probability distribution over all possible classes in the following way:
For every hidden state $\bh_t$, an affine transformation is applied to create the distribution over possible output symbols $\bo_t$ (eq. \ref{eq:vanilla-rnn-proj}) with $\bo_t \in \mathbb{R}^{|\mathcal{V}|}$, where $\mathcal{V}$ denotes the set of all symbols, or vocabulary. As this distribution usually does not meet the requirements of a proper probability distribution, it is then typically normalized using the \emph{softmax} function, so that the total probability mass amounts to $1$ (eq. \ref{eq:vanilla-rnn-softmax})

\begin{equation}\label{eq:vanilla-rnn-proj}
    \botilde_t = \bW_{ho}\bh_t + \bb_{ho}
\end{equation}
\begin{equation}\label{eq:vanilla-rnn-softmax}
    \bo_{ti} = \frac{\exp(\botilde_{ti})}{\sum_{j=1}^J\exp(\botilde_{tj})}
\end{equation}

where $\bW_{ho} \in \mathbb{R}^{|\mathcal{V}| \times N}$ and $\bb_{ho} \in \mathbb{R}^{|\mathcal{V}|}$.

When training this kind of network, it is often observed that gradients shrink more and more the longer the input sequence is, not allowing for any meaningful improvements of the parameters, which is called the \emph{vanishing gradient problem}: When calculating the gradients for the network's parameters, a modification to the earlier introduced backpropagation algorithm, \emph{backpropagation through time} (\gls{BPTT}) has to be used, as the recurrency of the network conditions later outputs on earlier processes. If we now evaluate the resulting product of partial derivatives, we often find gradients to tend to zero when the network's error is sufficiently small.\footnote{Or in other words, $\lim\limits_{d \to \infty}\nabla_\theta\mathcal{L}^d = 0$ holds for the gradient of the parameters and an input lag $d$, i.e.\ the number of time steps the error is being propagated back in time, if the spectral radius (or biggest absolute eigenvalue) of $\nabla_\theta\mathcal{L}$ is smaller than one \citep{hochreiter2001gradient}.}\\

The Long-Short Term Memory network (\gls{LSTM}) by \cite{hochreiter1997long}, which is the basis for the models used in this work, tries to avoid this problem by augmenting the standard \gls{RNN} recurrency by adding several gates:

\begin{equation}\label{eq:forget-gate}
    \mathbf{\underline{f}}_t = \sigma(\bW_{h\underline{f}}\bh_{t-1} + \bW_{x\underline{f}}\bx_t + \bb_{h\underline{f}} + \bb_{x\underline{f}})
\end{equation}
\begin{equation}\label{eq:input-gate}
    \mathbf{\underline{i}}_t = \sigma(\bW_{h\underline{i}}\bh_{t-1} + \bW_{x\underline{i}}\bx_t + \bb_{h\underline{i}} + \bb_{x\underline{i}})
\end{equation}
\begin{equation}\label{eq:output-gate}
    \mathbf{\underline{o}}_t = \sigma(\bW_{h\underline{o}}\bh_{t-1} + \bW_{x\underline{o}}\bx_t + \bb_{h\underline{o}} + \bb_{x\underline{o}})
\end{equation}
\begin{equation}\label{eq:cell-update}
    \mathbf{c}_t = \mathbf{\underline{f}}_t \circ \mathbf{c}_{t-1} + \mathbf{\underline{i}}_t \circ \text{tanh}(\bW_{hc}\bh_{t-1} + \bW_{xc}\bx_t + \bb_{hc} + \bb_{xc})
\end{equation}
\begin{equation}\label{eq:lstm-hidden}
    \bh_t = \mathbf{\underline{o}}_t \cdot \text{tanh}(\mathbf{c}_t)
\end{equation}

which are called the forget gate $\mathbf{\underline{f}}_t$, input gate $\mathbf{\underline{i}}_t$ and output gate $\mathbf{\underline{o}}_t$ (not to be confused with the output \emph{activations} $\bo_t$). A new type of activations, the cell state $\mathbf{c}_t$, is also added. The activations of all gates consist of linear combinations of affine transformations of the last hidden state $\bh_{t-1}$ and the current input $\bx_t$ (\cref{eq:forget-gate,eq:input-gate,eq:output-gate}).\\

However, each of them is assigned a different role within the cell's architecture: The input gate determines to which extent information is flowing into the cell state, the forget gates modulates the amount of information inside the cell state that is being erased and the output gate defines how much information flows from the cell state to the new hidden state $\bh_t$ (eq. \ref{eq:lstm-hidden}). The role of the cell state $\mathbf{c}_t$ is to act as a form of memory storage that is being updated based on the current input.
Not only does this enable the \gls{LSTM} to model more long-term dependencies in its input sequence\footnote{This is what seems to give rise to all kinds of linguistic abilities in \glspl{LSTM}, see e.g.\ \cite{liu2018lstms, linzen2016assessing, gulordava2018colorless, bernardy2018can, jumelet2018language}.}, the formulation of the cell state update in eq.~\ref{eq:cell-update} allows gradients to still ``flow'' through the forget gate when computing the derivative $\partial \mathbf{c}_t / \partial \mathbf{c}_{t-1}$, which in turn avoids vanishing gradients.\\

\subsection{Language Modeling with \glspl{RNN}}\label{sec:lm}

In Language Modeling we try to model the probalities of words an sequences. Considering a sequence of $T$ words (or sentence) $\mathcal{S} = \langle w_1, \ldots, w_T\rangle$, the question arises how to assign a probability to it.
The joint probability $p(w_1^T)$ is commonly factorized into the product of the conditional probabilities of every word given all its predecessors

\begin{equation}
    p(w_1^T) = p(w_1)p(w_2|w_1)p(w_3|w_2, w_1)\ldots = p(w_1)\prod_{t=2}^T p(w_t|w_1^{t-1})
\end{equation}

As shown in the previous section, the output probability distribution over words $\bo_{t}$ is dependent on the hidden representation $\bh_t$, which is in turn dependent on the current input embedding for a word $\bx_t$ and the previous hidden state $\bh_{t-1}$. Therefore, we can see the \gls{RNN} modeling this exact conditional probability $p(w_t|w_1^{t-1})$, where the output distribution tries to predict the next word in the sequence. That means that the probability of the next word $w_{t+1}$ being the $c$-th word in the vocabulary $\mathcal{V}$ assigned by the model can be written as follows:

\begin{equation}
  p(w_{t+1}=c|w_1^t) = \bo_{tc}
\end{equation}

To train the model parameters $\btheta$, a loss function $\mathcal{L}$ then measures the difference of the output distribution $\bo_t$ to the target prediction, or in this case, the actual following word, $w_{t+1}$. Then, the gradient descent rule from the previous section (or any other update rule) is applied. For language modeling, the most popular choice of loss function is the cross-entropy loss:

\begin{equation}\label{eq:cross-entropy}
  \mathcal{L}_{CE} = - \frac{1}{T}\sum_{t=1}^T\sum_{c=1}^{|V|}\indicator{w_{t+1}=c}\log(\bo_{tc})
\end{equation}

where $\indicator{\cdot}$ is a binary function that is $1$ when $\bo_{tc}$ corresponds to the probability of the actually observed token at time step $t+1$ and $0$ otherwise. Intuitively, this loss measures the difference between two probability distributions and penalizes when the predicted output distribution $\bo_{t}$ deviates from a target distribution, where only the observed token should be assigned a probability of $1$.

\section{Bayesian Deep Learning}\label{sec:bayesian-deep-learning}

In this section I deliver some information about the incorporation of Bayesian statistics into Deep Learning. Bayesian statistics allows us to also express a degree of belief, such as a personal prior belief about the probability of an event happening. This differs from the purely frequentist approach, which simply observes the likelihood of certain events. Transferring this framework to Deep Learning introduces some problems - these and a possible mitigation are layed out in section \ref{sec:vi}. Methods to estimate the confidence of a model in its predictions using methods based on Bayesian Deep Learning are outlined in sections \ref{sec:confidence-dropout} and \ref{sec:confidence-ensembles}, which will be exploited in later chapters by the recoding mechanism.

\subsection{Variational inference}\label{sec:vi}

Regular feedforward neural networks are prone to overfitting and do not incorporate uncertainty into their predictions, which often results in overly confident predictions \citep{blundell2015weight}. This problem can be alleviated by trying to harmonize models with the Bayesian approach: The training regime explained in the previous section is equivalent to retrieve the \emph{Maximum Likelihood Estimate} (\gls{MLE}) of the model parameters $\btheta$ given a data set $\mathcal{D}_{\text{train}} = \{(\bx_i, y_i)\}_1^N$ of data points $\bx_i$ and corresponding labels $y_i$:

\begin{equation}
    \btheta_{MLE} = \argmax_{\btheta} \log p(\mathcal{D}|\btheta) =  \argmax_{\btheta}\sum_{i=1}^N \log p(y_i|\bx_i, \btheta)
\end{equation}

Intuitively, this solution retrieves the parameters that explain the data best, or maximize the \emph{likelihood}  $p(\mathcal{D}|\btheta)$. However, one can also try to find the most likely parameters given a data set $p(\btheta|\mathcal{D})$, which can be obtained by using Bayes' theorem:

\begin{equation}
  p(\btheta|\mathcal{D}) = \frac{p(\mathcal{D}|\btheta)p(\btheta)}{p(\mathcal{D})}
\end{equation}

where we call $p(\btheta|\mathcal{D})$ the \emph{posterior} distribution, $p(\btheta)$ the \emph{prior} and $p(\mathcal{D})$ the \emph{evidence}. Maximizing this quantity (or obtaining the \emph{Maximum a posteriori estimate} (\gls{MAP})) by finding the best model parameters $\btheta$, we can ignore the evidence, because it is not dependent on them:

\begin{equation}
      \btheta_{MAP} = \argmax_{\btheta} \log p(\btheta|\mathcal{D}) = \argmax_{\btheta} \log p(\mathcal{D}|\btheta) + \log p(\btheta)
\end{equation}

Unfortunately, a problem arises when trying to perform exact Bayesian inference on a new data point $\bx^\prime$ in order to predict its label $y^\prime$. This requires one to evaluate the predictive distribution

\begin{equation}\label{eq:intractable-post}
    p(y^\prime|\bx^\prime) = \mathbb{E}_{p(\btheta|\mathcal{D})}[p(y^\prime | \bx^\prime, \btheta)] = \int p(y^\prime | \bx^\prime, \btheta) p(\btheta|\mathcal{D}) d\btheta
\end{equation}

because of the fact that the model parameters have to be marginalized out:

\begin{equation}
\int p(y^\prime | \bx^\prime, \btheta) p(\btheta|\mathcal{D}) d\btheta = \int p(y^\prime, \btheta| \bx^\prime) d\btheta = p(y^\prime| \bx^\prime)
\end{equation}

This however is intractable in most practical scenarios as we have to integrate over all values of the network parameters $\btheta$. Overcoming this problem is the main motivation for using variational inference: We approximate the posterior $p(\btheta|\mathcal{D})$ with an easy to evaluate variational distribution
$q_\bphi(\btheta)$ parameterized by variational parameters $\bphi$. To find the optimal parameters $\bphi^*$ (and therefore the distribution $q_\bphi(\btheta)$ that approximates the true posterior best), we try to minimize
the evidence lower bound (\gls{ELBO}) or variational free energy

\begin{equation}
    \bphi^* = \argmin_\bphi \text{KL}[q_\bphi(\btheta)||p(\btheta)] - \mathbb{E}_{q_\bphi(\btheta)}[\log p(\mathcal{D}|\btheta)]
\end{equation}

which can be derived from evaluating $\text{KL}[q_\bphi(\btheta)||p(\btheta|\mathcal{D})]$ (for the full derivation see appendix \ref{appendix:elbo}). $\text{KL}[\cdot||\cdot]$ here denotes the Kullback-Leibler divergence, an asymmetric similarity measure to quantify the distance between two probability distributions, in our case the real posterior distribution and the easier substitute we choose.\\

\cite{blundell2015weight} uses this framework to propose \emph{Bayes by Backprop}, a way to use the Backpropagation algorithm to learn \emph{distributions} over weights instead of point estimates. However, this approach requires the model to store additional distribution parameters and results in overall slower training. Therefore I will showcase other works trying to find more scalable ways to achieve this in the next section.

\subsection{Estimating model confidence with Dropout}\label{sec:confidence-dropout}

Dropout \citep{srivastava2014dropout} is a popular stochastic regularization technique utilized to avoid overfitting in neural networks. It is applied during training by randomly cutting out varying neurons, forcing the network to not rely on specific units for a successful prediction.\\

It was shown in the work of \cite{gal2016dropout} that applying this scheme can be seen as variational inference in Deep Gaussian processes. The intractable posterior in equation \ref{eq:intractable-post} is now approximated by a variational distribution over the model weight matrices $q(\bomega)$, whose elements are randomly set to zero\footnote{As this distribution is not put onto the model's biases, we denote the set of weight matrices $\bomega$ to distinguish it from all model parameters $\btheta$.}, according to the Dropout training regime. From there, they try to match the moments  (i.e.\ the mean and variance) of the approximate predictive distribution using $K$ Monte Carlo estimates

\begin{equation}
    \mathbb{E}_{q(y^\prime|\bx^\prime)}[f_\btheta(\bx^\prime)] \approx \frac{1}{K}\sum_{k=1}^K f_\btheta(\bx^\prime,  \bomega^{(k)})
\end{equation}

\begin{equation}\label{eq:mcd-var}
    \text{Var}_{q(y^\prime|\bx^\prime)}[f_\btheta(\bx^\prime)] \approx \tau^{-1}\mathbf{I}_D + \frac{1}{K}\sum_{k=1}^K f_\btheta(\bx^\prime,  \bomega^{(k)})^Tf_\btheta(\bx^\prime,  \bomega^{(k)})
\end{equation}

which they refer to as \emph{MC Dropout} (\gls{MCD}). $f_\btheta(\bx^\prime,  \bomega^{(k)})$ simply denotes the prediction of the network for the new data point using a set of weight matrices $\bomega^{(k)}$ with a different dropout mask. The mask itself is sampled from a multivariate Bernoulli distribution, according to the original description of the technique in \cite{srivastava2014dropout}. The first term in the sum of eq. \ref{eq:mcd-var}, $\tau$, denotes the model precision, which corresponds to

\begin{equation}
    \tau = \frac{pl^2}{2N\lambda}
\end{equation}

This term results from using Dropout in order to obtain the approximate posterior distribution and using a weight prior of the form $p(\bomega) = \mathcal{N}(0, l^{-2}\mathbf{I})$, with $p$ being the dropout probability, $l$ the length scale of the weight prior distribution (expressing our prior belief about the realization of the model weights), $N$ the number of training samples and $\lambda$ the weight decay parameter\footnote{Placing a Gaussian prior on the weights is known to result in $l_2$-regularization, also called \emph{weight decay} \citep{mackay2003information}.}. In a regression setting, this finally results in a predictive variance approximated by

\begin{equation}\label{eq:pred-var}
    \text{Var}_{q(y^\prime|\bx^\prime)}[f_\btheta(\bx^\prime)] \approx \tau^{-1}\mathbf{I}_D + \frac{1}{K}\sum_{k=1}^K f_\btheta(\bx^\prime,  \bomega^{(k)})^Tf_\btheta(\bx^\prime,  \bomega^{(k)}) - \mathbb{E}_{q(y^\prime|\bx^\prime)}[y^\prime]^T\mathbb{E}_{q(y^\prime|\bx^\prime)}[y^\prime]
\end{equation}

In other words, in order to approximate the predictive variance of the variational posterior, we conveniently only have to perform $K$ forward passes with our model, each one of them using a different Dropout mask. This whole approach is applicable to \glspl{MLP}. Because this work is concerned with Recurrent Neural Networks, the corresponding extension of this approach is also being discussed.

\begin{figure}[h]
  \centering
  \includegraphics[width=0.9\textwidth]{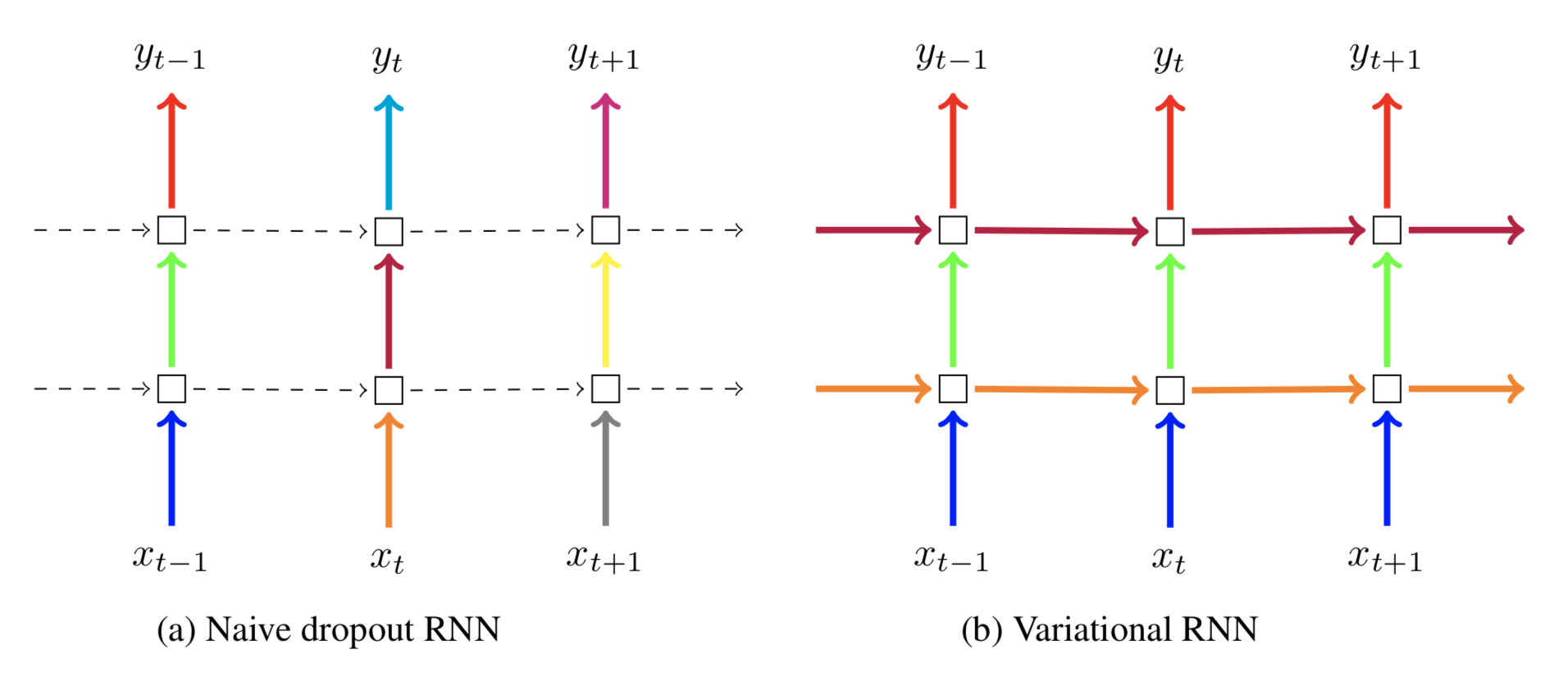}
  \caption[Comparison of a common application of Dropout to \glspl{RNN} with \cite{gal2016theoretically}]{Figure taken from \cite{gal2016theoretically}: While traditional approaches apply different kind of dropout masks to only a select set of connections (left), the variational \gls{RNN} Dropout approach always applies the same set of masks to the same connection when processing a sequence.}\label{fig:variational-rnn}
\end{figure}

When applying Dropout to \glspl{RNN}, \cite{gal2016theoretically} show that it is inhibitive to sample a different mask for every connection at every time step. Instead, they propose a scheme where only a single set of Dropout masks is sampled for every batch, reusing the same mask for the same type connection while processing the sequence (see fig. \ref{fig:variational-rnn}). The authors refer to this as the \emph{Variational \gls{RNN}}.
Thus the same way of estimating the predictive uncertainty as above can be used, with the exception that $f_\btheta(\bx^\prime,  \bomega^{(k)})$ now refers to the prediction of the network at time step $t$ and that $\bomega^{(k)}$ refers to the set of all weight matrices in the \gls{RNN} with the same set of Dropout masks applied to it throughout the whole input\footnote{$\bomega^{(k)}_{\text{RNN}} = \{\bW_{hx}^{(k)}, \bW_{hh}^{(k)}, \bW_{ho}^{(k)}\}$ for a vanilla \gls{RNN}.}.\\

Lastly, because in Language Modeling we are trying to predict the probability of a class, the method to estimate the model's confidence has to be adapted accordingly, as the probability vector that is the output of the model does not capture model confidence \citep{gal2016uncertainty}. Instead, we can consider the \emph{predictive entropy} of a model, which - in the case of a discrete probability distribution - is defined as

\begin{equation}\label{eq:pred-entropy}
    \mathbb{H}[y|\bx, \mathcal{D}_{\text{train}}] = - \sum_{c} p(y=c|\bx, \mathcal{D}_{\text{train}})\log p(y=c|\bx, \mathcal{D}_{\text{train}})
\end{equation}

Entropy is a measure from the field of information theory and captures the amount of information contained in a probability distribution. Minimum entropy is reached when all the mass of the distribution rests on a single class, the maximum value is attained when the underlying distribution is uniform. Again, the term $p(y=c|\bx, \mathcal{D}_{\text{train}})$ is intractable, but can be approximated using $K$ forward passes of the model with different Dropout masks (a full proof of this is given in appendix \ref{appendix:pred-entropy}), where $p(y=c|\bx, \bomega^{(k)})$ denotes the probability of the $c$-th word produced by a single forward pass with a Dropout mask:

\begin{equation}\label{eq:pred-entropy-approx}
  p(y=c|\bx, \mathcal{D}_{\text{train}}) \approx \frac{1}{K}{\displaystyle \sum_{k=1}^K}p(y=c|\bx, \bomega^{(k)})
\end{equation}

Therefore, when the mean distribution of these $K$ forward passes is almost uniform, the model is not confident which class to predict; conversely, if all passes amass all the probability on a single class, the model is confident that this class corresponds to the right prediction. In the next section I will lay out another, non-variational approach to obtain this quantity.

\subsection{Estimating model confidence with ensembles}\label{sec:confidence-ensembles}

\cite{pearce2018uncertainty} choose another, Bayesian but non-variational approach to estimating the predictive uncertainty by using ensembles: Although using an ensemble of deep models had already been realized before \citep{lakshminarayanan2017simple}, it lacked a Bayesian motivation. \cite{pearce2018uncertainty} achieve to overcome this blemish by a largely unknown inference method called
\emph{randomized \gls{MAP} sampling}: It is known that for a multivariate normal likelihood with parameters $\bmu_{\text{like}}$, $\bSigma_{\text{like}}$ and prior distribution with $\bmu_{\text{prior}}$, $\bSigma_{\text{prior}}$, the maximum a posteriori solution for the mean $\bmu$ is given by

\begin{equation}
    \bmu_{\text{MAP}} = (\bSigma_{\text{like}}^{-1} + \bSigma_{\text{prior}}^{-1})^{-1}(\bSigma_{\text{like}}^{-1}\bmu_{\text{like}} + \bSigma_{\text{prior}}^{-1}\bmu_{\text{prior}})
\end{equation}

In randomized \gls{MAP} sampling, a distribution of $\bmu_{\text{MAP}}$ solutions is produced by using a noise-injecting mechanism. As the source of noise, we replace $\bmu_{\text{prior}}$ with a random noise variable $\btheta_0$, which produces a function $\mathbf{f}_{\text{MAP}}$ based on the current $\btheta_0$:

\begin{equation}
   \mathbf{f}_{\text{MAP}}(\btheta_0) = (\bSigma_{\text{like}}^{-1} + \bSigma_{\text{prior}}^{-1})^{-1}(\bSigma_{\text{like}}^{-1}\bmu_{\text{like}} + \bSigma_{\text{prior}}^{-1}\btheta_0)
\end{equation}

This way, every time $\mathbf{f}_{\text{MAP}}$ is applied to a noise variable $\btheta_0$, it produces a new $\bmu_{\text{MAP}}$ solution. The parameters of the \emph{anchor noise distribution} that the variable $\btheta_0$ is sampled from approximates the true posterior $\mathcal{N}(\bmu_{\text{MAP}}, \bSigma_{\text{MAP}})$. It can then be found by (see \cite{pearce2018uncertainty}, appendix A for full derivation)

\begin{equation}\label{eq:back-mu-prior}
    \bmu_0 = \bmu_{\text{prior}}
\end{equation}
\begin{equation}\label{eq:back-sigma-prior}
    \bSigma_0 = \bSigma_{\text{prior}} + \bSigma_{\text{prior}}^2\bSigma_{\text{like}}^{-1}
\end{equation}

\begin{figure}[h]
  \centering
  \includegraphics[width=0.975\textwidth]{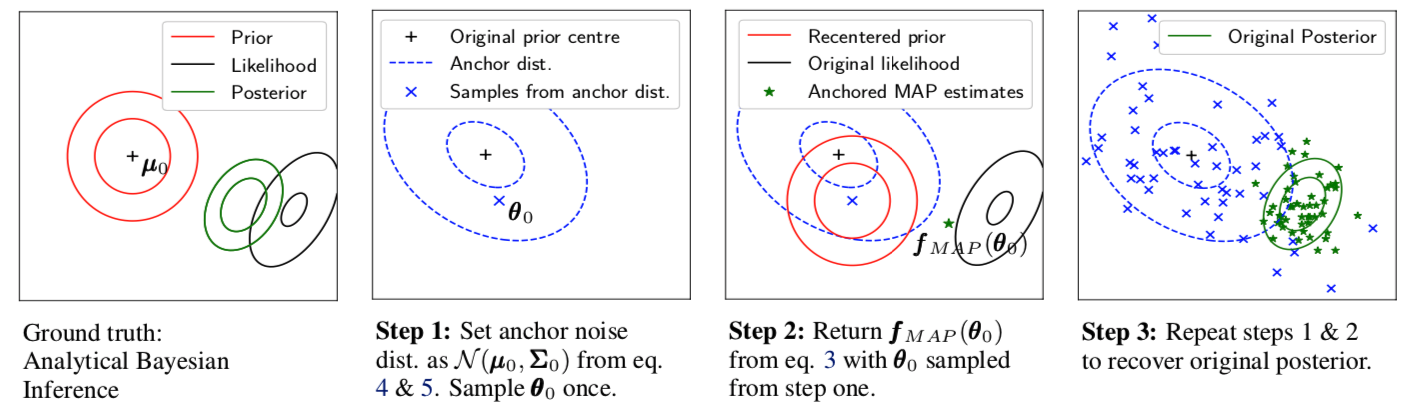}
  \caption[Illustration of the randomized \gls{MAP} sampling procedure]{Illustration of the randomized MAP sampling procedure, taken from the work of \cite{pearce2018uncertainty}.}\label{fig:randomized-map}
\end{figure}

Therefore, repeatedly sampling $\btheta_0 \sim \mathcal{N}(\bmu_0, \bSigma_0)$ and returning $ \mathbf{f}_{MAP}(\btheta_0)$ recovers the original posterior, as illustrated in figure \ref{fig:randomized-map}. Transferring this concept to deep neural networks, it is known that a training loss with weight decay can be interpreted as \gls{MAP} estimates with a normal prior centered at zero \citep{mackay2003information}. In a classification setting, this results in a cross-entropy loss of the form

\begin{equation}
  \mathcal{L}_{\text{CE}} = - \frac{1}{T}\sum_{t=1}^T\sum_{c}\indicator{y=c}\log p(y=c) + \frac{1}{T}||\bGamma^{1/2}\btheta||
\end{equation}

where $\bGamma$ is a diagonal matrix of the weight decay parameter $\lambda$. We can modify this loss to return \gls{MAP} estimates with prior estimates not centered at zero values:

\begin{equation}
  \mathcal{L}^{(k)}_{\text{CE}} =  - \frac{1}{T}\sum_{t=1}^T\sum_{c}\indicator{y=c}\log p(y=c) + \frac{1}{T}||\bGamma^{1/2}(\btheta^{(k)} - \btheta^{(k)}_0)||
\end{equation}

for every member $k$ of a ensemble of $K$ \glspl{NN}. Finally, in order to draw $\btheta_0^{(k)}$ with the parameters from equations \ref{eq:back-mu-prior} and \ref{eq:back-sigma-prior}, we approximate the latter as $\bSigma_0 = \bSigma_{prior}$, where \cite{pearce2018uncertainty} show that this approximation improves as the correlation between network parameter increases. Using this loss for every member of the ensemble simply ensures that we still obtain the real posterior.\\

In order to now estimate the confidence of the model, we can take the output distribution of every member in the ensemble to estimate the predictive entropy in eq. \ref{eq:pred-entropy} by using the approximation in eq. \ref{eq:pred-entropy-approx}. A justification for this in given in appendix \ref{appendix:pred-entropy}.

\section{Recap}

In this chapter I layed out the theoretical foundations for the next chapter. The main model type used in this work, the \gls{LSTM} network described in \ref{sec:rnn}, is especially suited for the task of Language Modeling (section \ref{sec:lm}) due to its ability to be trained on modeling sequences while also being able to capture linguistic information. As recoding, the mechanism proposed in this thesis, aims to be unsupervised, we also explored two different approaches to approximate the true weight posterior distribution by using Dropout (section \ref{sec:confidence-dropout}) and Bayesian anchored ensembles (section \ref{sec:confidence-ensembles}), both of which also enable the user to approximate model confidence in the form of predictive entropy. In the following chapter, I combine all these ideas into a single framework, called the recoder.

\chapter{Recoding Framework}\label{chapter:recoding-framework}

Having discussed related work and the necessary background for this thesis, I will use this chapter to layout the form and details of the proposed framework. The main idea stems from the work of \cite{giulianelli2018under}, where activation corrections were named \emph{interventions}, which is now being explored in more detail.

\section{Interventions}\label{sec:interventions}

In order to show how recoding is a generalization of \emph{interventions}, let us first discuss the latter.
 \cite{giulianelli2018under} examine the ability of \glspl{LSTM} to track subject-verb agreement, i.e.\ whether the models recognize the right verb number corresponding to the subject. Consider the sentence \emph{She only talks goobledygook} and its ungrammatical counterpart $^*$\emph{She only talk goobledygook}. In the second case, the grammatical violation manifests in the differing number between subject and verb, i.e.\ \emph{she} requiring the 3rd person singular form \emph{talks} when \emph{talk} is used instead. If \glspl{LSTM} store this information, then we would expect a Language Model to consistently assign a higher probability to the grammatical sentence than to the one in violation to the English grammar.\\

That this is indeed the case was already shown in the work of \cite{gulordava2018colorless}, however, \cite{giulianelli2018under} extend these results further:
They use Diagnostic classifiers \citep{hupkes2018visualisation, dalvi2019one, conneau2018you} (\glspl{DC}), linear classifiers that are trained on the activations of the models in order to predict the subject's number. If the classifier reaches a high accuracy, it implies that the number information is reliably encoded in the activations and can thus be used for a successful prediction. Not only do they show that this information is indeed tracked, but also that it can be corrupted by \emph{attractors}, intermediate nouns of a different number.\\

Their key contribution to this work now lies in trying to rectify these deviations by the use of \emph{interventions}: Using the binary cross-entropy loss of the \gls{DC} compared to the true subject number allows them to compute the the gradient of this loss w.r.t.\ to the \textbf{activations}. After performing an update step akin to the gradient descent step they are able to conclude that this procedure indeed restores the damaged number information.\\

This idea will now be extended and generalized into a framework called \emph{recoding}. Notwithstanding the similarities, I want to emphasize some key differences of the interventions in \cite{giulianelli2018under} compared to the proposed framework:

\begin{enumerate}
  \item \textbf{Interventions were only performed during inference} Intervention gradients were applied to an already trained \gls{LSTM}, while in this work recoding will already be applied during training.
  \item \textbf{Interventions were performed with a secondary objective in mind} While the model used in \cite{giulianelli2018under} was also trained to model language, the main focus of the work
  was to examine the ability of the model to retain information of the subject's number. Interventions were helpful here to avoid the corruption of this information while processing a sequence without changing the models perplexity significantly. In contrast, this work aims to utilize recoding directly to improve the main performance metric of the task.
\end{enumerate}

Therefore, when using additional labels to compute the error signal, it has to be considered to which extent the information contained in this labels is actually useful for the model to optimize
the primary objective. Furthermore, useful secondary labels can be hard to come by or have to be created manually, which is undesirable for the generalizational ability of the recoding approach.
The next sections thus focus on a more general concept.

\section{Recoding Mechanism}
Let us call the corrective mechanism that is added to the model the \emph{recoding mechanism} or \emph{recoder}. In its most abstract form, it simply consists of another computational step regarding the hidden activations of a model, which does not include any additional parameters. Let us define the mechanism as a function $r: \mathbb{R}^{N} \mapsto \mathbb{R}^N$ that maps a real-valued vector of activations to another vector of the same dimensionality.\footnote{While it could also be possible to imagine a version that maps multiple activations to a single new one, we will only consider this version in this work for now.} In the following, we will focus on the case of a \gls{RNN} that produces times-step dependent hidden activations $\bh_t \in \mathbb{R}^N$. Thus, we can define the recoding function as a modification of the hidden state using the Delta-rule \citep{widrow1960adaptive}:

\begin{equation}\label{eq:recoding-update}
  r(\bh_t) \equiv \bh_t^\prime = \bh_t - \alpha_t\nabla_{\bh_t}\delta_t
\end{equation}

where $\bh_t^\prime$ denotes the modified hidden state that will subsequently be used by the network, $\alpha$ the recoding \emph{step size} and $\delta_t$ a time-dependent \emph{error signal}. Naturally, as the anti-\emph{recoding gradient} $-\nabla_{\bh_t}\delta_t$ points into the direction from $\bh_t$ that minimizes the error signal $\delta_t$, taking a step $\alpha_t$ into its direction will result in a decrease or at least stagnation of the signal (see appendix \ref{appendix:theoretical-guarantees} for a full proof). Not specifying the kind of error signal enables us to consider vastly different methods, as we will see later.
The step size can be chosen to be constant, learned as an additional parameter or even parameterized using a whole different network (more about this in section \ref{sec:step-size}).
It is no coincidence that this modification mirrors the gradient descent update rule. Using this particular form allows us to derive certain theoretical guarantees about its efficacy while leaving room for extensions (similar to how the gradient descent update rule was modified using e.g.\ momentum \citep{polyak1964some} and can be used with any arbitrary objective function).

\begin{figure}[h]
  \centering
  \includegraphics[width=0.75\textwidth]{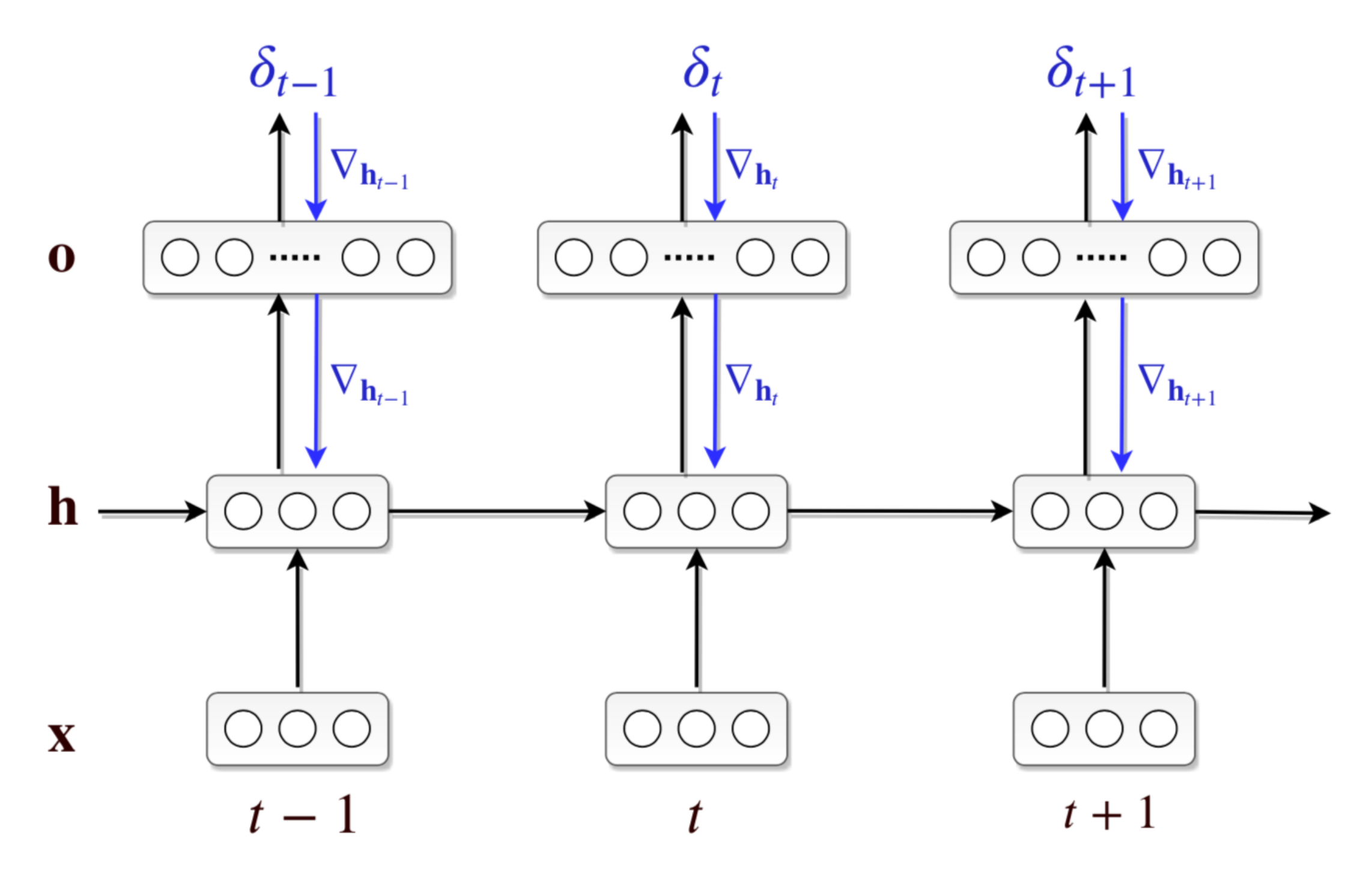}
  \caption[Illustration of the recoding framework]{Illustration of the recoding framework on a unidirectional, single-layer \gls{RNN}. The error signal $\delta_t$ is computed based on the hidden activations $\bh_t$, or, in this case, the output activations $\bo_t$ of every time step. Subsequently, it is used to compute the recoding gradient $\nabla_{\bh_t}\delta_t$ (blue) which is used in the update step to produce the recoded activations $\bh^\prime_t$ (\emph{recoding}), which are the new activations utilized in the next \gls{RNN} step.}\label{fig:recoding}
\end{figure}

The whole procedure is illustrated in figure \ref{fig:recoding} for a normal, unidirectional single-layer \gls{RNN}. However, the whole procedure can easily be extended to any \gls{RNN} variant, either uni- or bidirectional, with an arbitrary amount of layers, albeit recoding gradients will become increasingly smaller for lower layers the more
distant they are from the error signal $\delta_t$, as the chain of partial derivatives necessary to evaluate the gradient grows longer.

The next sections are dedicated to exploring different choices of error signals (\S\ref{sec:error-signals}) and methods to determine the recoding step size (\S\ref{sec:step-size}). Some additional information can be found in the appendix, e.g.\ theoretical guarantees for this framework in appendix \ref{appendix:theoretical-guarantees} and practical considerations during implementation in appendix \ref{appendix:practical-consid}.

\section{Error Signals}\label{sec:error-signals}

In this section, different sources for the error signal $\delta_t$ in a Language Modeling setting are explored. This ranges from using additional labels (\ref{subsec:weakly-supervised}),
to the ``surprisal'' of a model encountering a new token (\ref{subsec:surprisal}) or approximations of the predictive entropy of a model in sections \ref{subsec:mc-dropout} and \ref{subsec:anchored-ensembling}.

\subsection{Weak Supervision}\label{subsec:weakly-supervised}

Using this new framework, we can easily transfer the interventions from \cite{giulianelli2018under} into a recoding setting. Additional binary labels $\langle y\rangle_1^T = \langle y_1, \ldots, y_T \rangle$ with $y_t \in \{0, 1\}$ are used in combination with logistic regression classifiers to perform recoding. These classifiers simply predict the probability of one class by using an affine transformation of the hidden state. A loss function like binary cross-entropy loss can then be used to compute the difference between the predicted label and actual label:

\begin{equation}
    \delta_t \equiv \mathcal{L}_{\text{BCE}}(\sigma(\mathbf{w}^T\bh_t + \bb), y_t)
\end{equation}

where $\sigma$ denotes the sigmoid function and $\mathbf{w}, \bb$ pre-trained weights and bias. These parameters have been trained based on the hidden activations of the model itself. This could further be generalized by using labels belonging to an arbitrary number of classes and using more complex classifiers than logistic regression. Because this still carries the problem of having to use additional labels, this approach is not regarded further in this thesis.

\subsection{Surprisal}\label{subsec:surprisal}

\begin{figure}[h]
  \captionsetup[subfigure]{justification=centering}
  \centering
  \begin{subfigure}[b]{0.3\columnwidth}
    \centering
    \includegraphics[width=0.95\columnwidth]{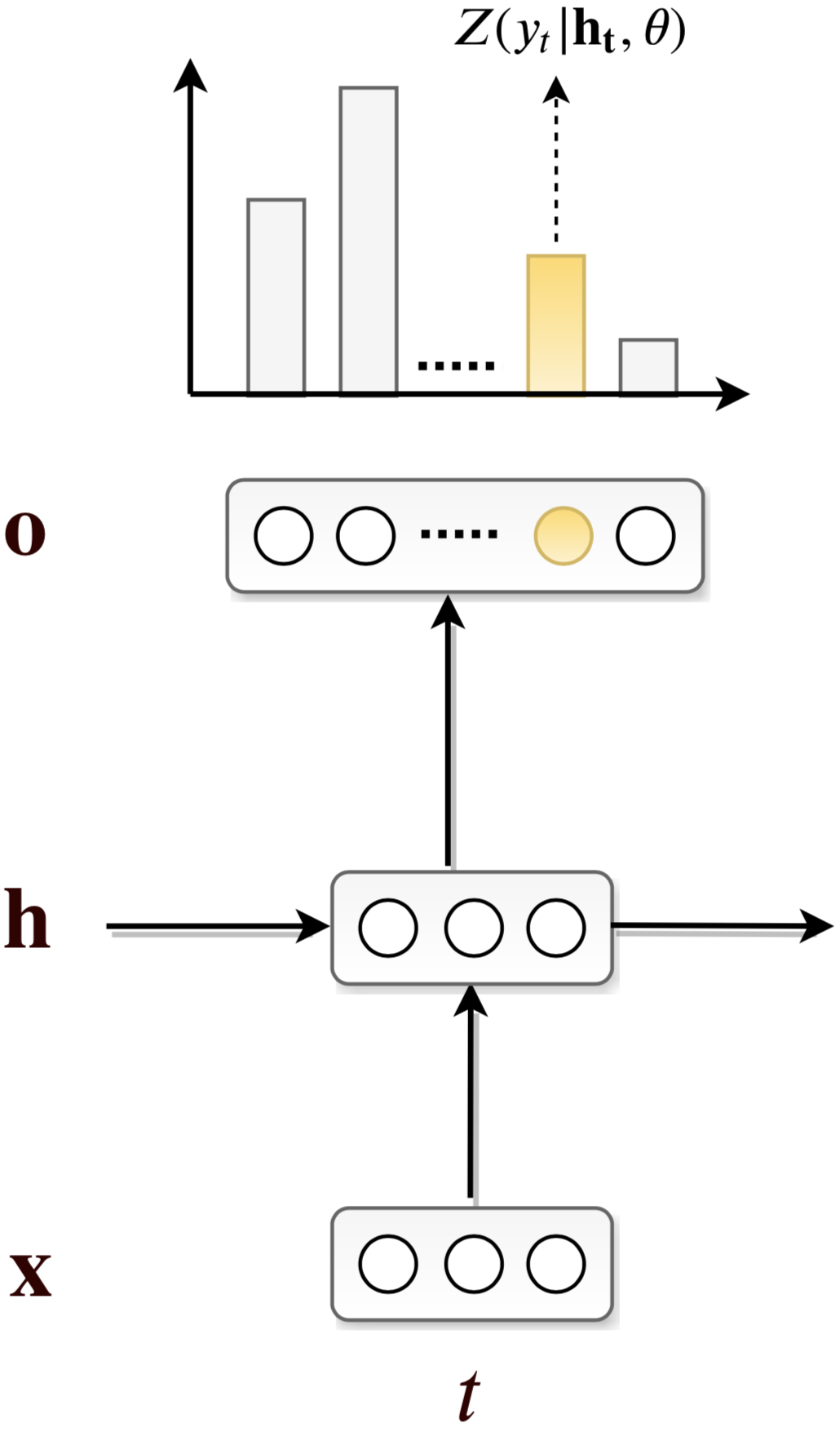}
    \caption{Surprisal}\label{fig:error-signals-surprisal}
  \end{subfigure}\hfill%
  \begin{subfigure}[b]{0.3\columnwidth}
    \centering
    \includegraphics[width=0.95\columnwidth]{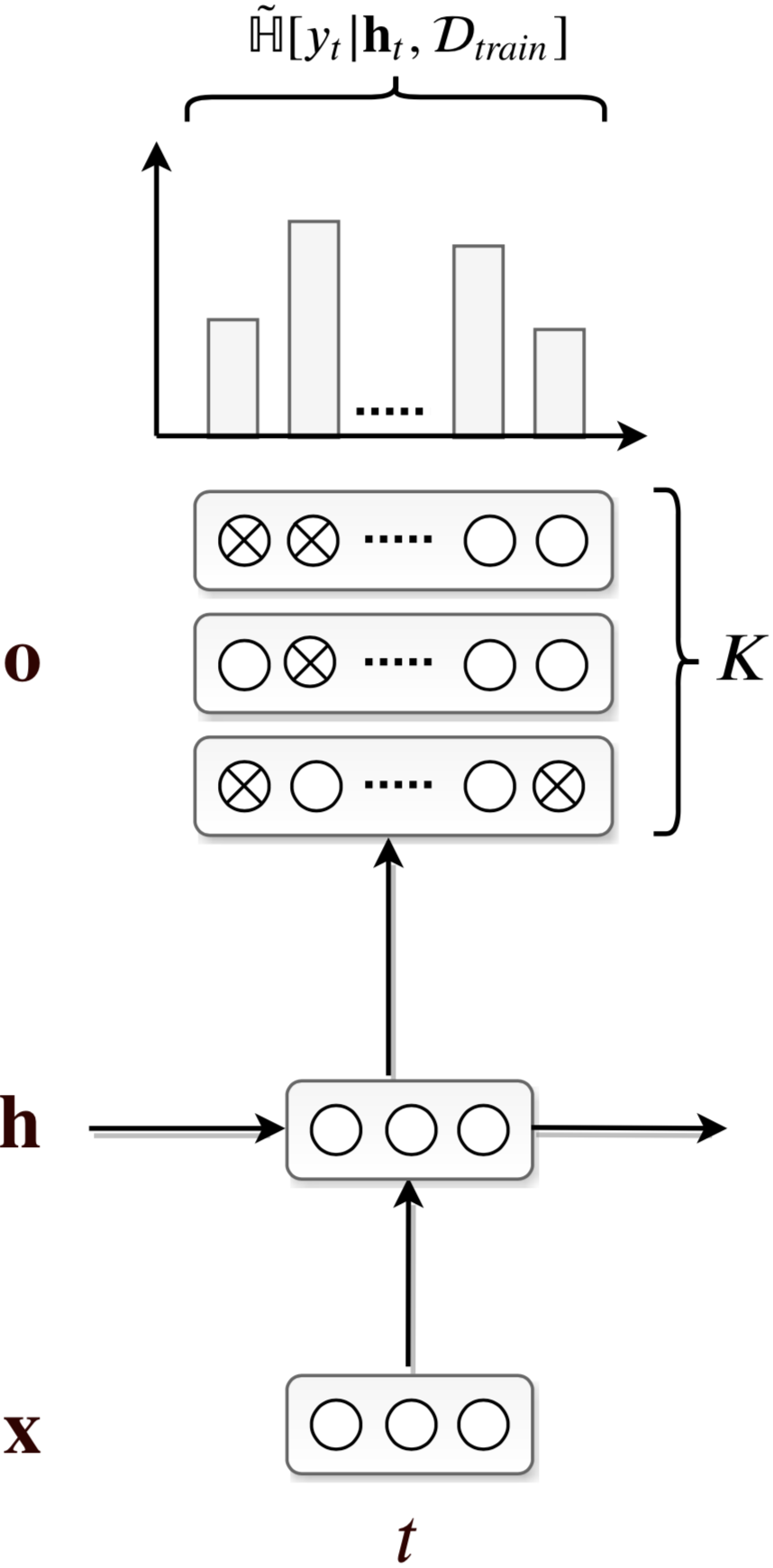}
    \caption{MC Dropout}\label{fig:error-signals-mcd}
  \end{subfigure}\hfill%
  \begin{subfigure}[b]{0.3\columnwidth}
    \centering
    \includegraphics[width=0.95\columnwidth]{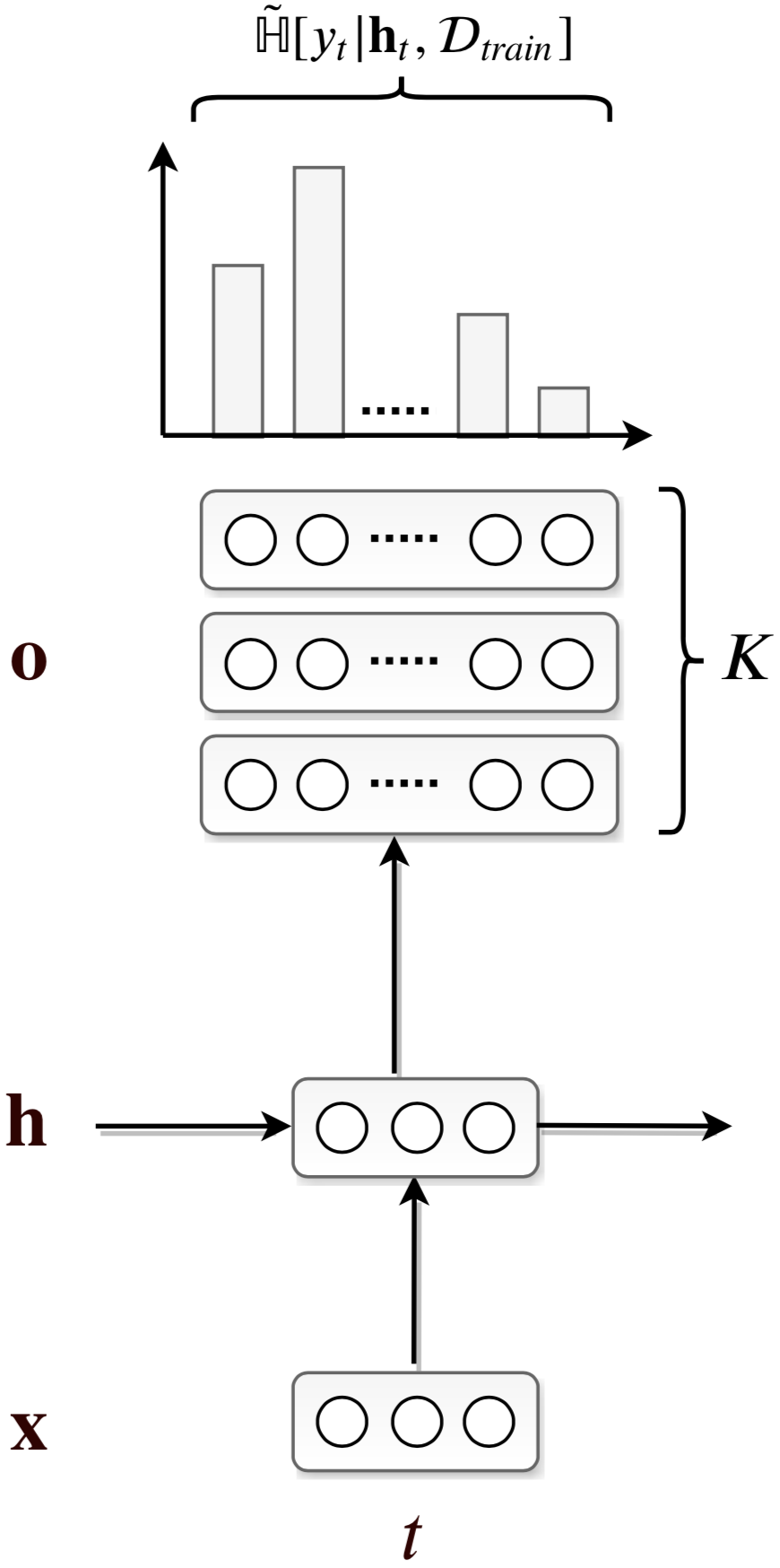}
    \caption{Anchored Ensemble}\label{fig:error-signals-bae}
  \end{subfigure}
  \caption[Illustration of different error signals.]{Illustration of the different error signals used in this work. (a) After generating the output distribution, the probability of the gold token (actual next word) is selected to compute the surprisal score. (b) $K$ forward passes with different Dropout masks are averaged, after which the predictive entropy of the resulting distribution is computed. (c) Like (b), but $K$ distributions stem from several ensemble members.}\label{fig:error-signals-illustrations}
\end{figure}

One of the easiest error signals to define in the context of language modelling would be the perplexity of the target token at a given time step, which we will call \emph{surprisal} in order to distringuish it from our evaluation metric. In general, the surprisal for a sequence of events $\langle x_1, \ldots, x_N \rangle$ is defined in terms of the entropy of their propability distribution $p(x)$:

\begin{equation}
  \text{ppl}(x_1^N) = 2^{-\frac{1}{N}\sum_{x_i} p(x_i)\log_2 p(x_i)}
\end{equation}

If we now exclusively consider the probability of the gold token $p(w_{t+1}|w_1^t)$,
this expression can be simplified. Let us denote the surprisal of a model as $\text{Z}(y_t|\bh_t, \theta)$, were $y_t = w_{t+1}$ in the language modelling case:

\begin{equation}\label{eq:surprisal-error-signal}
  \delta_t \equiv \text{Z}(y_t|\bh_t, \theta) = \sum_{c}\indicator{y_t = c}2^{-\bo_{tc}\log_2(\bo_{tc})} - 1 = \sum_{c}\indicator{y_t = c}\bo_{tc}^{-\bo_{tc}} - 1 \wnl
\end{equation}

where we simply add $-1$ so that the minimum surprisal value is $0$. This does not influence the expression of the recoding gradient, which can be derived to be

\begin{equation}
    \nabla_{\bh_t}\text{Z}(y_t|\bx_t, \btheta) = -\mathbf{A}\Big(\log(\bo_t) + \mathbf{1}\Big)\exp\Big(-\mathbf{A}\bo_{t}^T\log(\bo_{t})\Big)\Big(\text{diag}(\bo_{t}) - \bo_{t}\bo_{t}^T\Big)\bW_{ho}
\end{equation}

The full derivation for this is given in appendix \ref{appendix-surprisal-recoding-grad}. Intuitively, this error signal moves the hidden activations into a direction where the resulting surprisal score is being lowered, in other words the prediction based on these activations $\bh_t^\prime$ would assign a higher probability to the gold token than the one based on the original $\bh_t$.\\

Equipped with these results, we can apply surprisal recoding in the way described in fig. \ref{fig:code-surp-recoding} in appendix \ref{appendix:pseudocode} and illustrated in fig. \ref{fig:error-signals-surprisal}:
For every time step, we use the probability of the gold token to compute the surprisal score, which in turn is used to compute its gradient w.r.t.\ to the hidden state $\bh_t$. After the update step, the network continues with the recoded hidden activations. Two things should be noted here: Firstly, this approach is still supervised, but only uses the gold labels already included in the training set. Secondly, the output distribution $\bo_t$ cannot be recomputed based on the new recoded hidden activations $\bh_t^\prime$ as this would trivialize the task.\footnote{The recoding gradient induces knowledge about the gold token into the model. Therefore recomputing the output predictions would give the model an unfair advantage, as it already possesses knowledge about its target.}
However, in theorem \ref{theorem:error-reduction-tt} we have proven that recoding also has a positive influence on the value of future error signals.

\subsection{MC Dropout}\label{subsec:mc-dropout}

Another approach to determine an error signal lies in using the model's predictive entropy. Here, MC Dropout \citep{gal2016dropout} and Variational RNN Dropout \citep{gal2016theoretically}, introduced in section \ref{sec:bayesian-deep-learning}, provide a procedure to estimate this measure by performing several forward-passes of the model using different Dropout masks. We apply these techniques in two distinct ways:

\begin{enumerate}
  \item Apply \cite{gal2016dropout} by making the simplifying assumption that the decoder weights $\bW_{ho}$ can be seen as a single-layer \gls{MLP}. For every time step $t$, sample $K$ Dropout masks to estimate the predictive entropy and use that in order to recode the hidden activations $\bh_t$.
  \item Apply \cite{gal2016theoretically} by sampling $K$ Dropout masks at the beginning of each batch corresponding to every type of connection in the RNN. For every time step $t$, estimate the predictive entropy this way and use that in order to recode all hidden activations $\bh_t^{(k)}$.
\end{enumerate}

First, we focus on the simplified approach: Here, we apply the standard Dropout procedure by using a Dropout mask sampled from a Bernoulli distribution with probability $p$ s.t.

\begin{equation}
    \bZ_{ij}^{(k)} \sim \text{Bernoulli}(p)\ \forall i = 1, \ldots, N\ \ \forall j= 1, \ldots, |\mathcal{V}|
\end{equation}
\begin{equation}
    \bW_{ho}^{(k)} = \bW_{ho}\odot\bZ^{(k)}
\end{equation}

In this case, the set of weights $\bomega^{(k)}$ using for a stochastic forward pass is simply $\bomega^{(k)} = \{\bW_{ho}^{(k)}\}$ and the probability of a class given those weights $p(y_t = c|\bx_t, \bomega^{(k)}) = \bo_{tc}^{(k)}$.
Using eqs. \ref{eq:pred-entropy} and \ref{eq:pred-entropy-approx}, we can define the error signal in terms of the approximate predictive entropy, now conditioned on the hidden state $\bh_t$ due to the recurrent setting:

\begin{equation}
  \delta_t \equiv \tilde{\mathbb{H}}[y_t|\bh_t, \mathcal{D}_{\text{train}}] = - {\displaystyle\sum_{c=1}^{|\mathcal{V}|}} \bohat_{tc}\log(\bohat_{tc}) = -\bohat_t^T\log(\bohat_t)
\end{equation}

where

\begin{equation}
  \bohat_t = \frac{1}{K}{\displaystyle\sum_{k=1}^K} \bo_t^{(k)} \wnl
\end{equation}

The corresponding recoding gradient can be identified as

\begin{equation}
  \nabla_{\bh_t} \tilde{\mathbb{H}}[y_t|\bh_t, \mathcal{D}_{\text{train}}] = - \Big(\log(\bohat_t) + \mathbf{1}\Big)\frac{1}{K}{\displaystyle \sum_{k=1}^K}(\text{diag}(\bo_t^{(k)}) - \bo_t^{(k)}(\bo_t^{(k)})^T)\bW_{ho}^{(k)}
\end{equation}

The full derivation is given in appendix \ref{appendix-pred-entropy-recoding-grad}. On an intuitive level, this means that the recoding gradient based on this error signal moves the hidden activations in such a direction such that the \emph{confidence} of the model's prediction based on these activations is increased. However, because the predictive entropy is approximated using Monte Carlo methods, the quality of the gradient depends on the number of samples $K$ used.
The procedure is illustrated in fig. \ref{fig:error-signals-mcd} and written in pseudocode in fig. \ref{fig:code-mcd-recoding} in appendix \ref{appendix:pseudocode}.
Note that because this error signal is being computed in an entirely supervised manner, we are allowed to recompute the output distribution $\bo_t$ here.

The approach using the the variational \gls{LSTM} Dropout is also tested: Here we can approximate the predictive entropy by sampling a different Dropout masks for every type of connection as illustrated in figure \ref{fig:variational-rnn}. To do so, we have to calculate the network's predictions under $K$ different sets of Dropout masks. We can also imagine this as an ensemble of $K$ \glspl{RNN}, each equipped with a different set. Practically, although this results in less Dropout masks to sample in total\footnote{$K \times T$ masks in the simplified version vs. $3 \times K + (2 \times L \times K)$ for an \gls{LSTM} with a number of layers $L$, therefore less masks are sampled when $3 + 2 \times L < T$.}, it is most efficient to keep track of all intermediate hidden, cell and output activations by transforming batches into pseudo-batches of with $K \times B$ instances, where we save each batch instance with a different Dropout mask. This effectively slows down the model training again, but also provides more theoretically grounded estimates.
The derived recoding gradient does not change, except that $\bh_t$ is now a concatenation of all hidden activations produced by the network under different sets of Dropout masks s.t. $\bh_t = [\bh_t^{(1)} \oplus \ldots \oplus \bh_t^{(K)}]$.

\subsection{Bayesian Anchored Ensembling}\label{subsec:anchored-ensembling}

Bayesian Anchored Ensembling \citep{pearce2018uncertainty} (\gls{BAE}), as introduced in \ref{sec:bayesian-deep-learning}, allows for another way to estimate predictive uncertainty that does not require the costly sampling of Dropout masks. In contrast, we
use an ensemble of $K$ decoders with decoder weights $\bW_{ho}^{(k)}$ to produce a set of different output activations, which is depicted in fig. \ref{fig:error-signals-bae}. In the context of the recoding framework, this approach comes with some caveats: Firstly, the theoretical foundations of this approach have not been extended to RNNs yet and are only valid for \glspl{MLP}. Straightforwardly, one could just ensemble multiple \gls{RNN} cells by sampling anchor points for every set of weights in the cell. However, ensembling $K$ full \glspl{RNN} this way increases training time tremendously. We will therefore employ
the same simplification from the previous subsection by only applying this technique to the decoder, although it is debatable whether certain assumptions still hold under this circumstances: \cite{pearce2018uncertainty} claim that Bayesian anchored ensembling gives good results for sufficiently wide NNs, as these are known to display an increasing correlation between parameters over the course of a training. Nevertheless, we proceed with this simplification for now.

The derivation of the recoding gradient $\nabla_{\bh_t}\delta_t$ for anchored ensembles is equivalent to that of MC Dropout in section \ref{subsec:mc-dropout} and is therefore not repeated here. The main difference is that $\bomega^{(k)}$ now does not denote the weights of the output layer with a specific Dropout mask but the weights of the $k$-th member of the anchored ensemble and $\bo_t^{(k)}$ to the normalized output probabilities produced by these weights. A justification that this can also be used to approximate the predictive entropy is given in appendix \ref{appendix:pred-entropy}. Notwithstanding the formal resemblance to the gradient derived for MC Dropout recoding, the actual gradients
differ in practice, as $\bomega^{(k)}$ refers to the same set of weights with different Dropout masks in the case
of MC Dropout, but to a set of independently initialized weights in this circumstance.\\

Again, the resulting recoding gradient based on this error signal points into a direction such that the \emph{confidence} of the model's prediction based on these activations is increased. In similar fashion to the MC Dropout error signal, the quality of the gradient is expected to depend on the number of members in the ensemble $K$.\\

The full pseudocode for this procedure is given in fig. \ref{fig:code-bae-recoding} in \ref{appendix:pseudocode}. As calculating the additional weight decay loss term $\mathcal{L}_{\text{anchor}}^{(k)}$ and learning all different $K$ members of the ensemble separately slows down training, I employ a variant of this approach where we only sample a single anchor point for all ensemble members and average their losses to one single term:

\begin{equation}
    \mathcal{L}_{\text{total}} = \frac{1}{K}\sum_{k=1}^K \big( \mathcal{L}^{(k)}_\text{CE} + \mathcal{L}_{\text{anchor}}^{(k)}\big)
\end{equation}

This armortized form of Bayesian Anchored Ensembling actually proved to train significantly faster and produce better results in preliminary experiments.

\section{Step Size}\label{sec:step-size}

Lastly, we are going to focus on different choices of the step size $\alpha$. The theorems in appendix \ref{appendix:theoretical-guarantees} rely on the optimal step size which can be derived from the smallest Lipschitz constant $L$. In practice however, this constant is hard to obtain (this will be discussed in more detail in section \ref{sec:future-step-size}). I therefore experiment with three different alternatives:\\

\begin{enumerate}
  \item \textbf{Constant step size} The simplest approach just treats the step size as a hyperparameter and uses the same value $\alpha$ across different time steps, layers and activation types.
  \item \textbf{Learned step size} The next approach defines the step size as an additional parameter, which is learned via gradient descent alongside the model's other parameters. This has two advantages: We do not have to perform any hyperparameter search for the step size and we can add multiple step size parameters for every layer and activation type.
  \item \textbf{Predicted step size} Here we parameterize the step size using an additional network with parameters $\bkappa$ s.t. $\alpha_t = f_\bkappa(\bh_t)$. This way, step sizes can flexibly be determined based on the content of the sentence encoded in $\bh_t$. Furthermore, we can also use of these predictors for every activation type and layer, using the corresponding layer's hidden activations as input. These predictors can also be learned using gradient descent, but thus add additional training time.
\end{enumerate}

\section{Recap}\label{sec:recoding-recap}

In this chapter I presented an extension and generalization of the interventions in \cite{giulianelli2018under} that performs gradient-based updates of activations, called \emph{recoding}. I furthermore presented three error signals that will be examined in the next chapter: Surprisal, which based recoding gradients on the probability of the gold token and can therefore be classified as a supervised recoding approach, as well as MC Dropout and Bayesian Anchored Ensembling recoding. The latter two function in an entirely unsupervised manner and approximate the predictive entropy of the model at every processing time step. I also introduced three distinct ways of determined the step size of the recoding update step. All of these variants will now be tested extensively.

\chapter{Experiments}\label{chapter:experiments}

In this chapter I present the results of several experiments applying the models described in the previous chapter to a Language Modeling task, assessing the impact of important hyperparameters and model components.\\

The data set is being introduced first in section \ref{sec:exp-dataset} along with the training conditions in section \ref{sec:exp-training}. Secondly, I evaluate the different models with a language modeling objective in section \ref{sec:exp-lm}. More precisely,
the effect of the most prominent hyperparameters on performances is dissected in sections \ref{sec:exp-step-size}, \ref{sec:exp-num-samples} and \ref{sec:exp-dropout-rate}.
Additional experiments in section \ref{sec:learned-step-size} focus on the impact of the dynamic step sizes like introduced back in chapter \ref{sec:step-size}. Lastly, some ablation experiments try to explain the interplay between the \gls{LSTM} and the recoder in chapter \ref{sec:ablation}.\\

Implementing these models in a framework like \verb|PyTorch| come with some caveats, which are described in appendix \ref{appendix:practical-consid}.

\section{Dataset}\label{sec:exp-dataset}

The data set used for the experiments is the Penn Treebank (\gls{PTB}) \citep{marcus1993building}, consisting of documents in American English originating from different sources like e.g.\ radio transcripts, governmental documents and financial news reports. The corpus is annoted with linguistic information like the syntactic tree structure and Part-of-Speech tags, which is discarded for the purpose of this work. End-of-sentence tokens are also added. Notwithstanding its small size compared to modern Deep Learning data sets (about 4.5 million words in total), it is still a benchmark for state-of-the-art models in language modelling and was purposefully chosen, as its limited size allows for rapid prototyping of new models and allows full training even for slower models, which helps to ensure comparability.

\section{Training}\label{sec:exp-training}

Apart from the parameters of the model whose values are learning during the training, the procedure and model also comes equipped with a number of parameters that have to be determined beforehand, called \emph{hyperparameters}. In this instance, hyperparameters are e.g.\ the number of layers of a model, the choice of optimizer, the learning rate (schedule), and many more. Due to the sheer number of possible hyperparameters, models and model variations as well as computational resource constraints I conduct the hyperparameter search differently than performing the rather na\"ive grid search. Figure \ref{table:hyperparams} in the appendix gives a (non-exhaustive!) list of the main hyperparameters of the different models. For some models like recoding using MC Dropout, this results in $12$ possible hyperparameters to tune. Just performing grid-search with $4$ options per parameter would result in $12^4 = 20736$ runs to evaluate, which is intractable under the given circumstances.\\

For this reason, I employ two strategies: First, we reduce the search space by adapting the hyperparameters for the core \gls{LSTM} found by \cite{gulordava2018colorless}, which can be seen in table \ref{subfig:all-hyperparams}. While the authors do not necessarily have a strong motivation for their chosen options, they perform an extensive grid search. Secondly, sampling hyperparameters from a distribution instead of arranging them in a grid is more efficient, as not all parameters are equally important \citep{bergstra2012random}, which is why this strategy is preferred over the common grid search.
The details of this procedure are described in appendix \ref{sec:hyperparameter-search}. Even this approach will not produce the best hyperparameters, but it should be made clear that this is not the intention of this work.
Instead, the aim is to produce a \emph{good enough} set of hyperparameters that allows us to examine the effect of adjusting individual options, make comparisons to the baseline and probe the models for interesting behavior and properties. The final hyperparameters listed in table \ref{table:hyperparameter-best} in the appendix will stay fixed in the following unless explicitly mentioned otherwise.\\

With the final set of parameters, the models are trained on the corpus for $8$ epochs with a learning rate schedule corresponding to \cite{gulordava2018colorless}, where the value is halfed when no improvement has been observed for an entire epoch using mini-batch stochastic gradient descent and up to four NVIDIA 1080TI graphics cards at a time, each running one separate instance. Models are selected using early stopping based on their perplexity score on a validation set. As it is common practice for Language Modeling, the data set is partitioned into batches of a certain size and sequence length, where sentences do not necessarily end at the end of a batch but continue into the same instance of the next batch. More details about the evaluation is given in the next section.

\section{Evaluation}\label{sec:exp-lm}

The models are evaluated on the test set consisting of about $3600$ sentences based on their perplexity scores. For a sequence of $T$ words $\langle w\rangle_1^T = \langle w_1, \ldots, w_T\rangle$, the perplexity of a model is defined as

\begin{equation}
  \text{ppl}(w_1^T) = 2^{-\frac{1}{T}\sum_{t=1}^T \log_2 p(w_t)}
\end{equation}

In the following subsections, I explore the impact of different hyperparameters and model components on the performance of the models compared to the baseline. To account for variance between model runs due to their random initialization, the results are given as the average over four runs, including their standard deviation. A single-tailed Student's t-test is used to validate possible improvements over the baseline.

\subsection{Recoding Step Size}\label{sec:exp-step-size}

\begin{figure}[H]
  \centering
  \begin{tabular}{cc}
    \includegraphics[width=0.495\textwidth]{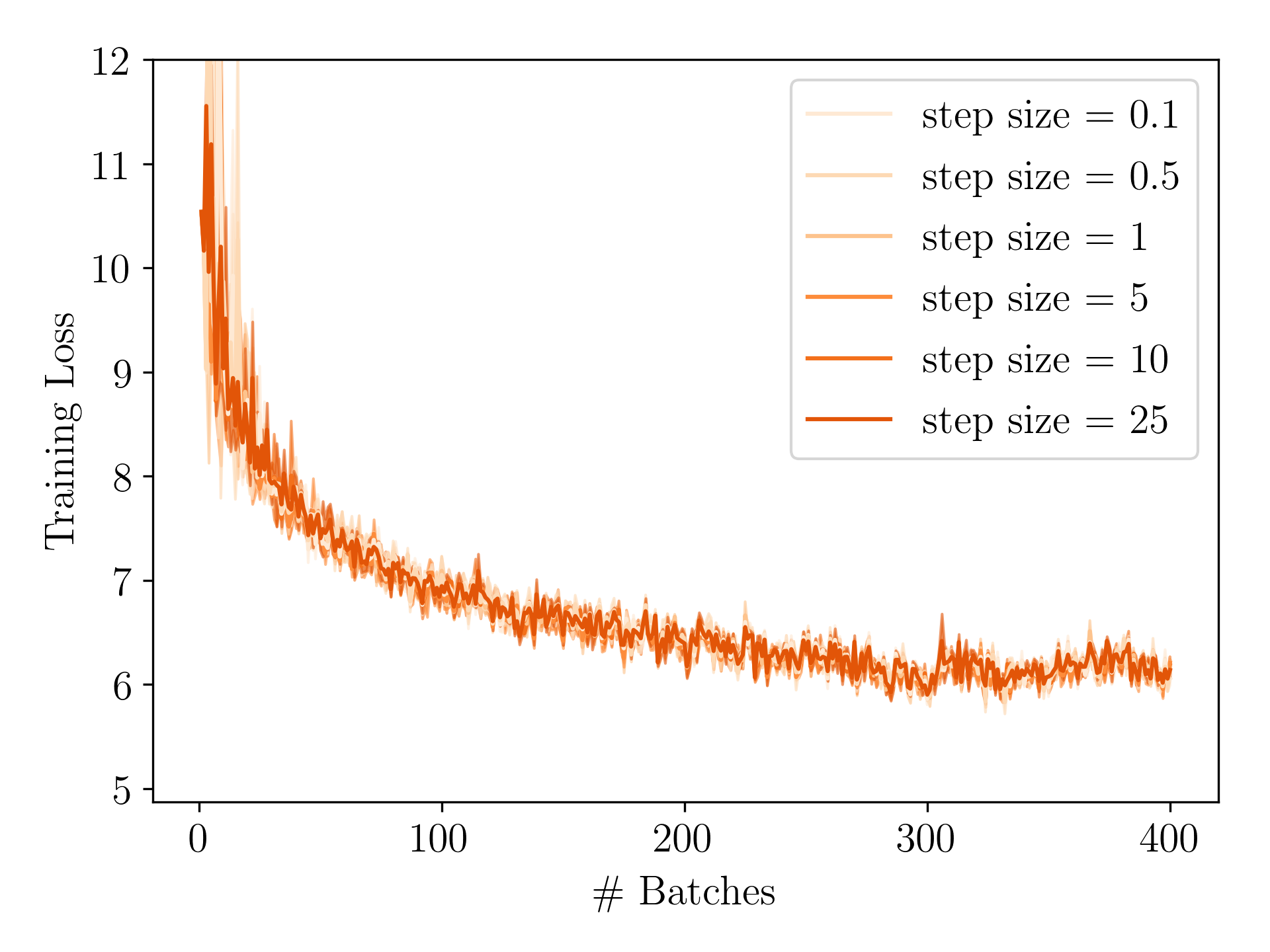} &
    \includegraphics[width=0.495\textwidth]{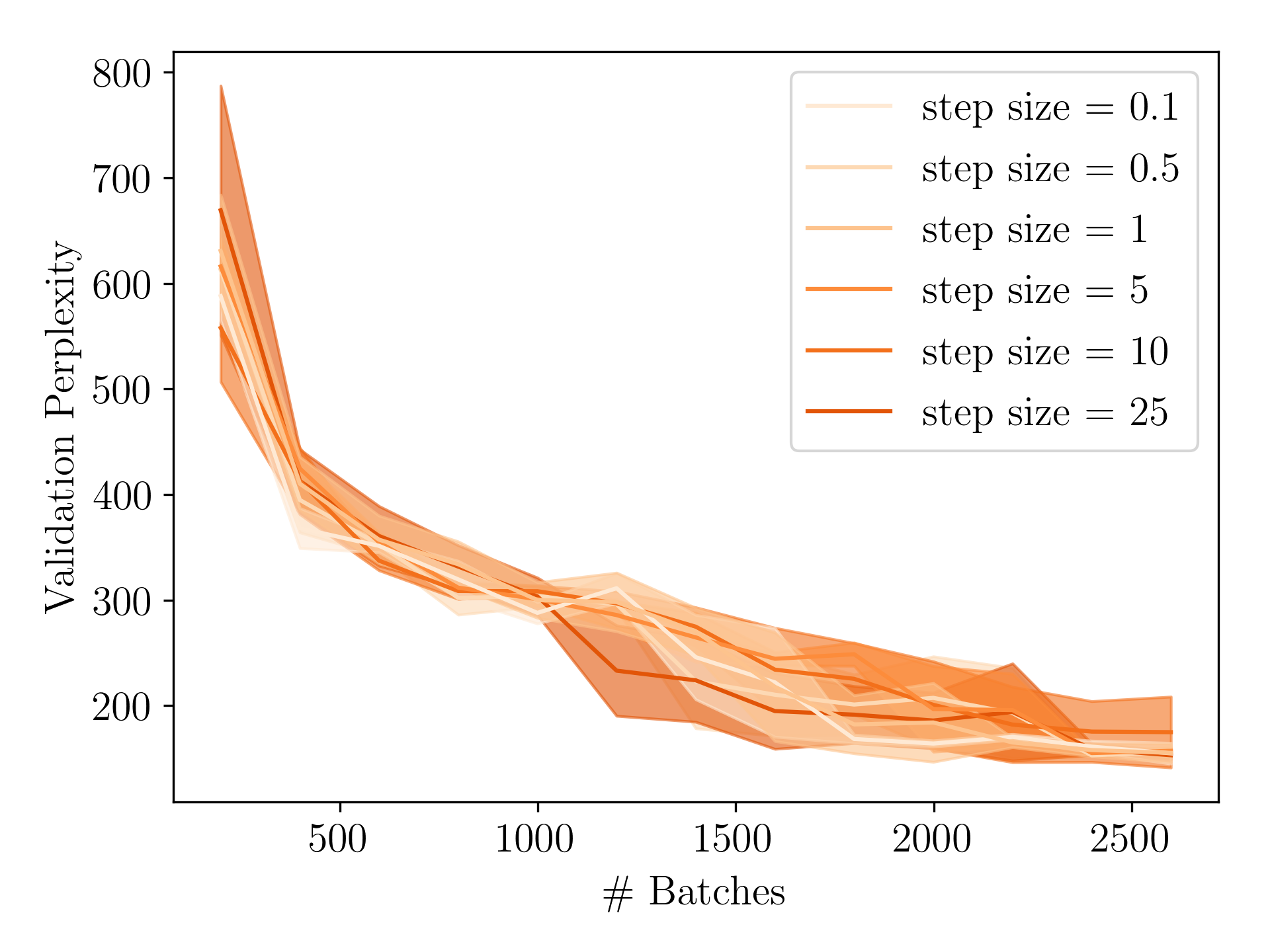} \\
    \includegraphics[width=0.495\textwidth]{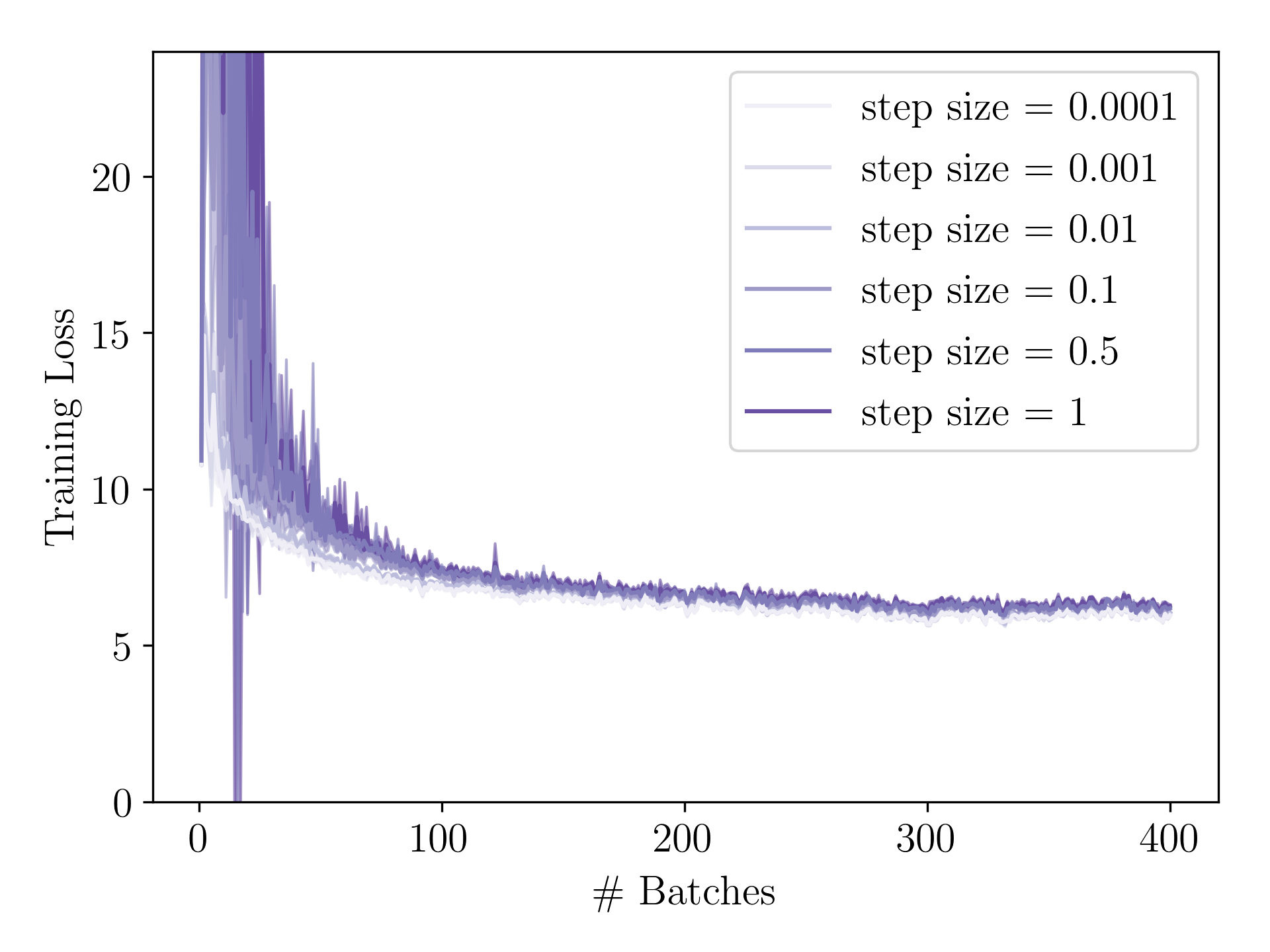} &
    \includegraphics[width=0.495\textwidth]{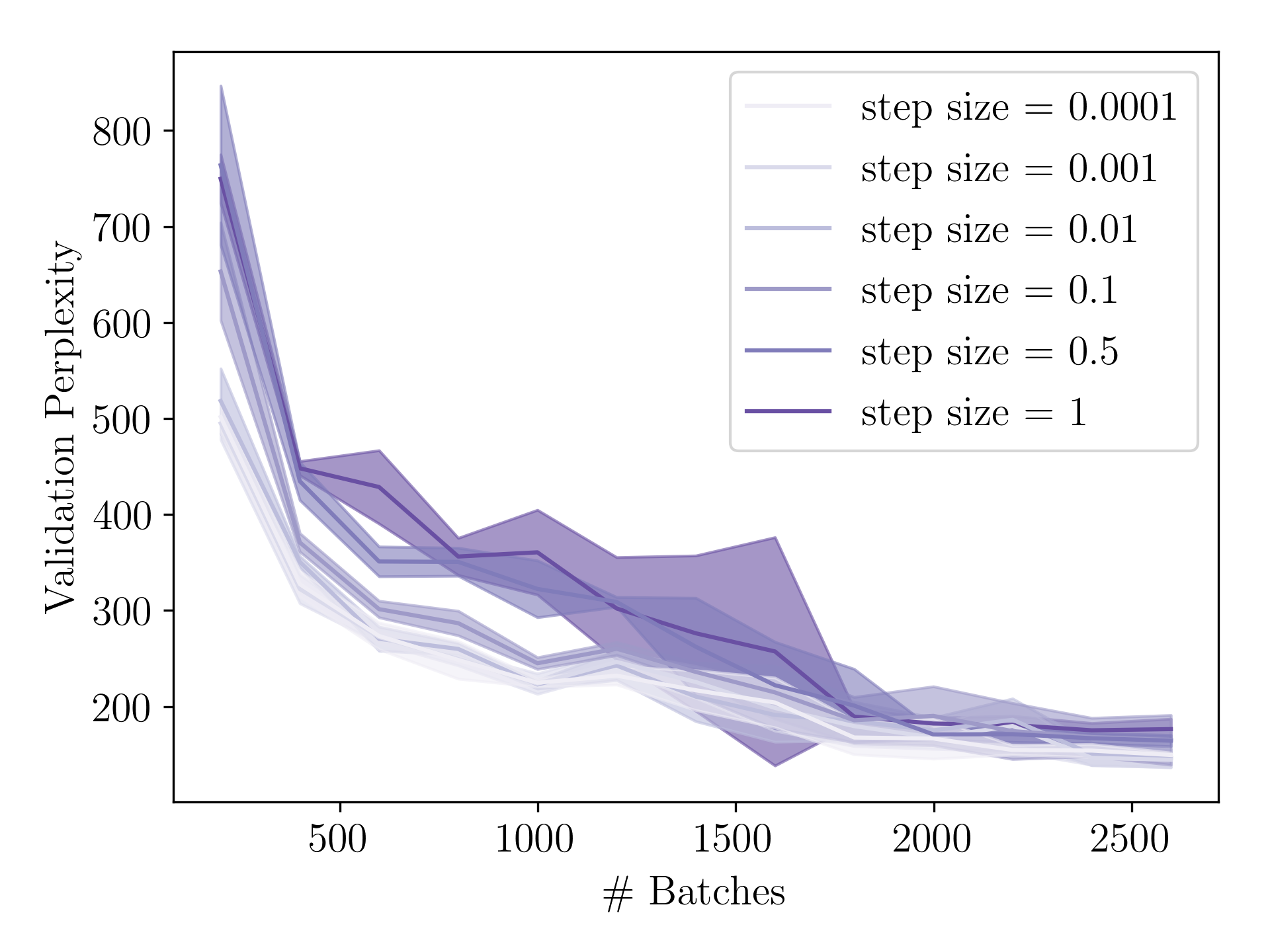} \\
    \includegraphics[width=0.495\textwidth]{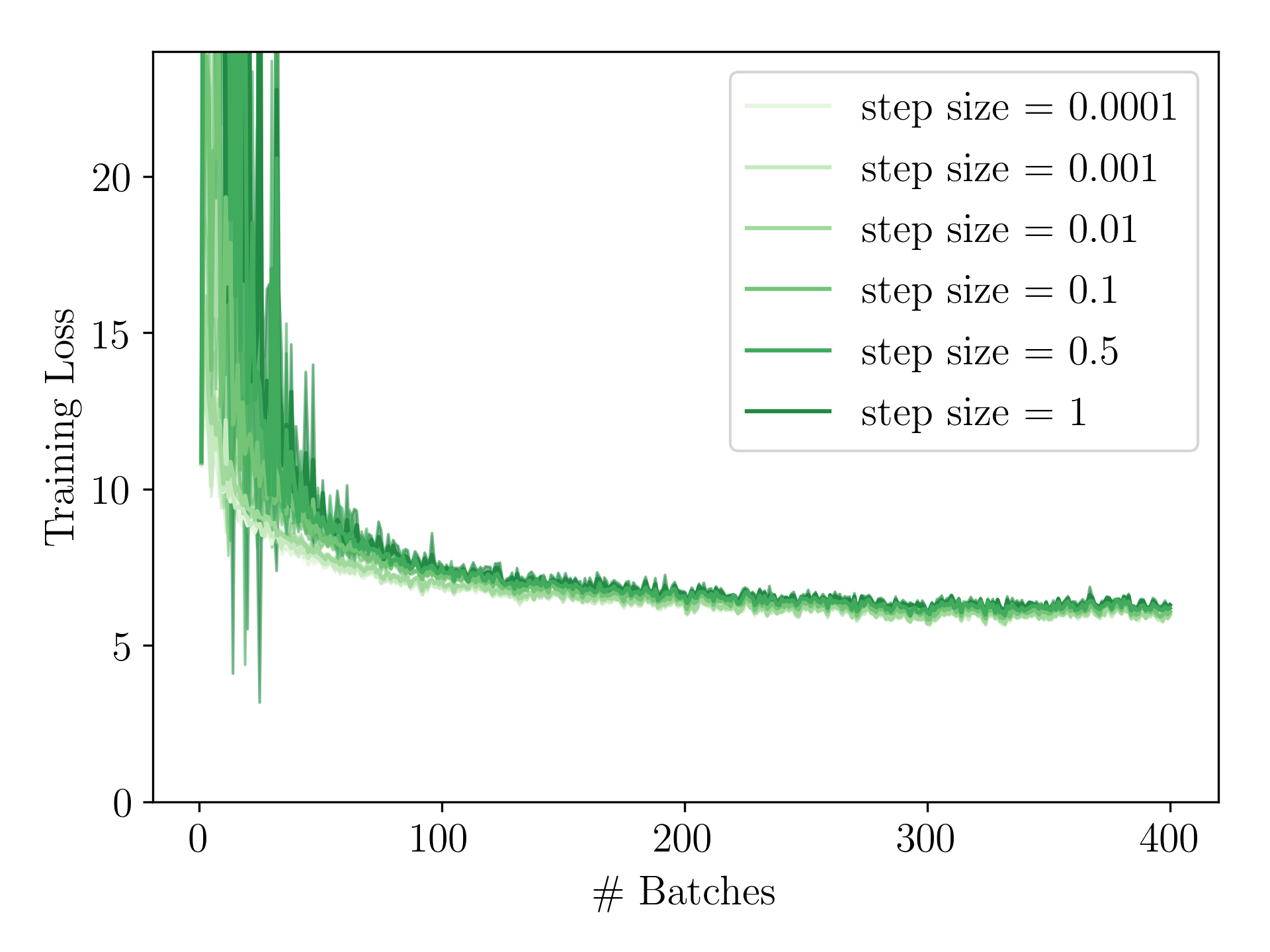} &
    \includegraphics[width=0.495\textwidth]{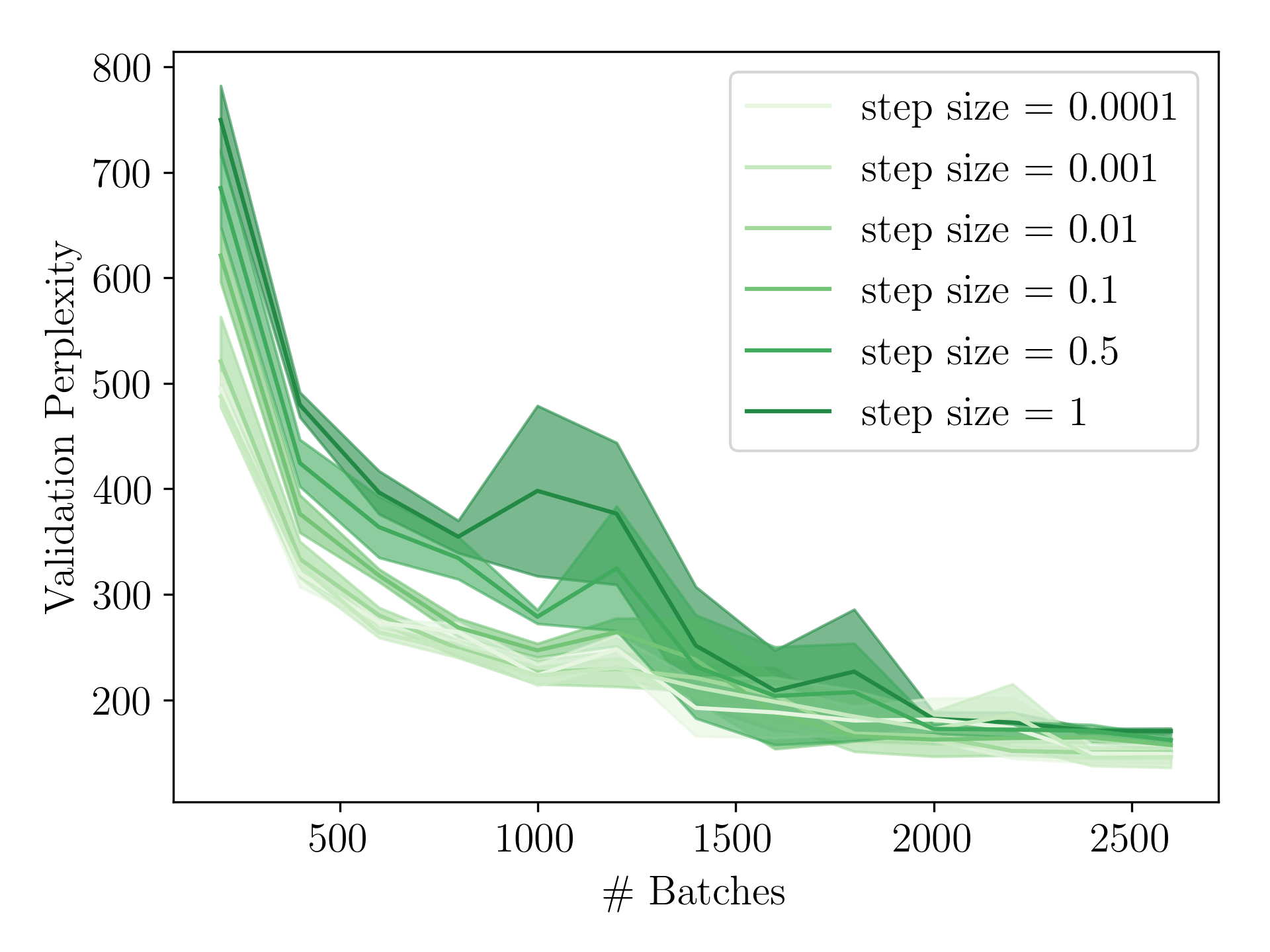} \\
  \end{tabular}
  \caption[Training losses for different recoding step sizes]{Training loss (left column) and validation perplexities (right column) on \gls{PTB} for different recoding step sizes using \surpcol{Surprisal}, \mcdcol{MC Dropout} and \baecol{Bayesian anchored ensemble} recoding. Curves denote the mean over $n=4$, with intervals signifying a standard deviation. Best viewed in color.}\label{fig:exp-step-sizes-curves}
\end{figure}

I begin by examining the impact of the most integral parameter, the recoding step size $\alpha$. At first, we just look at the effect of setting the step size to fixed values, the results obtained by learning or predicting the step sizes are given in a later section \ref{sec:learned-step-size}. Although the following experiments cover a wide range of step sizes (from $10^{-4}$ to $25$), not all of those are applicable to all models. For example, the surprisal given the target token often results in value below $1$. In contrast, the approximate predictive entropy reaches values in the hundreds, therefore not all of the step sizes that are experimented on are applicable to all models, which is why some of the results have been omitted in table \ref{fig:exp-step-sizes-curves}, where all scores are listed.
 It should also be mentioned that using variational \glspl{RNN} on the basis of \cite{gal2016theoretically} to compute the error signals produced results magnitudes worse than all other considered model variants in preliminary experiments and corresponding outcomes were therefore disregarded in this and later sections.\\

It seems immediately clear that there exists an advantageous range of values for the step size for all recoding types alike. Fig. \ref{table:exp-step-sizes-results} shows that for all recoding variants, more extreme values let models converge to worse training and validation scores. This is also reflected in the test perplexity scores in table \ref{table:exp-step-sizes-results}, which are obtained for a step size of $5$ for surprisal and $0.001$ for \gls{MCD} and \gls{BAE} recoding. In the case of the latter two, a lower step size than $0.001$ seems to result in a worse performance again.\\

\begin{table}[h]
  \centering
  \setlength{\tabcolsep}{8pt}
  \def\arraystretch{1.5}
  \resizebox{.85\textwidth}{!}{
  \begin{tabular}{@{}rccc@{\hspace{1cm}}c@{}}
       \toprule[1.5pt]
        & \multicolumn{3}{c}{Recoding Models} & \\ \cline{2-4}
       Step size & \surpcol{Surprisal} & \mcdcol{MC Dropout} & \baecol{BAE} & Baseline \\
       \toprule[1.15pt]
        $10^{-4}$ & -                   & $144.03 \pm 3.54$   & $143.66 \pm 8.79$ & $\mathbf{126.60 \pm 4.06}$ \\
        $10^{-3}$ & -                   & \mcdcol{$139.61 \pm 7.28$}   & \baecol{$139.40 \pm 7.24$} & \\
        $0.01$    & -                   & $142.43 \pm 8.12$   & $144.29 \pm 4.17$ & \\
        $0.1$     & $126.93 \pm 1.12$   & $157.91 \pm 22.03$  & $151.61 \pm 4.09$ & \\
        $1$       & $126.94 \pm 0.86$   & $169.89 \pm 7.36$   & $164.79 \pm 1.84$  & \\
        $5$       & \surpcol{$\mathbf{125.33 \pm 1.32}$}   & $198.95 \pm 7.66$   & $204.00 \pm 12.12$  & \\
        $10$      & $170.51 \pm 30.78$  & -                   & - &  \\
        $25$      & $235.14 \pm 76.99$  & -                   & - & \\
       \bottomrule[1.15pt]
  \end{tabular}}
  \vspace{0.2cm}
  \caption[Test perplexities for different recoding step sizes]{Test perplexity on the test set for different models and a variety of recoding step sizes. Results are averaged of $n=4$ runs. Best result overall is given in bold, best result per model in its corresponding color.}\label{table:exp-step-sizes-results}
\end{table}

However, none of the novel models presented manage to outperform the baseline in a statistically significant manner. This could be due to one or a combination of the following reasons: Firstly, as seen in this line of experiments, the choice of step size seems to play a crucial role in the performance of the new models. Using a fixed step size therefore seems to be unlikely to be the best choice of the model under all circumstances encountered during testing. If we assume that decreasing the error signal $\delta_t$ has a positive impact on the model performance, then the theorems in appendix \ref{appendix:theoretical-guarantees} suggest that the possible improvements depend on using a step size that depends on the nature of the error function that produces $\delta_t$. The ``error signal surface'', however, changes with every parameter update of the model, and therefore the step size should be adjusted as well, but the exact adjustment is hard to determine (for a more in-depth discussion for this see chapter \ref{chapter:discussion}).
Therefore, models using a more flexible step size are evaluated in section \ref{sec:learned-step-size}.\\

Another critical point to be considered is whether the chosen error signals actually provide the model with some advantage to solve the task. This is easier to argue for in the case of surprisal, because it is directly based on the probability of the target token. In the case of \gls{MCD} and \gls{BAE} recoding, i.e.\ when using predictive entropy, this is less clear. Here, recoding intuitively should improve the model's confidence encoded in its activations, but it is uncertain whether increasing the confidence actually helps to avoid mispredictions later on and whether error occur by making very confident wrong predictions. These points are also considered in a later part of this work, namely chapter \ref{sec:overconfident}. To further assess the impact of certain hyperparameters, the effect of the number of samples $K$ is evaluated next.

\subsection{Number of samples}\label{sec:exp-num-samples}

\begin{figure}[h]
  \centering
  \begin{tabular}{cc}
    \includegraphics[width=0.495\textwidth]{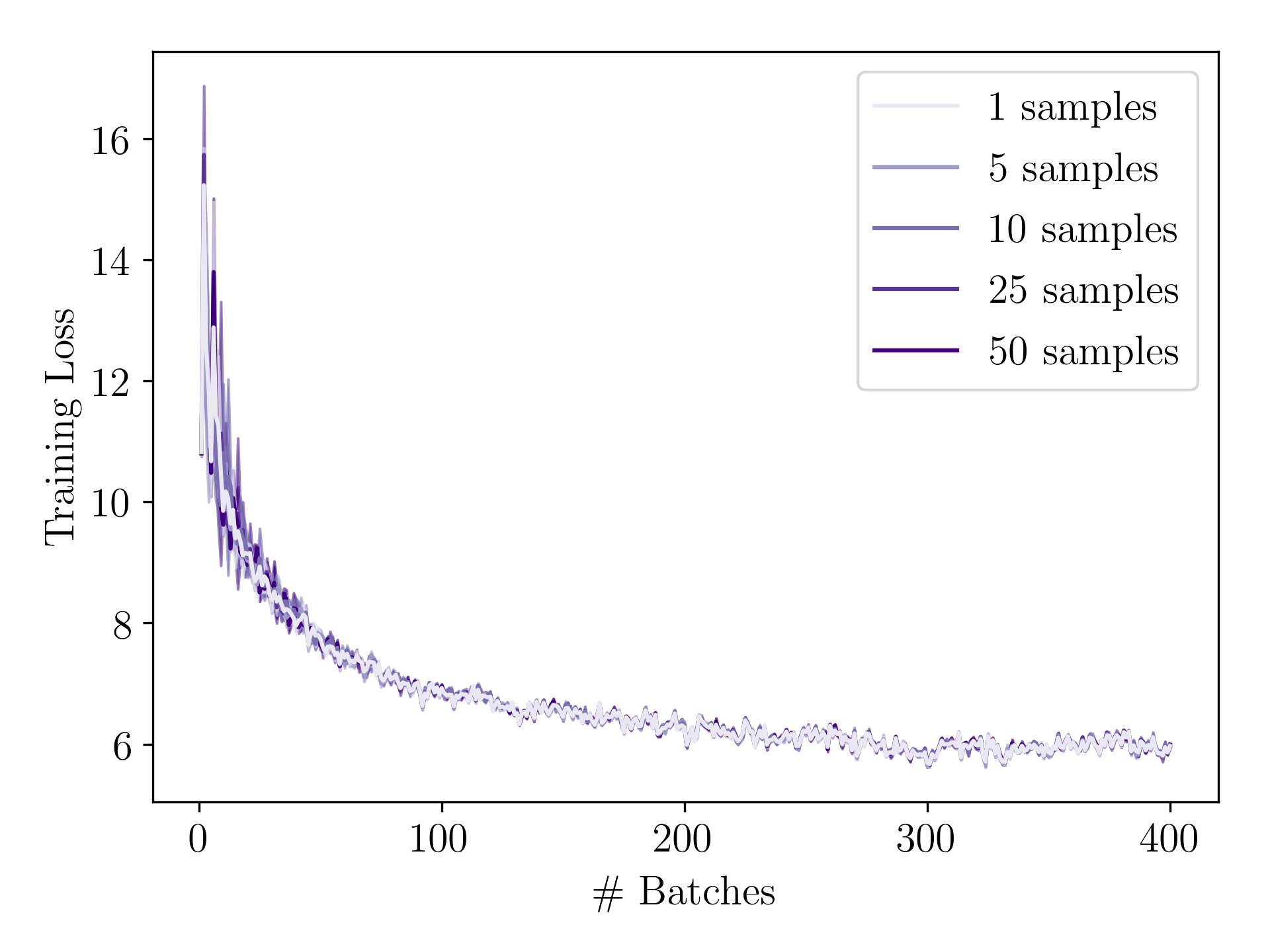} &
    \includegraphics[width=0.495\textwidth]{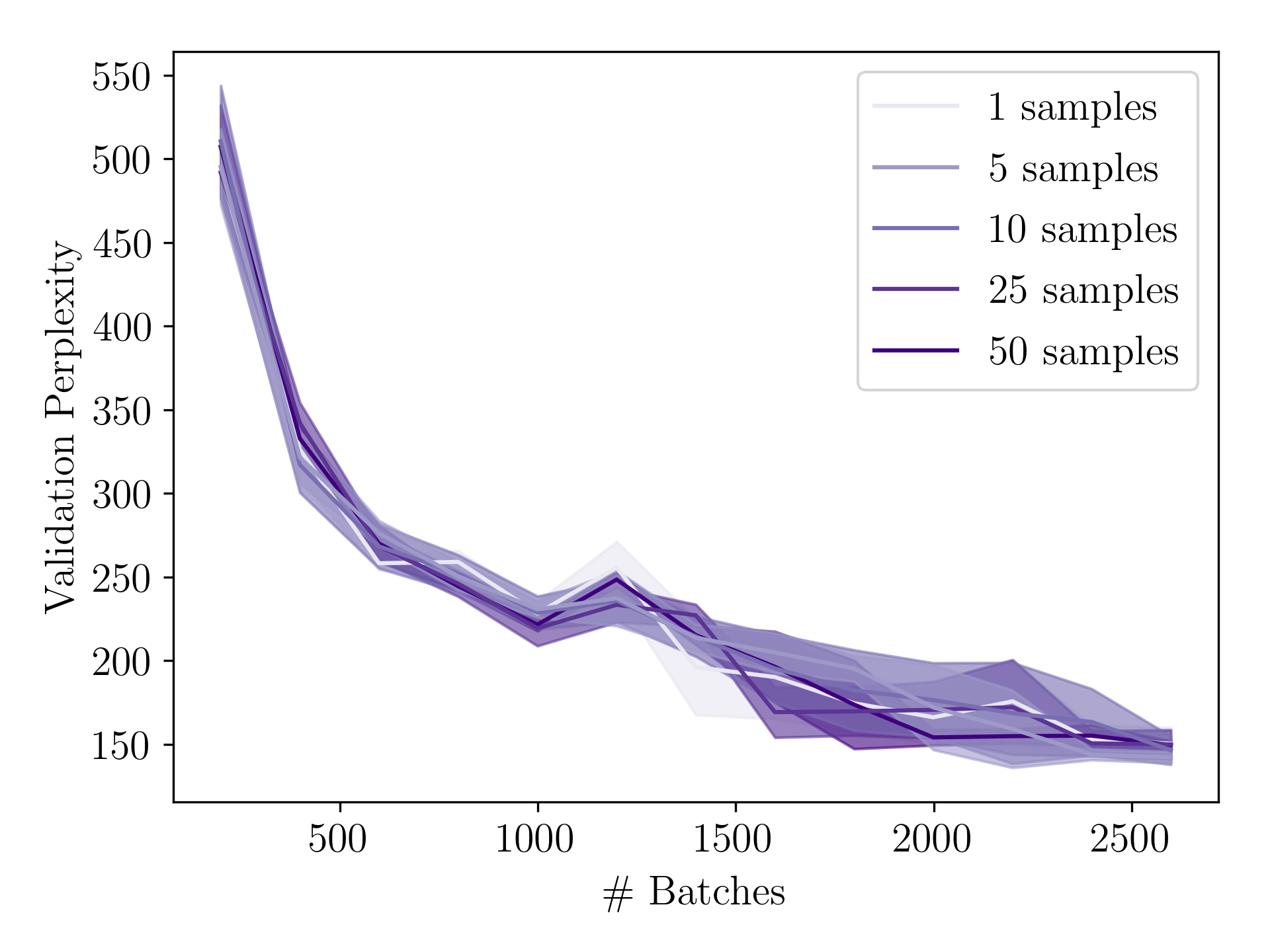} \\
    \includegraphics[width=0.495\textwidth]{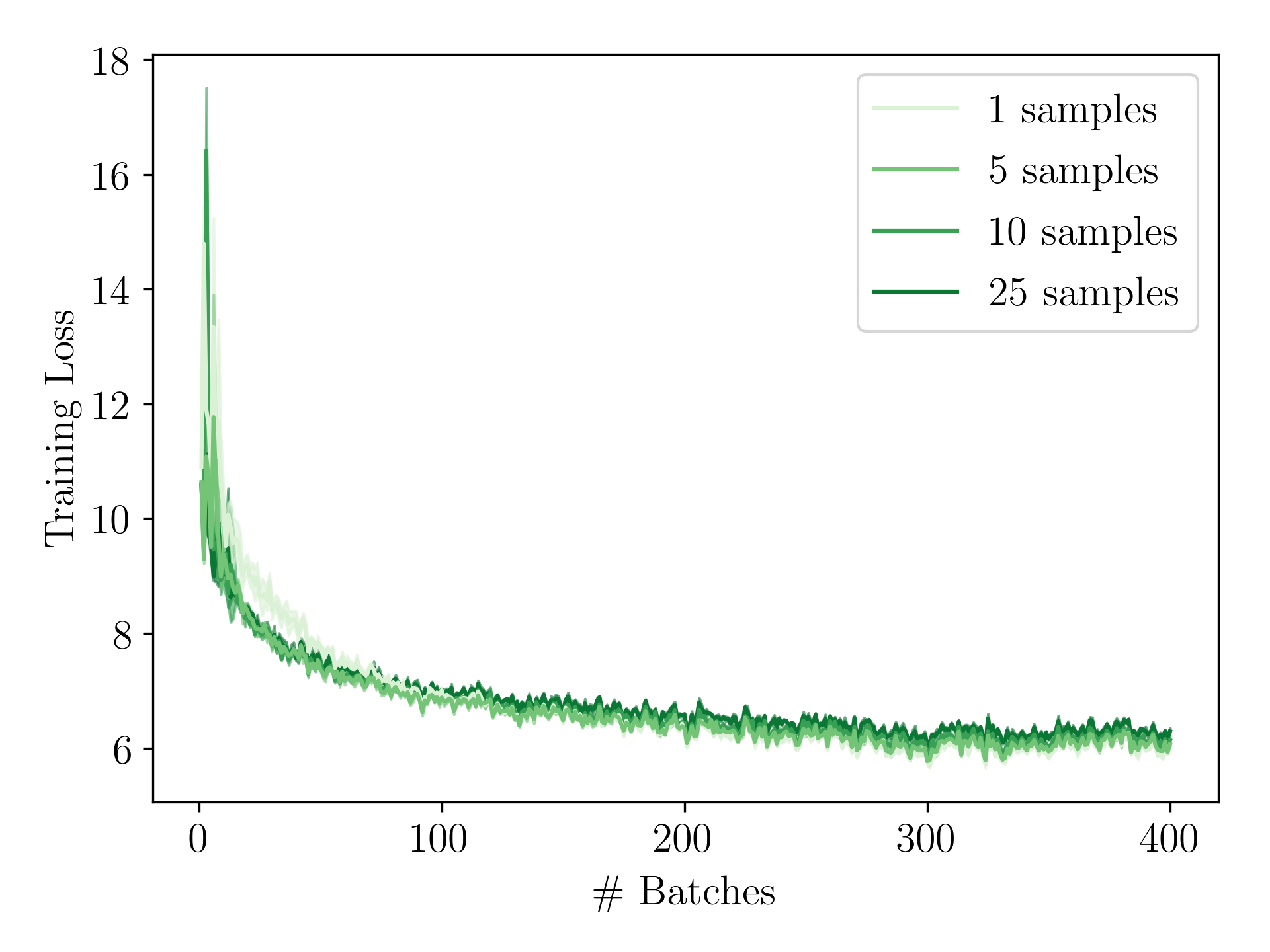} &
    \includegraphics[width=0.495\textwidth]{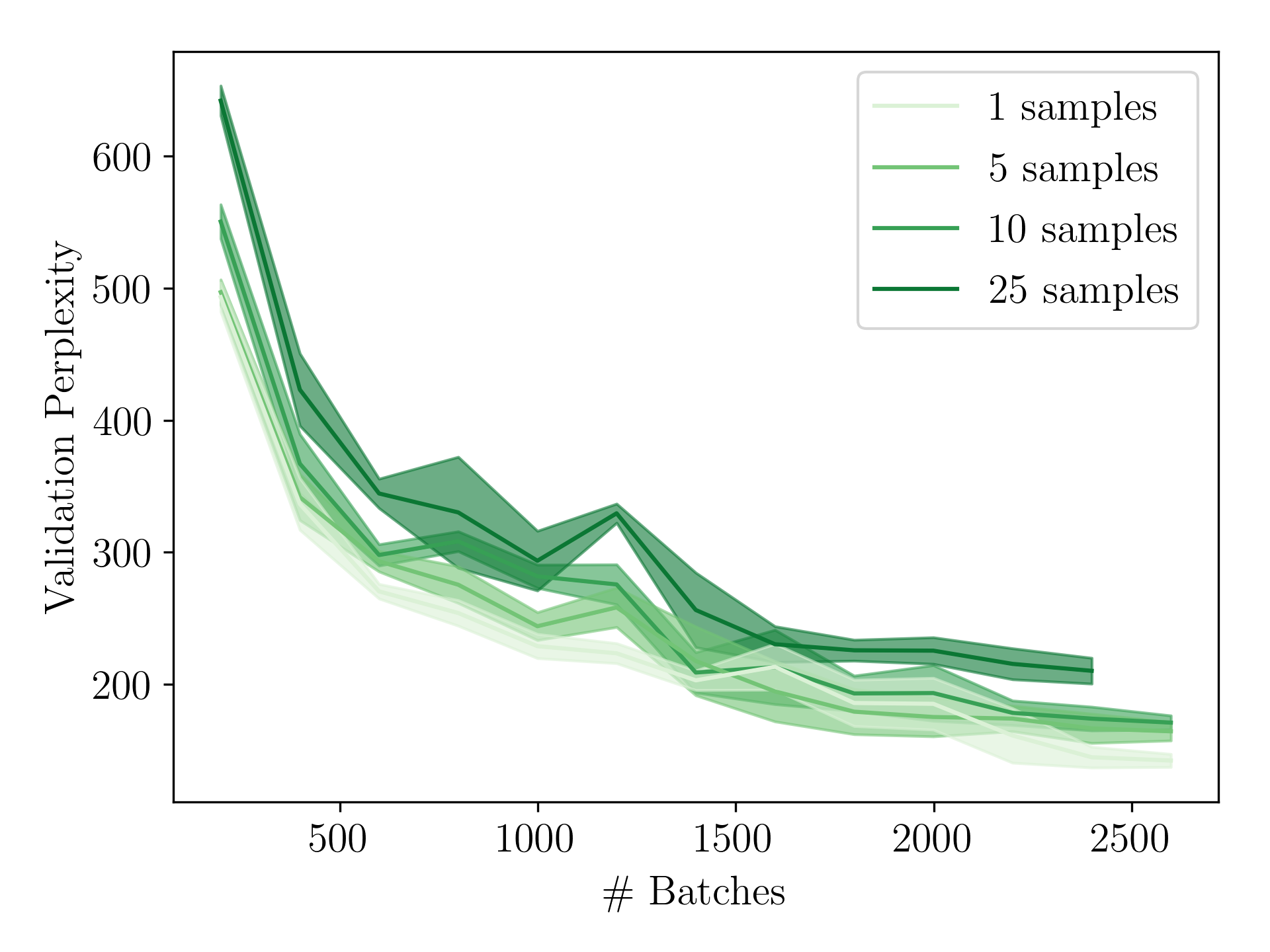} \\
  \end{tabular}
  \caption[Training losses for different numbers of samples]{Training loss (left column) and validation perplexities (right column) on \gls{PTB} using a different amount of samples to approximate predictive entropy using \mcdcol{MC Dropout} and \baecol{Bayesian anchored ensemble} recoding. Curves denote the mean over $n=4$, with intervals signifying the standard deviation. Best viewed in color.}\label{fig:exp-num-samples-curves}
\end{figure}

For \gls{MCD} and \gls{BAE} recoding, the error signal, i.e.\ the predictve entropy of the model, cannot be computed precisely but has to be approximated using a number of samples. To understand the extent to which the quality of this approximation influences the model performance, the two model variants are trained using an increasing amount of samples $K$. This happens on multiple levels: Similar to the previous section, first loss curves and test set results are given in fig. \ref{fig:exp-num-samples-curves} and table \ref{table:exp-num-samples-results}, respectively.\\

\begin{table}[h]
  \centering
  \setlength{\tabcolsep}{8pt}
  \def\arraystretch{1.5}
  \resizebox{.95\textwidth}{!}{
  \begin{tabular}{@{\extracolsep{4pt}}rllllll@{}}
       \toprule[1.5pt]
        & \multicolumn{4}{c}{Recoding Models} & & \\ \cline{2-5}
        & \multicolumn{2}{c}{\mcdcol{MC Dropout}} & \multicolumn{2}{c}{\baecol{BAE}} & \multicolumn{2}{c}{Baseline} \\ \cline{2-3} \cline{4-5} \cline{6-7}
        \# Samples & Perplexity & $\varnothing\ $Speed & Perplexity & $\varnothing\ $Speed & Perplexity & $\varnothing\ $Speed \\
       \toprule[1.15pt]
        $1$  & $145.77 \pm 8.99$ & $66.85$  & \baecol{$138.54 \pm 4.24$} & $59.15$ & $\mathbf{126.60 \pm 4.06}$ & $\mathbf{406.35}$ \\
        $5$  & \mcdcol{$138.02 \pm 2.79$} & $48.3$   & $161.30 \pm 9.22$ & $32.2$ & & \\
        $10$ & $140.81 \pm 7.60$ & $31.15$  & $170.41 \pm 5.22$ & $18.55$ & & \\
        $25$ & $143.88 \pm 7.03$ & $14$     & $215.23 \pm 12.16$ & $8.05$ & & \\
        $50$ & $143.85 \pm 2.81$ & $7.35$   & - & - & & \\
       \bottomrule[1.15pt]
  \end{tabular}}
  \vspace{0.2cm}
  \caption[Test perplexities for different numbers of samples]{Test perplexity on the test set for different models and a variety of recoding step sizes as well as average inference speed in tokens per second. Results are based on $n=4$ runs. Speed is based running the evaluation on a single NVIDIA 1080TI graphics card.}\label{table:exp-num-samples-results}
\end{table}

Here we can identify very stark differences between the two models: Although training and loss curves seem to overlap almost completely, increasing the sample size seems to only yield diminishing returns after $5$ samples for \gls{MCD} recoding. For \gls{BAE} recoding, the best result is achieved by just using a single (!) ensemble member. In both instances, the loss in speed is immediately apparent, as presented in the additional column of table \ref{table:exp-num-samples-results}: In the case of MC Dropout, additional Dropout masks have to be sampled for every $k$. In the case of Bayesian Anchored Ensembling, an additional member of the ensemble has to make a prediction. This combined with the additional gradient computations slow down both models by a factor of around $6$ in the case of a single $k$.\\

\begin{figure}[h]
  \centering
  \begin{tabular}{cc}
    \includegraphics[width=0.495\textwidth]{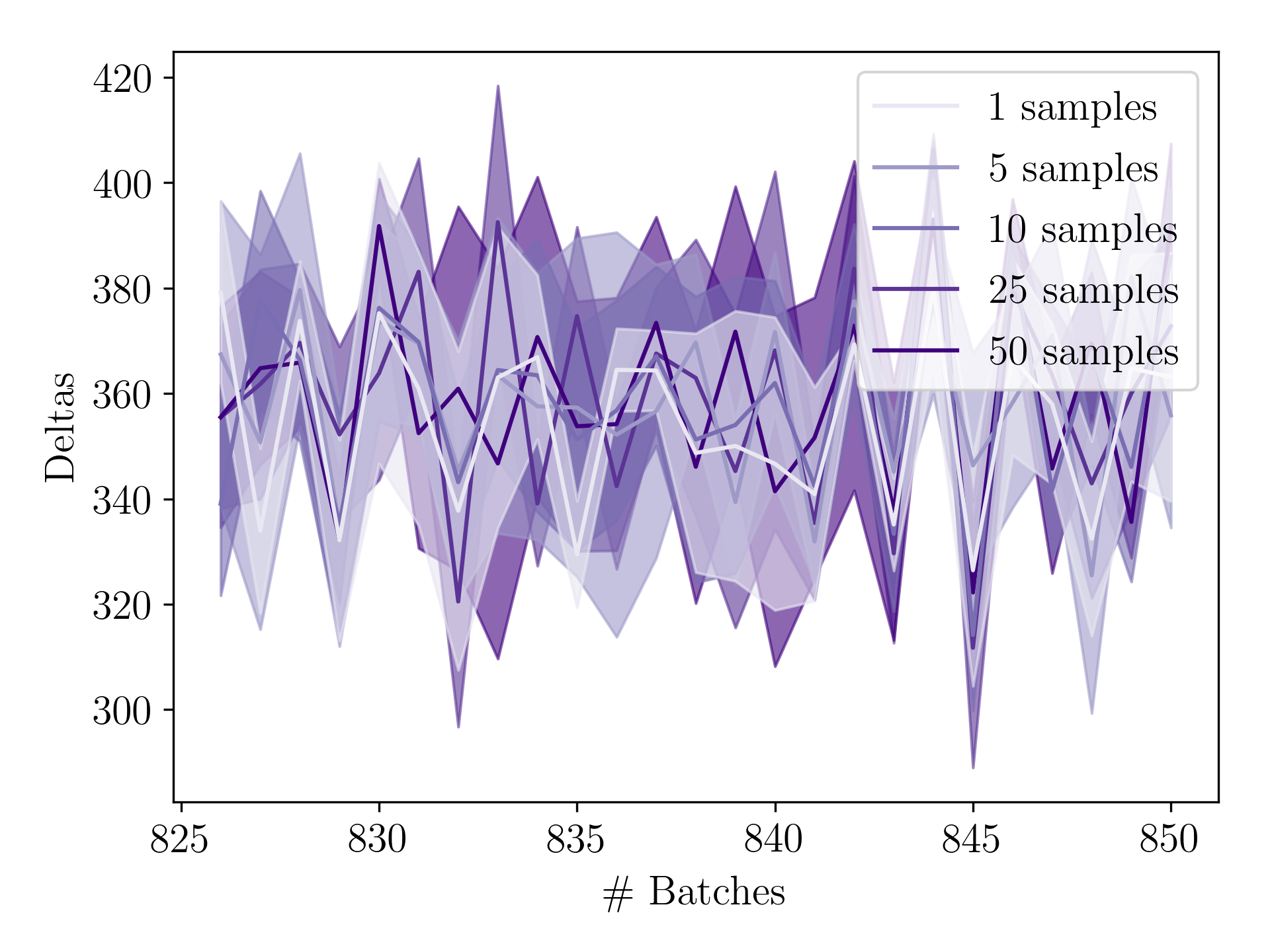} &
    \includegraphics[width=0.495\textwidth]{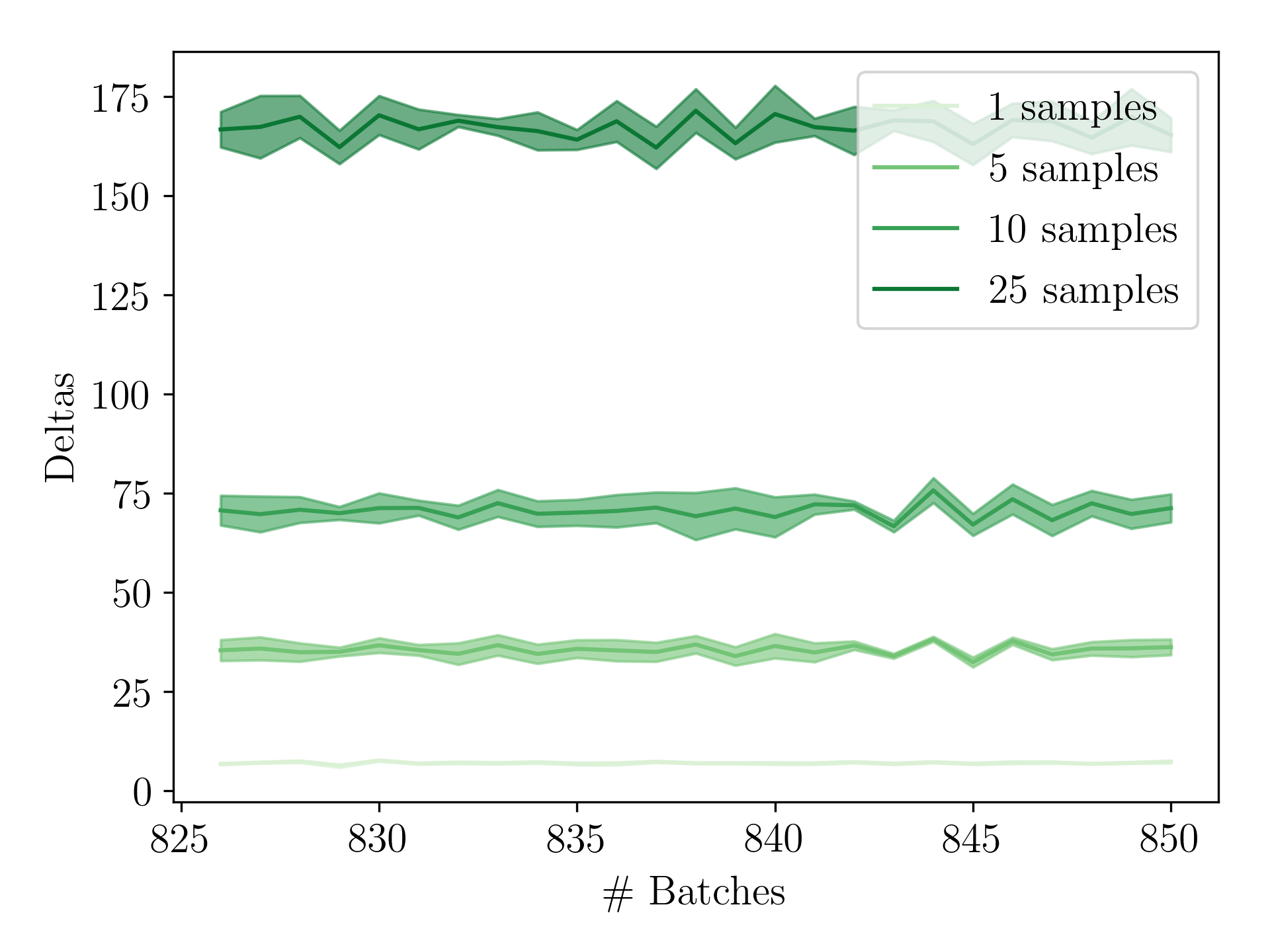} \\
  \end{tabular}
  \caption[Uncertainty estimates for different numbers of samples]{Error signals on \gls{PTB} using a different amount of samples on a later stage of the training using \mcdcol{MC Dropout} and \baecol{Bayesian anchored ensemble} recoding. Curves denote the mean over $n=4$, with intervals signifying the standard deviation. Best viewed in color.}\label{fig:exp-num-samples-deltas}
\end{figure}

Lastly, let us briefly investigate the agreement between the uncertainty estimates of both models as well as the agreement among instances of the same model using a different number of samples. Fig. \ref{fig:exp-num-samples-deltas} displays the average estimated uncertainty per batch during a later stage in the training. These plots seem to elucidate why \gls{BAE} recoding performs best only using a single member: While the different estimates produced by MC Dropout seem to highly correlate and to some degree be refined by more samples, they appear completely unrelated in the \gls{BAE} case. This comes as a surprise given the theoretical justification for using \gls{BAE} to estimate the predictive uncertainty in appendix \ref{appendix:pred-entropy}. Yet an amortization of the training loss was necessary to make training in the recurrent recoding context feasible, which appears to have a fatal effect on the quality of uncertainty estimates. It thus gives an explanation to the best score being achieved by a mere single ensemble member, corresponding simply to the entropy of the output distribution.

\subsection{MC Dropout rate}\label{sec:exp-dropout-rate}

\begin{figure}[!tbp]
  \centering
  \captionsetup[subfigure]{justification=centering}
  \begin{subfigure}[b]{0.38\columnwidth}
    \centering
    \setlength{\tabcolsep}{8pt}
    \def\arraystretch{1.5}
    \resizebox{.8\textwidth}{!}{
    \begin{tabular}{@{}rc@{}}
         \toprule[1.5pt]
         Dropout & Perplexity \\
         \toprule[1.15pt]
          $0.1$ & $141.32 \pm 4.83$ \\
          $0.2$ & $140.37 \pm 4.49$ \\
          $0.3$ & $145.51 \pm 4.97$ \\
          $0.4$ & $139.14 \pm 8.29$ \\
          $0.5$ & $143.95 \pm 4.17$ \\
          $0.6$ & $143.25 \pm 3.91$  \\
         \bottomrule[1.15pt]
    \end{tabular}}
    \vspace{0.2cm}
    \caption{Test perplexities.}\label{table:exp-dropout-rate-results}
  \end{subfigure}%
  \hfill
  \begin{subfigure}[b]{0.58\columnwidth}
    \centering
    \includegraphics[width=0.95\textwidth]{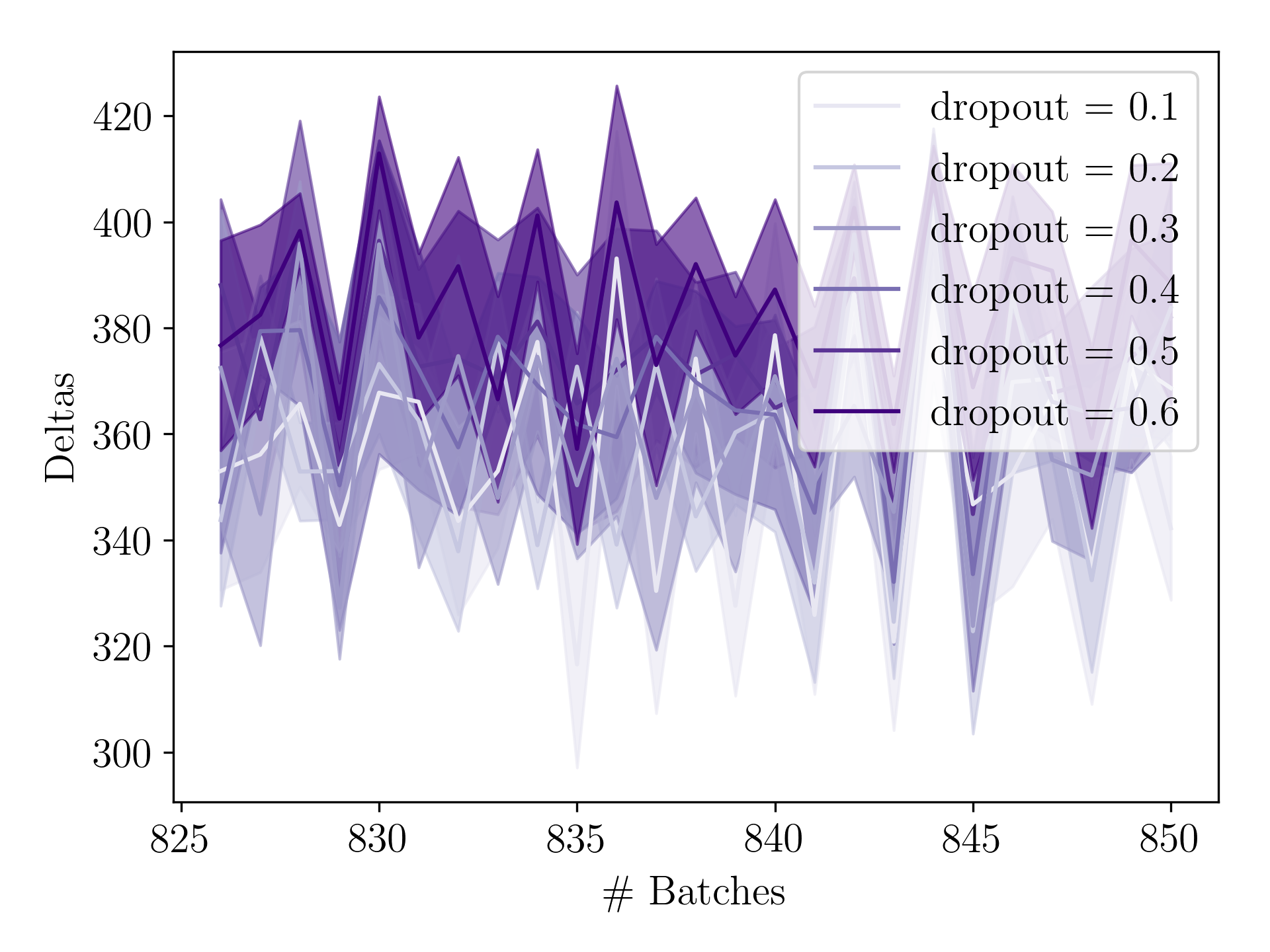}
    \caption{Uncertainty estimates.}\label{fig:exp-dropout-deltas}
  \end{subfigure}%
  \caption[Results for using \gls{MCD} recoding with different Dropout rates.]
  {Results for different Dropout rates using MC Dropout recoding. (a) Perplexities on the test set. (b) Uncertainty estimates during a later stage in training. Curves denote the mean over $n=4$ runs, with intervals signifying the standard deviation. Best viewed in color.}\label{fig:exp-dropouts}
\end{figure}

Now we look at the impact of using different values for the MC Dropout rate. The test perplexities for different Dropout rates are given in table \ref{table:exp-dropout-rate-results}. The results hint at the fact that the Dropout rate does not seem to produce significantly different outcomes for the options tested, a notion which is reinforced by the training and validation curves in fig. \ref{fig:exp-dropout-curves} in the appendix.\\

More interesting however is to look at the uncertainty estimates for the same window during training as in the previous section, which is given in fig. \ref{fig:exp-dropout-deltas}. Here we can clearly identify a relationship
between the Dropout probability and the estimates of the predictive uncertainty: The lower the rate, the higher the uncertainty.

This interdependency can be explained by the following observation: An increased Dropout probability leads to more neurons whose activations become zero. As each neuron outputs the probability of a predicted word, averaging over $K$ distributions with different masks, this will lower the probability of the same word when averaging samples. Ultimately, this leads to a flatter distribution (or increased uniformity) and therefore a higher predictive entropy.\\

Based on the results, it remains unclear how to select the ``best'' Dropout rate to estimate the predictive uncertainty. \cite{gal2016dropout} note that for some of their experiments, using different Dropout rates the models actually converged to the same uncertainty estimates over time. This cannot be said in our case, where the estimates differ in similar proportions even approaching the end of the training (not shown here). However, as the choice of Dropout rate does not seem to produce any significant differences, we proceed in the following experiments with the Dropout rates found during hyperparameter search. This value of $0.15$ for the Language Model's decoder is also in in line with the experiments performed by \cite{gal2016dropout}, who similarly choose small values for models of very limited size.

\subsection{Dynamic Step Size}\label{sec:learned-step-size}

So far, we only considered a recoding step size $\alpha$ that stays fixed throughout training. However, as shown in theorem \ref{theorem:error-reduction}, a reduction of the error signal $\delta_t$ in the hidden activations $\bh_t$ is only guaranteed using the correct step size, which in turn depends on the error function's Lipschitz constant $L$. This constant is not known, and although making the step size a learnable parameter is also not guaranteed to find the correct value, it does seem more promising than determining it manually (\emph{learned step size}). Because we saw earlier that the recoding step size required varies highly depending on the model and should not be constrained to a certain range, regularization like weight decay - if applicable - is not applied to this parameter. Furthermore, the softplus function (eq. \ref{eq:softplus}) is chosen to guarantee a positive step size.\footnote{It was observed in earlier experiments that applying a \gls{ReLU} to the step size would tend to have the step size parameter dying off, resulting in a step size of $0$ that would disable recoding and could not be recovered from.}
We also consider the even more dynamic case described in chapter \ref{sec:step-size}, where we parameterize the recoding step size at a time step $t$ with a neural network with parameters $\bkappa$ s.t. $\alpha_t = f_\bkappa(\bh_t)$ (\emph{predicted step size}). As before, the softplus function is used on the final output to restrict it to positive values. In this case, the predictor network was chosen to be a two-layer \gls{MLP} with $300$ and $100$ hidden units, respectively.\\

\begin{figure}[h]
  \centering
  \captionsetup[subfigure]{justification=centering}
  \begin{subfigure}[b]{0.5\columnwidth}
    \setlength{\tabcolsep}{8pt}
    \def\arraystretch{1.5}
    \resizebox{.95\textwidth}{!}{
      \begin{tabular}{@{}rrll@{}}
        \toprule[1.5pt]
        Model & Step type & Perplexity & Gain \\
        \toprule[1.5pt]
        Baseline                          & -                 & $126.60 \pm 4.06$ & N/A \\
        \surpcol{Surp.}                   & \surpcol{Learned} & $126.25 \pm 0.77$ & \downdiff{-0.92} \\
                                          & \surpcol{Pred.}   & $129.01 \pm 2.42$ & \downdiff{-3.68} \\
        \mcdcol{MCD}                      & \mcdcol{Learned}  & $140.50 \pm 6.78$ & \downdiff{-2.48} \\
                                          & \mcdcol{Pred.}    & $154.48 \pm 14.32$ & \downdiff{-16.46} \\
        \baecol{BAE}                      & \baecol{Learned}  & $142.20 \pm 8.84$ & \downdiff{-3.66} \\
                                          & \baecol{Pred.}    & $139.55 \pm 6.30$ & \downdiff{-1.01} \\
        \bottomrule[1.15pt]
      \end{tabular}
    }
    \vspace{0.2cm}
    \caption{Results for dynamic step sizes.}\label{subfig:dynamic-steps-perplexities}
  \end{subfigure}\hfill%
  \begin{subfigure}[b]{0.5\columnwidth}
    \centering
    \includegraphics[width=0.985\textwidth]{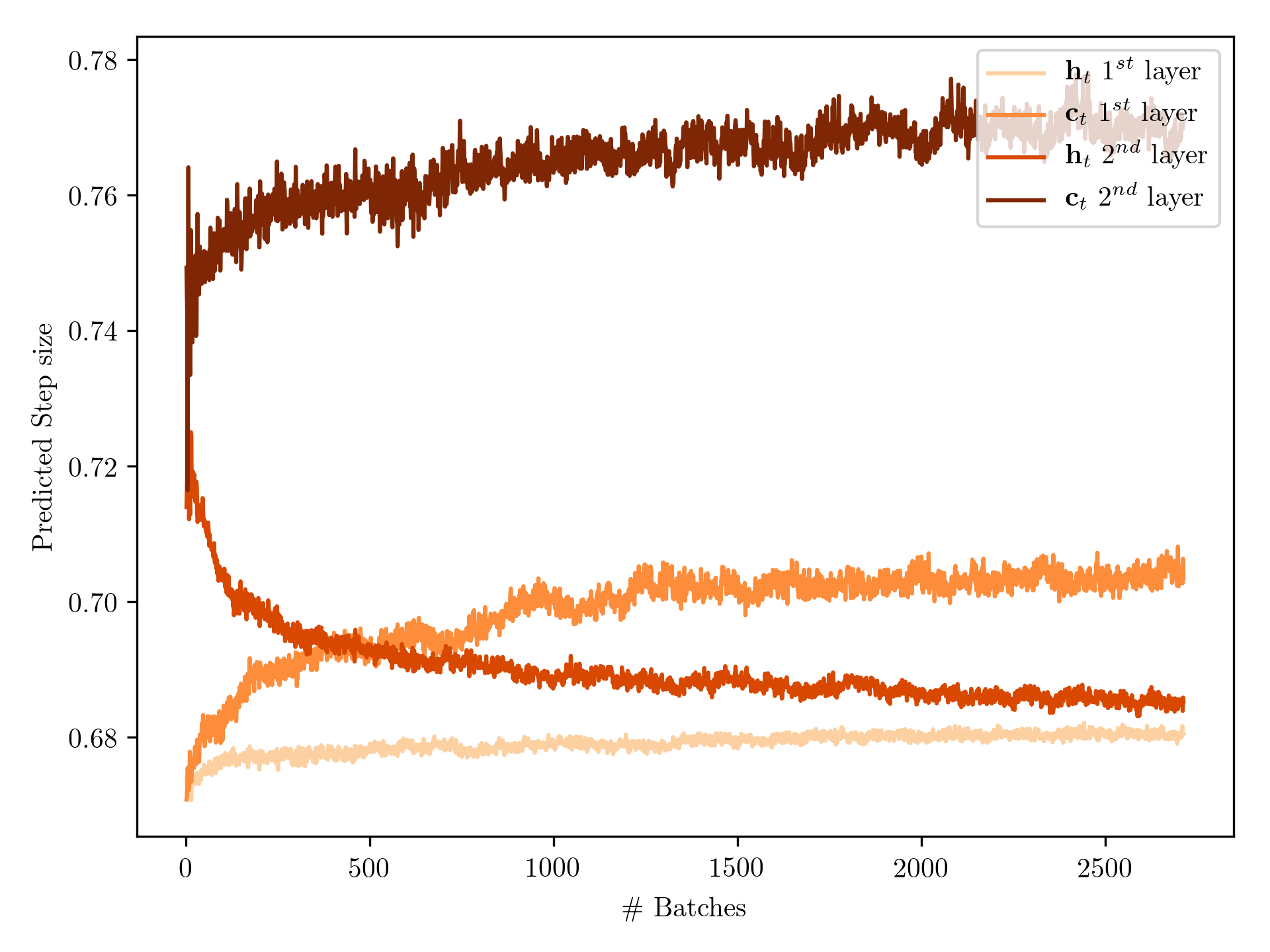}
    \caption{Learned steps for single surprisal model.}\label{subfig:perlexity_learned_single}
  \end{subfigure}\\
  \begin{subfigure}[b]{0.5\columnwidth}
    \centering
    \includegraphics[width=0.985\textwidth]{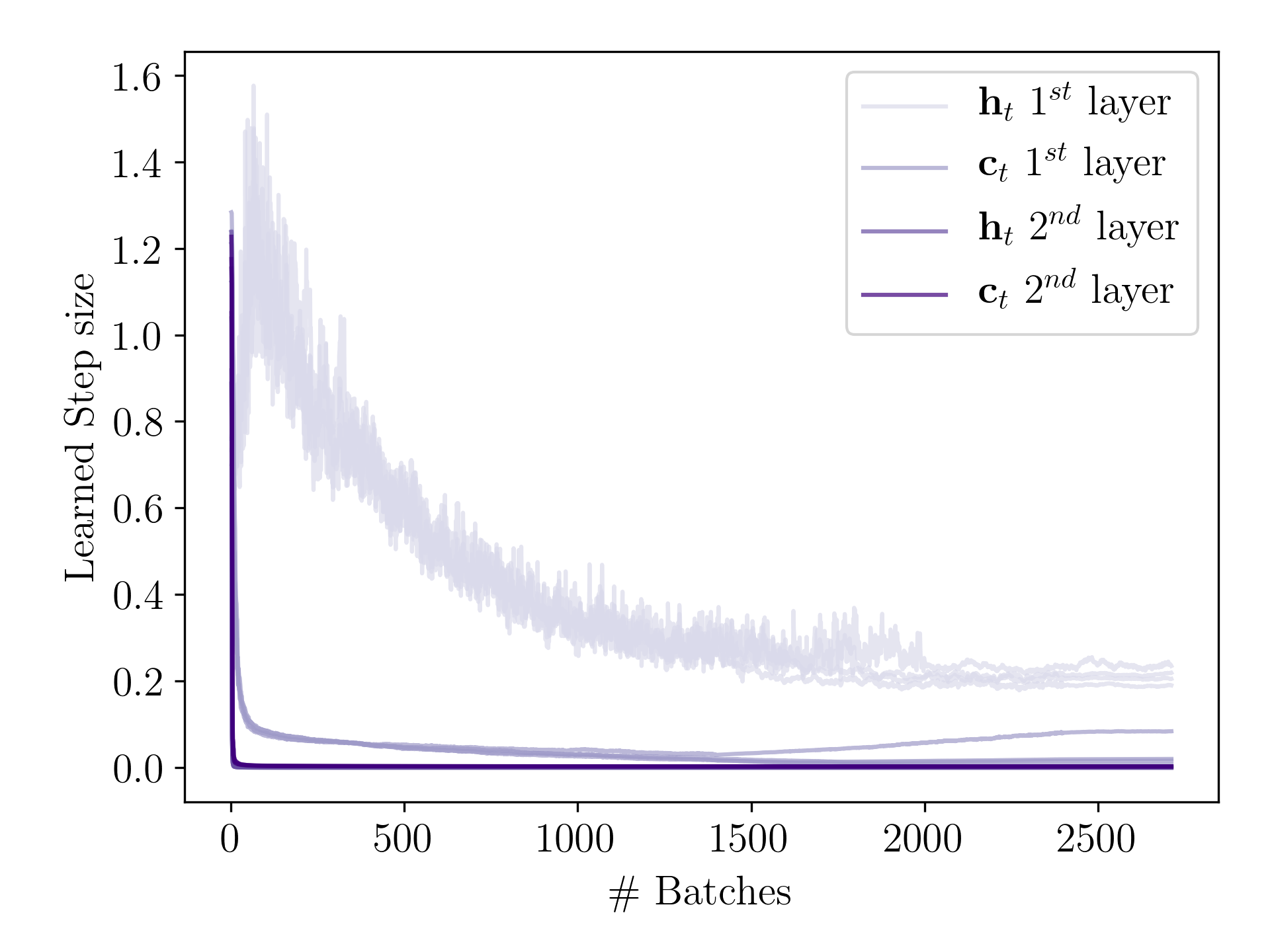}
    \caption{Learned steps for $n=4$ \gls{MCD} models.}
  \end{subfigure}\hfill%
  \begin{subfigure}[b]{0.5\columnwidth}
    \centering
    \includegraphics[width=0.985\textwidth]{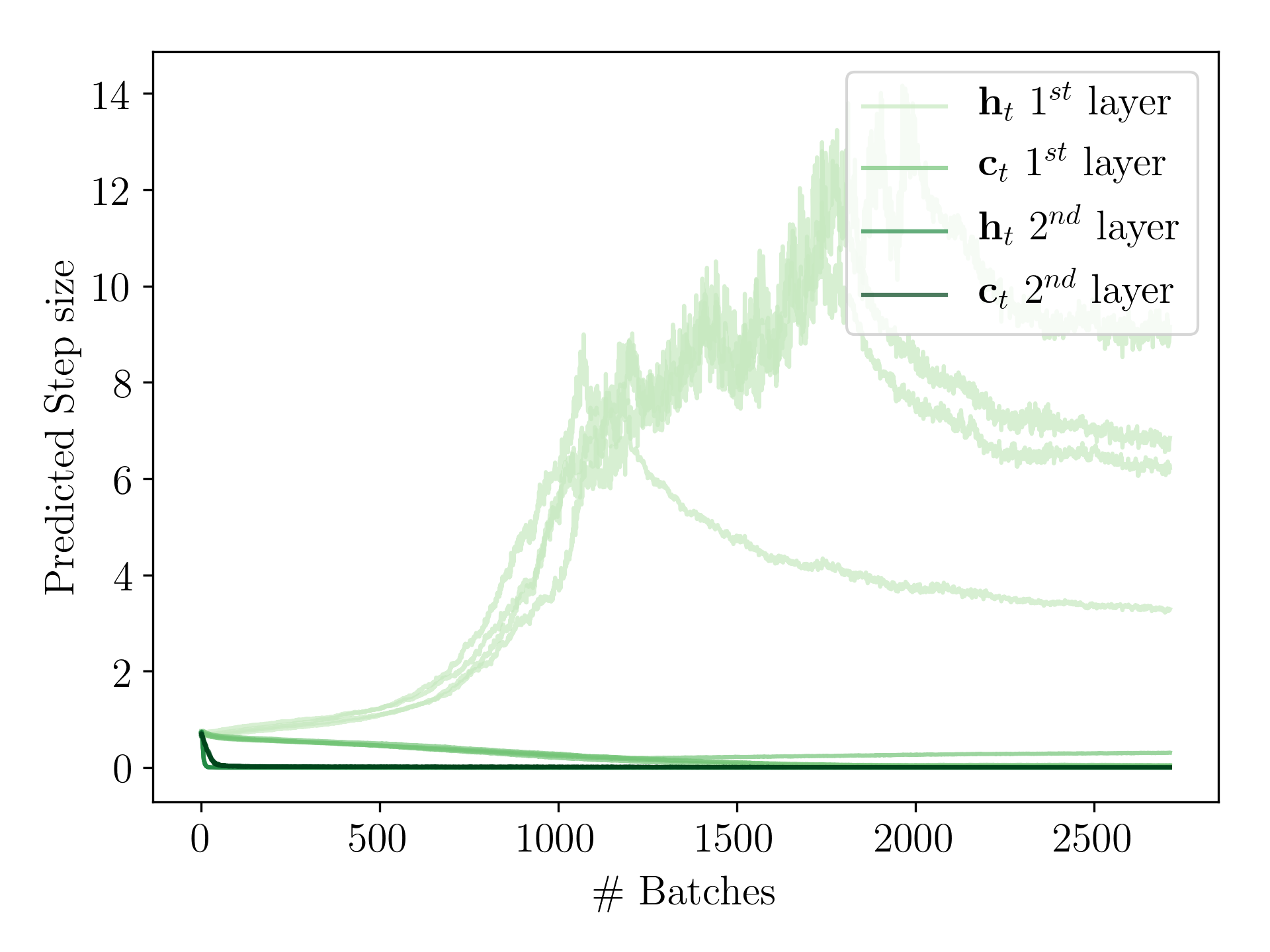}
    \caption{Predicted steps for $n=4$ \gls{BAE} models.}
  \end{subfigure}
  \caption[Overview over results with dynamic step size models.]{Overview over the results with dynamic step size models. (a) Perplexities on the test set
  including the difference to best results for every model type so far. (b) - (d) Development of the average step size per batch for a certain layer and activation type during training for the three recoding models using either learned or predicted steps.}\label{fig:dynamic-steps-results}
\end{figure}

The results given in table \ref{subfig:dynamic-steps-perplexities} reveal the difficulty in determining the step size: No matter the model variant or manner to pick the recoding step, all models perform worse. Furthermore, all models disagree on the relative importance of correcting hidden activations: When predicting the update step with a \gls{MLP}, the cherry-picked surprisal model depicted in fig. \ref{subfig:perlexity_learned_single} places emphasis on recoding the cell states of the \gls{LSTM}, preferring the upper layer, in contrast to the \gls{MCD} and \gls{BAE} models, which recode the hidden activations of the lowest layer the most and has other sizes tend close to $0$. Curiously, the averaged step sizes - either learned or predicted - seem to be very similar in these two cases regardless of parameterization.
This cannot be said for surprisal recoding, where in either cases both methods converge to distinct solutions in different runs (cp. with fig. \ref{fig:dynamic-steps-additional-results} in appendix). This seems to hint at a high variability in the training process of these recoding step parameters which does not find advantageous solutions and ultimately inhibits performance. This fragility could stem from multiple origins, including the flux of the error function surface the step size has to correspond to and the unregularized predictors. An overall reflection on the issue of step size is given later in chapter \ref{sec:discussion-step-size}.

\subsection{Ablation}\label{sec:ablation}

After reviewing previous results, several questions arise: Why does surprisal encoding only perform marginally better? Can we improve an independently trained baseline model by adding the recoder? Does recoding using the model's approximate predictive entropy result in worse performance by virtue of the recoding mechanism or does training with the mechanism result in suboptimal weights?\\

I attempt to answer these questions by performing a series of ablation experiments: Here the original best evaluation scores per model type are compared against some modified instances, namely all recoding models \emph{without} their respective recoders as well as the baselines \emph{adding} the different mechanisms on top. All of this is solely performed during testing, using the models trained in earlier sections. The results of these experiments are given in fig. \ref{table:exp-ablation}, where the unmodified models are used for comparison.\\

\begin{figure}
  \captionsetup[subfigure]{justification=centering}
  \centering
  \begin{subfigure}[b]{0.5\columnwidth}
    \centering
    \setlength{\tabcolsep}{8pt}
    \def\arraystretch{1.5}
    \resizebox{.95\textwidth}{!}{
      \begin{tabular}{@{}rrccc@{}}
           \toprule[1.5pt]
           Model & Step & Modification & Result & Gain \\
           \toprule[1.15pt]
            \multirow{3}{*}{\surpcol{Surp.}} & \surpcol{Fixed}   & \multirow{3}{*}{$\vast\}$ w/o recoder}  & $126.11 \pm 1.38$ & \downdiff{-0.78} \\
                                             & \surpcol{Learned} &                                         & $126.25 \pm 0.77$ & \samediff{0} \\
                                             & \surpcol{Pred.}   &                                         & $129.01 \pm 2.42$ & \samediff{0} \\
            \bottomrule[1pt]
            \multirow{3}{*}{\mcdcol{MCD}}   & \mcdcol{Fixed}   & \multirow{3}{*}{$\vast\}$ w/o recoder}  & $138.01 \pm 2.79$ & \updiff{+0.01} \\
                                            & \mcdcol{Learned} &                                         & $145.74 \pm 8.95$ & \downdiff{-5.24} \\
                                            & \mcdcol{Pred.}   &                                         & $159.60 \pm 10.93$ & \downdiff{-5.12} \\
            \bottomrule[1.15pt]
            \multirow{3}{*}{BAE}            & \baecol{Fixed}   & \multirow{3}{*}{$\vast\}$ w/o recoder}  & $138.54 \pm 4.24$ & \samediff{0} \\
                                            & \baecol{Learned} &                                         & $142.36 \pm 7.76$ & \downdiff{-0.16} \\
                                            & \baecol{Pred.}   &                                         & $140.51 \pm 5.48$ & \downdiff{-0.96}  \\
            \bottomrule[1.15pt]
      \end{tabular}}
      \vspace{0.2cm}
      \caption{Models without recoder.}\label{table:exp-ablation-removal}
  \end{subfigure}\hfill%
  \begin{subfigure}[b]{0.5\columnwidth}
    \centering
    \setlength{\tabcolsep}{8pt}
    \def\arraystretch{1.5}
    \resizebox{.995\textwidth}{!}{
      \begin{tabular}{@{}rrccc@{}}
         \toprule[1.5pt]
         Model & Step & Modification & Result & Gain \\
         \toprule[1.15pt]
          \multirow{3}{*}{BL} & \surpcol{Fixed}   & \multirow{3}{*}{$\vast\}$ \surpcol{w/ Surp. recoder}} & $125.95 \pm 3.56$ & \updiff{+0.65} \\
                              & \surpcol{Learned} &                                                       & $126.55 \pm 3.63$ & \updiff{+0.05} \\
                              & \surpcol{Pred.}   &                                                       & $126.59 \pm 3.64$ & \updiff{+0.01} \\
          \bottomrule[1.15pt]
          \multirow{3}{*}{BL} & \mcdcol{Fixed}   & \multirow{3}{*}{$\vast\}$ \mcdcol{w/ MCD recoder}}   & $126.78 \pm 3.67$ & \downdiff{-0.18} \\
                              & \mcdcol{Learned} &                                                      & $127.04 \pm 4.11$ & \downdiff{-0.44} \\
                              & \mcdcol{Pred.}   &                                                      & $127.11 \pm 4.14$ & \downdiff{-0.51} \\
          \bottomrule[1.15pt]
          \multirow{3}{*}{BL} & \baecol{Fixed}   & \multirow{3}{*}{$\vast\}$ \baecol{w/ BAE recoder}}   & $126.72 \pm 3.65$ & \downdiff{-1.12} \\
                              & \baecol{Learned} &                                                      & $126.72 \pm 4.22$ & \downdiff{-1.12} \\
                              & \baecol{Pred.}   &                                                      & $126.72 \pm 4.22$ & \downdiff{-1.12} \\
         \bottomrule[1.15pt]
    \end{tabular}}
    \vspace{0.2cm}
    \caption{Baseline model with recoder.}\label{table:exp-ablation-replacement}
  \end{subfigure}
  \vspace{0.2cm}
  \caption[Results for ablation experiments.]{Test perplexity on the test set for different models using components from the baseline / recoding models. (a) Recoding models trained with the recoder but performing without it during testing. (b) The baseline model equipped with the recoding mechanisms of other models. Gains are calculated relative to the best results of this model type so far (i.e.\ Surprisal / \gls{MCD} / \gls{BAE} for (a) and the baseline for (b)).}\label{table:exp-ablation}
\end{figure}

Table \ref{table:exp-ablation-removal} brings an elusive interdependecy between the recoder and the underlying model to light: In almost all cases, removing the recoder results in worse or the same performance. Yet the discrepancy does not to appear as big as one might expect. This seems to suggest that the recoder loses some of its importance on the network's prediction as the model matures during training and that the underlying weights adapt to the mechanism. Moreover, the resulting models perform worse than the baseline ($126.60$) in the case of \gls{MCD} and \gls{BAE} recoding, from which we can conclude that these methods actually have an adverse impact on the learning of the model weights. Only in the case of surprisal recoding, the models still perform better, although statistically non-significant.\\

The notion that recoding based on predictive uncertainty has a negative effect on performance is further reinforced by table \ref{table:exp-ablation-replacement}: Using the corresponding recoding mechanisms with the baseline model results in a small drop in performance. Surprisal recoding also produces lower test perplexities here, yet miniscule in nature. To get a better understanding of the reasons behind these observations, the next chapter makes some observations about the models on a smaller scale.

\chapter{Qualitative Analysis}\label{chapter:qualitative}

Last chapter has provided a comprehensive comparison between the baseline and all proposed model variants in terms of test perplexity and other metrics on a corpus-level. However, this singular perspective fails to give insight into the core differences between the models: Because we would like to gain more introspection into these models to explain their differences, this chapter contains several experiments highlighting key model traits on a sentence or even word level by performing qualitative analyses. To this end, section \ref{sec:exp-behaviour} provides a check of cognitive plausibility according to the motivation layed on in the introduction.
It furthermore contains an empirical validation of theorems \ref{theorem:error-reduction} and \ref{theorem:error-reduction-tt} by measuring the amount of reduction in error signal recoding produces in section \ref{sec:exp-error-reduction} as well as an error analysis. Section \ref{sec:overconfident} tries to elucidate the curious performance gap between the baseline and models using predictive entropy as their error signal.

\section{Recoding Behaviour}\label{sec:exp-behaviour}

To analyze model behavior, we can track certain quantities on a word-to-word basis: One of them is the \emph{surprisal} of the Language Model. This approach has already been employed by other works to compare the surprisal of Language Models to human reading times \citep{van2018modeling} as a proxy for the difficulty of a sentence. However, this surprisal score differs from the recoding signal of the same name defined in chapter \ref{subsec:surprisal}, as it only consists of the base $2$ log probability of the word $-\log_2(w_t|w_1^{t-1}) \equiv -\log_2(\bo_{tc})$, where $c$ corresponds to the entry of word $w_{t+1}$ in the model's vocabulary. Also, we can track the recoding error signal $\delta_t$ emitted at some time step $t$. Besides these two pieces of information, we also record the surprisal of the model based on the recoded hidden activations
$-\log_2(\bo_{tc}^\prime)$, i.e.\ after re-computing the output distribution using $\bh_t^\prime$, along with the resulting new error signal $\delta_t^\prime$. The latter is not actually used for any computations, it just serves as a mean to disclose the impact of recoding.\\

Finally, experiments are conducted in a fashion that allows recoding only to be performed at specific time steps, which are appropriately marked or explained in the caption. This way, we can isolate the effect of single recoding actions and observe their side effects on the model later during a sequence.

\subsection{Cognitive Plausibility}

One motivation for the approach taken lies in the emulation of human behavior when processing challenging sentences.
To test the cogntive resemblence of the new augmented model, a small corpus of $20$ garden path sentences
is used that were collected from multiple online sources as well as $64$ sentence from the work of \cite{sturt1999structural}, resulting in a total of $84$ sentences containing difficult structural ambiguities.
Similar to the infamous ``The horse raced past the barn fell'' example
mentioned in chapter \ref{chapter:intro}, these sentences evoke a feeling of surprise when encountering a new pivotal
word which does not correspond to the parse built up by the reader thus far. Accordingly, the sentence representation has to be adjusted,
which was observed to lead to an increase in reading time among human participants \citep{milne1982predicting}. Accordingly, if the model is (less) susceptible to garden path effects, we can observe this by its surprisal scores.\\

\begin{figure}[h]
  \begin{tabular}{cc}
    \includegraphics[width=0.495\textwidth]{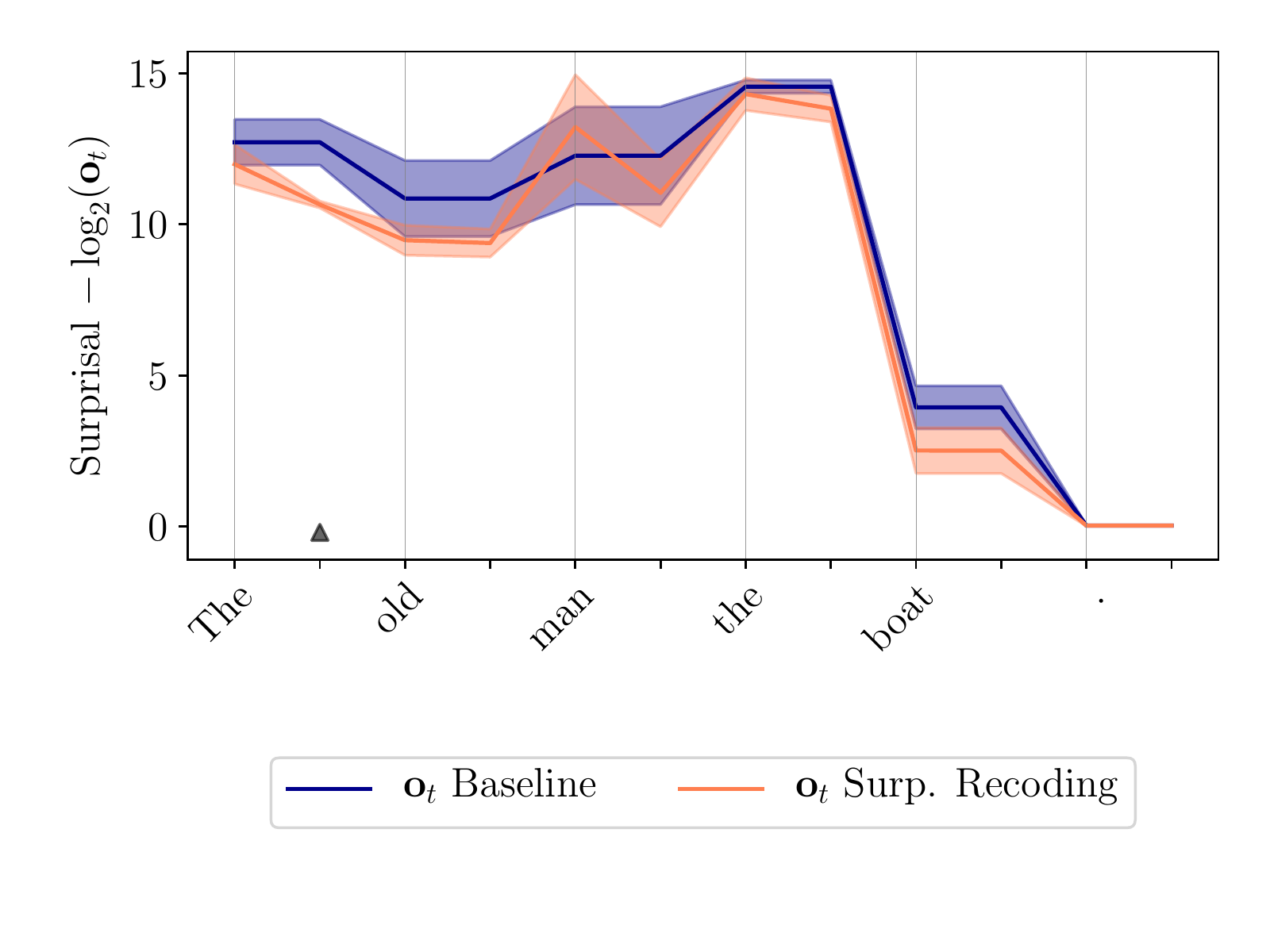} &
    \includegraphics[width=0.495\textwidth]{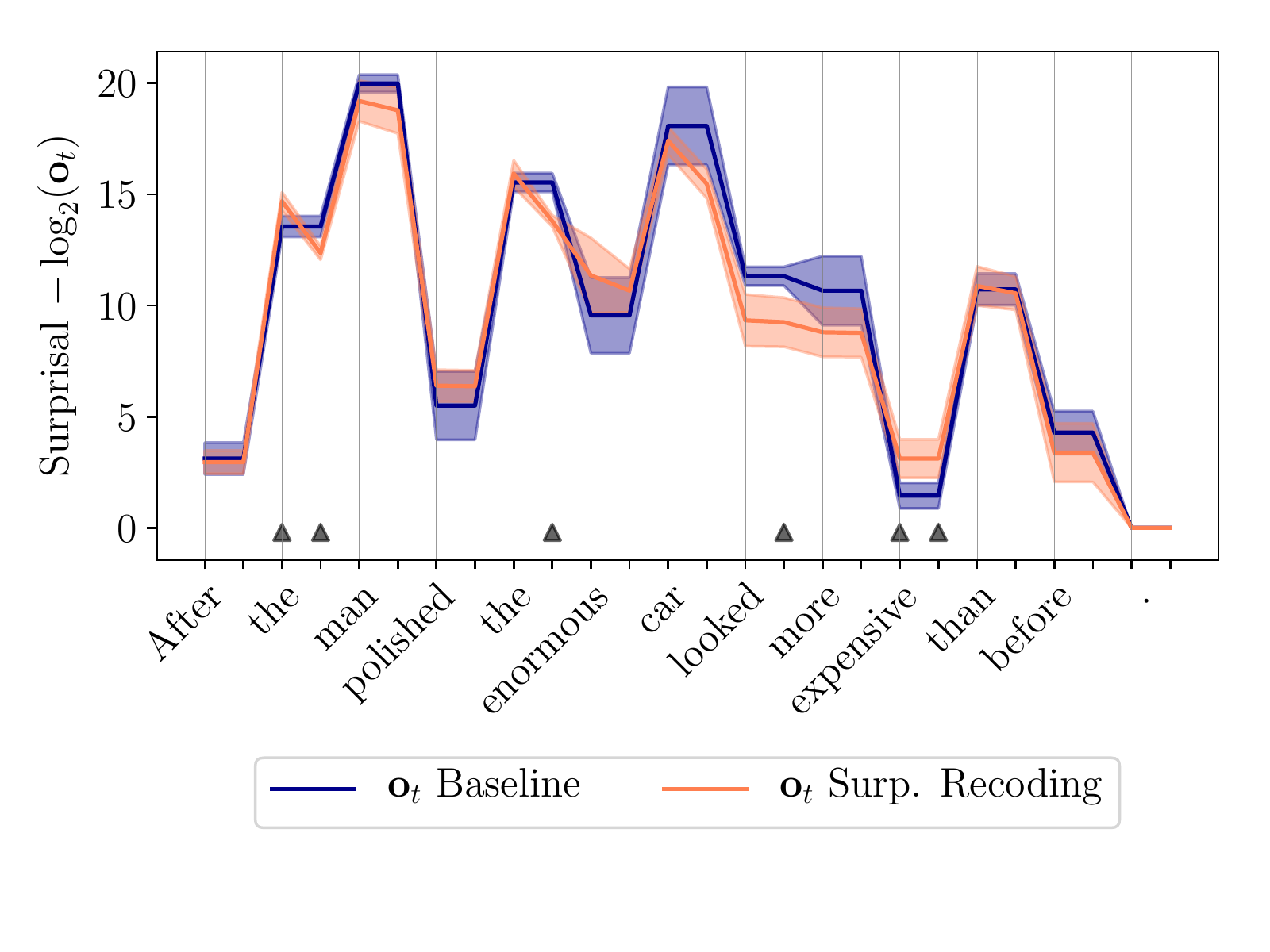} \\
  \end{tabular}
  \caption[Model surprisal scores on two garden path sentences]{Surprisal scores of the baseline and a surprisal recoding model on two garden path sentences. Curves denote the mean of $n=4$ runs, while the intervals depict one standard deviation. In case of the recoding models, the surprisal score based on the re-decoded output distribution is shown \emph{between} two time steps. Gray markers denote statistically significant differences (two-tailed Student's t-test, $p=0.05$). Best viewed in color.}\label{fig:garden-path-biological}
\end{figure}

In our case the processing time of new inputs is expected to remain nearly constant and does not depend on some notion of ``difficulty''; instead we would expect to see an increase
in the surprisal score and / or error signal $\delta_t$ when the model is confronted with an unexpected token, followed by
a decrease thereof as a consequence of the recoding update step. To test these assumptions, fig. \ref{fig:garden-path-biological} illustrates the behavior of the baseline model compared to
the best found surprisal-based recoding model (using a fixed step size of $5$) and depicts the surprisal scores over the course of a sentence, using samples from the garden path corpus.\\

Here we can make the following observations: Both models highly correlate in their surprisal, while this specific recoding model (orange curve) achieves slightly lower scores than the baseline (blue curve) most of the time. Furthermore, we can see in this instance that the surprisal scores improve or stay the same after recoding (data points between words), often in a statistically significant manner (gray triangular markers). There are examples where the baseline model produces lower scores (right plot, ``expensive''), yet these are far outnumbered by the reverse case in which the recoding model succeeds. Using the more dynamic step size approaches also did not influence these observations (not shown here).\\

However, all tested models only show a limited degree of cognitive plausibility. While it is difficult to directly constrast these results with human reading time\footnote{No direct reading time data is available and the setup often differs in the human case, i.e.\ people reading the sentences in chunks instead of a word-by-word basis.}, we would intuitively assume for the model's surprisal score to spike once they encounter the first word contradicting the sentences' first parse assumed by a human. In the first example, this is somewhat the case, as the models display higher surprisal when processing ``the'', which changes the preceeding ``man' from being the subject of the sentences with the attribute ``old'' to the former being the main verb and the latter to the subject (as in ``the old [people]''). In the second more complex sentence, ``looked'' gives the decisive clue that ``the enourmous car'' is not the object of the construction ``the man polished'' but indeed the subject of the main clause. Even though humans would be expected to take some more time to resolve this new-found ambiguity, the models actually emit \emph{lower} surprisal scores.\\

Other works have already pointed out some limits of relating the results of Language Models on garden path sentences to processes in the human cognitive system: \cite{van2018modeling} find that the reading time estimates produced by \glspl{LSTM} depend on the amount of training data and are incapable of completely modeling the processing time required by people. A similar observation is noted by \cite{futrell2019neural}, because their model trained on \gls{PTB} displays significantly smaller garden path effects than instances trained on larger data sets. Overall, we can conclude that the similarity of the models with respect to human processing is limited in two regards: First, by the syntactic abilities of \glspl{LSTM} (\glspl{LSTM} were shown to learn a syntactic state - as described e.g.\ in chapter \ref{sec:related-work-rnn-language} - but it might insufficient) and secondly the grammatical knowledge induced simply by the amount of training data, which is small compared to most other Deep Learning data sets.

\subsection{Error signal reduction}\label{sec:exp-error-reduction}

Now let us focus entirely on the error signal $\delta_t$. By plotting its development during
inference time several points about the methodology of this work can be verified: On one hand,
that recoding under the right choice of step size reduces the error signal $\delta_t$ (cp. theorem \ref{theorem:error-reduction}) and the error signal at a later time step (cp. theorem \ref{theorem:error-reduction-tt}), even when no additional recoding is used. On the other hand, this likewise gives an opportunity to analyze whether a suboptimal choice of step size actually leads to an \emph{increase} of the error signal, breaking the assumptions of both theorems with respect to the error function's Lipschitz constant.\\

\begin{figure}[h]
  \centering
  \begin{tabular}{cc}
    \includegraphics[width=0.495\textwidth]{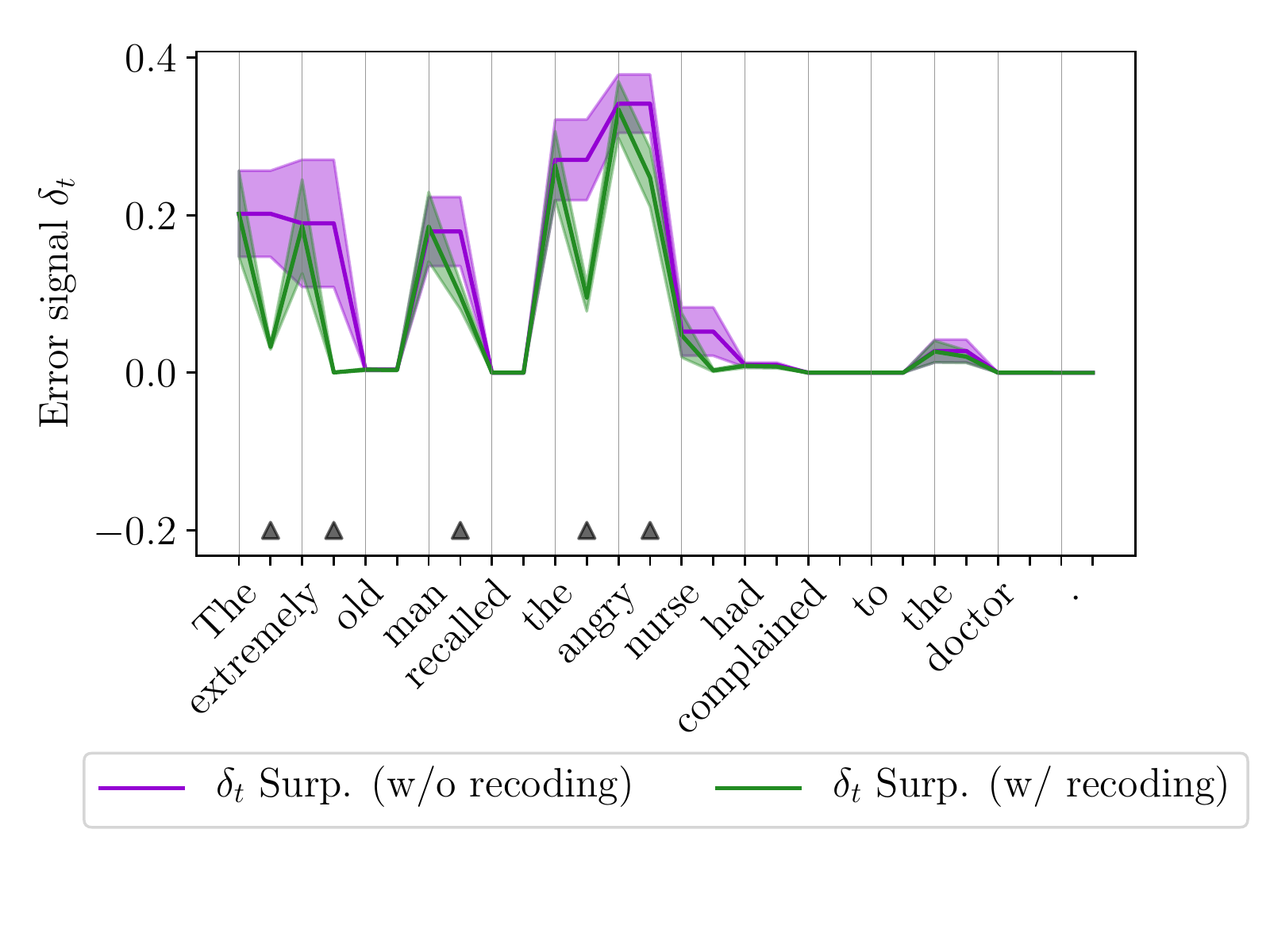} &
    \includegraphics[width=0.495\textwidth]{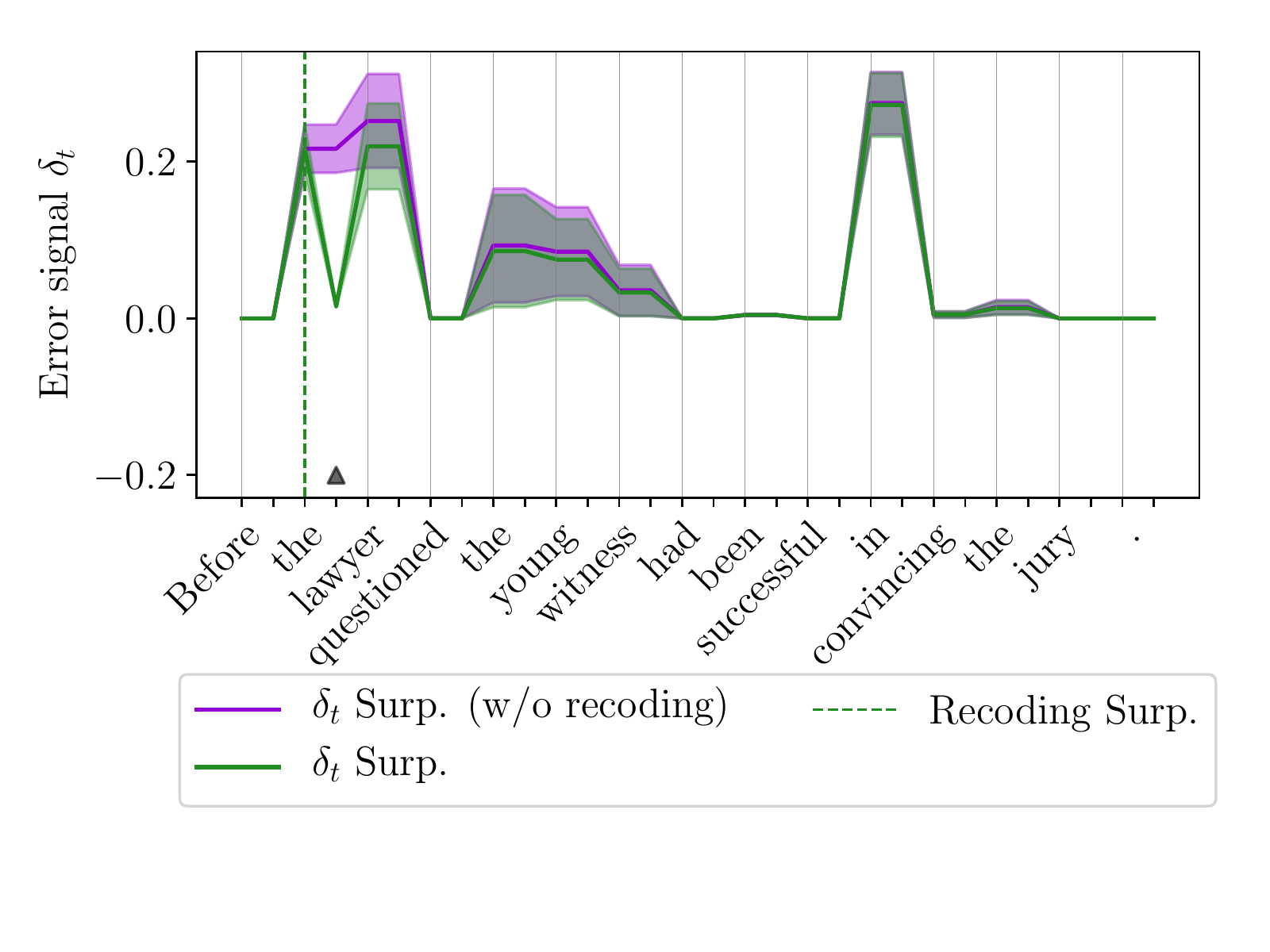} \\
  \end{tabular}
  \caption[Error reduction through recoding on garden path sentences]{Error reduction through recoding on garden path sentences for the surprisal-based recoding model with fixed step size for all time steps (left) and a single time step (right; dashed line). Curves denote the mean of $n=4$ runs, while the intervals depict one standard deviation. Gray markers denote statistically significant differences (two-tailed Student's t-test, $p=0.05$).}\label{fig:qualtitative-error-reduction}
\end{figure}

The following analysis focuses on the best observed surprisal-based recoding model.\footnote{Due to the stochasticity involved in \gls{MCD} a fair comparison of error signals for the same sequence cannot be guaranteed and the explanatory power of \gls{BAE} results is questionable.} In the left plot of figure \ref{fig:qualtitative-error-reduction}, it can be observed that recoding indeed consistenly reduced the error signal compared to the same model where the signal is recoded, but the recoding update step not performed. Although not impossible, it was also not observed in the given data set that recoding \emph{increased} the error signal, it at most stagnates after the activation adaption. In the right plot, another sentence is used to illustrate the effect of only recoding at time step $t=2$: Because recoding only takes place once, the difference between the surprisal scores fades over time, however, an offset is still noticable even after the time step in question, which is likely to be explained by the findings of theorem \ref{theorem:error-reduction-tt}. In many cases however, the effect afterwards is minimal or not noticeable, which can also be explained by the same theorem, which showed that the possible improvement becomes smaller by distance to the modification made by recoding. Lemon-picked examples of this outcome and more positive samples of both analyses are given in appendix \ref{fig:qualtitative-additional-error-reduction}.

\subsection{Effect of overconfident predictions}\label{sec:overconfident}

\begin{figure}[h]
  \centering
  \begin{tabular}{cc}
    \includegraphics[width=0.495\textwidth]{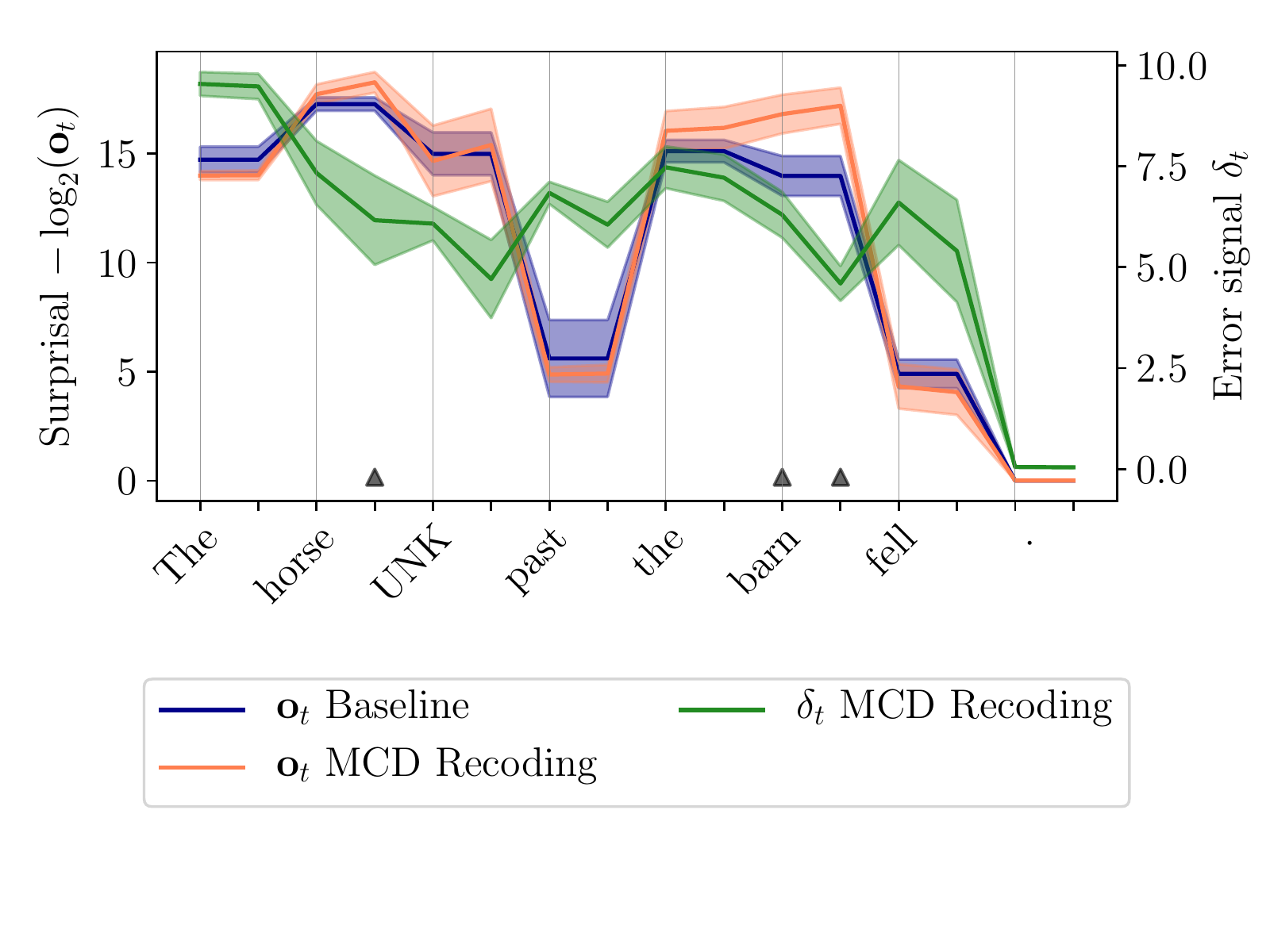} &
    \includegraphics[width=0.495\textwidth]{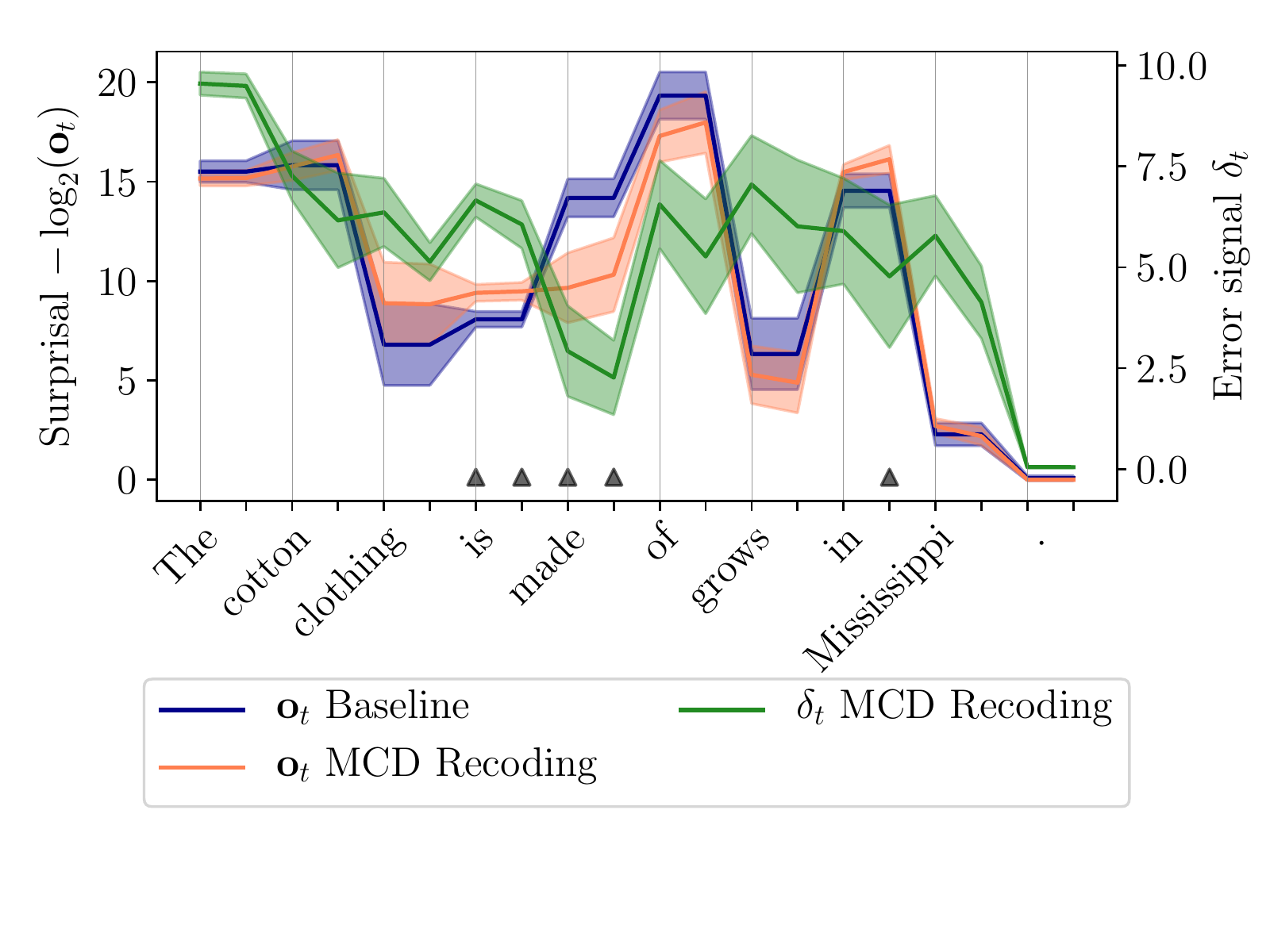} \\
  \end{tabular}
  \caption[The effect of uncertainty recoding on surprisal]{The effect of uncertainty recoding with \gls{MCD} on surprisal, using a fixed step size of $5$ and $K=10$ samples. Curves denote the mean of $n=4$ runs, while the intervals depict one standard deviation. Gray markers denote statistically significant differences in surprisal (two-tailed Student's t-test, $p=0.05$).}\label{fig:qualtitative-overconfident}
\end{figure}

In the previous chapter, we observed that the recoding models using approximate predictive entropy as their error signals actually performed significantly \emph{worse} than the baseline. The ablation study in chapter \ref{sec:ablation} revealed that this can at least partially be attributed to the fact that using those recoding mechanisms produces suboptimal weights (even without the mechanism, these models performed worse than the baseline). Nevertheless, another suspected reason for this discrepancy is that the confidence of the model approximated by \gls{MCD} and \gls{BAE} might lead to overconfident \emph{mis}-predictions, leading the model astray after the correction step. In contrast to surprisal recoding, these two variants do not have knowledge about the gold token and only incentivize the model to concentrate their probability around their most probable predictions following the update step. To investigate this suspicion, some plots showing the development of the error signal jointly with the model surprisal scores compared to the baseline in figure \ref{fig:qualtitative-overconfident}. Because the range of estimated predictive entropy scores is much lower on the garden path data set than on PTB, the fixed step $=0.1$ \gls{MCD} recoding model from chapter \ref{sec:exp-step-size} is reused.\\

Two observations can be made from both plots in fig. \ref{fig:qualtitative-overconfident}: First, we can see that recoding lowers the error signal $\delta_t$, validating theorem \ref{theorem:error-reduction} for this approach as well. In spite of this, the surprisal scores are \textbf{not} necessarily lowered by this. In the right plot, it does happen for the time step of the word ``grow'', but in many other instances for the shown samples, the opposite occurs: In the left plot, recoding the activations at the time step of ``horse'' lowers predictive entropy but increases surprisal such that the gap to the baseline's score becomes statistically significant. In the case of ``barn'' for the same sentence, the difference is exacerbated even further. This provides some evidence to the hypothesis that this behavior might contribute to the worse performance of \gls{MCD} and \gls{BAE} compared to the baseline. However, it is unclear whether this is an additional factor to the suboptimal weights identified in chapter \ref{sec:ablation} or actually the culprit behind this very same observation.

\chapter{Discussion}\label{chapter:discussion}

\epigraph{``The more [the scientist] analyzes the universe into infinitesimals, the more things he finds to classify, and the more he perceives the relativity of all classification. What he does not know seems to increase in geometric progression to what he knows.
"}{\vspace{0.4cm}\emph{The Wisdom of Insecurity - Alan W. Watts}}

The previous two chapters have tried to analyze the models and their operation from many angles. But even with all the experiments that have been conducted, this work cannot answer all questions that might be raised on the basis of the supplied results conclusively. Therefore I aggregate all the findings to far
and try to distill some general conclusions from them while also pointing out
some of the limits identified thus far.

\section{Step Size}\label{sec:discussion-step-size}

Many of the experiments in chapter \ref{chapter:experiments} were dedicated to determine the effect of step size on the recoding models. The step seems like one of the core pieces of recoding, as an improvement of the error signal can only guaranteed using the right size under theorem \ref{theorem:error-reduction}. The results show that an advantageous choice of step size heavily depends on the value range of the error signal and possibly also on the data set, as e.g.\ uncertainty estimates in chapter \ref{sec:overconfident} on the garden path data set differed from PTB. It also seems like a simplification to assume that all layers and activations types require the same constant step size at every point (the magnitude of correction by recoding probably should depend on the circumstances of the input).\\

To this end, it appears like a straightforward choice to parameterize the step size as a single learned parameter or by some secondary model, however the results from corresponding experiments in chapter \ref{sec:learned-step-size} were inconclusive. For different models and different error signals, found solutions preferred to correct activations in an inconsistent hierarchical order when sorted by size. It is not certain which type of activation has to be recoded most to achieve the best improvement. Besides, the resulting models performed worse than their counterparts using constant update steps. It hence remains unclear to which extent a constant step size and having the magnitude of the recoding solely determined by the strength of the recoding gradient $\nabla_{\bh_t}\delta_t$ suffices. Perhaps the learning of step sizes requires more exploration of regularization and hyperparameter search in the predicted case which lie outside the scope of this work.

\section{Role of Error Signal}\label{sec:role-error-signals}

The other moving part of the recoding framework is the choice of error signal $\delta_t$, which also provides
many points for discussion.\\

The key problem seems to be to identify an error signal that improves the objective, i.e.\ is ``conjugate''. This cannot be said for predictive entropy: Both quantitative and qualitative analysis in the two previous chapters have shown that these approaches, at least using \gls{MCD} and \gls{BAE}, result in worse weigths and confident mispredictions, resulting in higher perplexity than the baseline. These results underscore especially the difficulty of using unsupervised signals. In the original case of interventions \citep{giulianelli2018under}, replication experiments conducted in the context of this worked showed that these corrections also had a statistically non-significant effect on perplexity. Even other works that improved drastically on the results of the number prediction task could not identify improvements in perplexity \citep{merrill-etal-2019-finding}, which seems to suggest that this information does not improve the model's performance, either.\\

Even in the case of surprisal, which seems directly related to the cross-entropy objective as it involves the probability of the target token, only achieved minor improvements on the baseline in the best case. This could be explained by two reasons: Either the network does not profit from being supplied with the same learning signal twice (gold token probability during recoding \textbf{and} for the loss calculation) or the main source of error in a recurrent setting is not remedied by recoding. The process may correct corrupted information about the right prediction encoded in the hidden activations and therefore rectify the horizontal information flow in the RNN, but it does not influence any errors commited by the decoding layer of the LM itself or the insufficient encoding of the current input $w_t$ by the embeddings and inside the hidden layer. That this vertical direction could be the source of many other model errors might explain the limited improvement attained by the best recoding model.

\section{Approximations \& Simplifying Assumptions}

Moreover, I want to discuss the influence of approximations and simplifications on the results obtained in previous chapters.

\textbf{Approximation Quality of Predictive Entropy} In the case of \gls{MCD} recoding, the quality of the approximation of the model's predictive entropy relies on the number of samples and the MC Dropout rate. Although both factors seemed to not impact the performance to strongly (cp. chapters \ref{sec:exp-num-samples} and \ref{sec:exp-dropout-rate}), the results should be taken with a grain of salt, as the sample size of $n=4$ for different models runs was fairly small, although necessary due to enormous amount of models that had to be trained across all experiments. It has further been shown in chapter \ref{sec:exp-num-samples} that anchored ensembles do not produce correct uncertainty estimates when using the amortized training procedure. This simplification was paramount to make training of a decoder ensemble tractable in a recurrent setting; otherwise $K$ different weight gradients have to be computed at the end of every batch, which backpropagate back through the entire sequence. As the experiments elucidated, this produces inconsistent estimates and seems to break the theoretical guarantee layed out in appendix \ref{appendix:pred-entropy}, which might explain the loss in performance compared to MC Dropout recoding.
Recently, a faster method for ensembling derived from the topology of loss surfaces has been proposed \citep{garipov2018loss}, but lacks a Bayesian foundation and an extension to \glspl{RNN}. Until a scalable, Bayesian ensembling method in a recurrent setting is developed, this approach seems to be deemed to be outperformed by the other recoding model variants.

\textbf{LM Decoders as Independent Models} In order to transfer the ideas of MC Dropout and Bayesian Anchored Ensembling that apply to \glspl{MLP} to a \gls{RNN} setting, the simplifying assumptions was made to see the decoder layer of the Language Model as a sort of ``mini-\gls{MLP}''. While this allowed for an application of these techniques, it essentially results in only modeling the uncertainty of the decoder at a certain time step. For a recurrent model, we would like to model the uncertainty of the \emph{whole model}, which depends on previous time steps. \cite{gal2016theoretically} provides an adaptation of MC Dropout to \glspl{RNN}, yet preliminary experiments showed that this holistic solution is problematic in the recoding framework for two reasons: Ensembling multiple \glspl{RNN} using distinct sets of dropout masks for every type of connection is memory-intense and becomes computationally prohibitive in connection with additional gradient calculations at every time step.

\chapter{Conclusion}\label{chapter:conclusion}

\epigraph{\hfill \begin{CJK}{UTF8}{gbsn}
 \textbf{沧海桑田}
\end{CJK} - The blue sea turned into mulberry fields}{\vspace{0.4cm}\emph{Chinese idiom about the transformative changes in the world}}

The recoding framework proposed in this thesis comes with several theoretical guarantees and an appealing intuition. In spite of this, experimental results have shown that there are several caveats to the approach which make the application in practice challenging: The computation of the gradient imposes additional computational costs which are not justified by the results and only yield minor improvements in the best case. They also do not improve on the cognitive plausibility of the model. This shows that the most simple instantiation of the recoder is not sufficient to deliver on its promises. Nevertheless, this opens up further lines of research which are reflected on in the next section. The last part of this chapter, section \ref{sec:outlook}, is withal dedicated to put this work into the greater context of current AI research trends.

\section{Future Work}

Nonwithstanding the analyses in previous chaprers, the results should not be seen as the method's eulogy but as a first step that paves the road to future research ideas, some of which will be discussed here. The following sections hence deal with the issues of recoding signals (\S \ref{sec:future-recoding-signals}), step size (\S \ref{sec:future-step-size}), efficiency (\S \ref{sec:future-efficiency}) and possible modifications of training and architecture (\S \ref{sec:future-trainining}).

\subsection{Recoding signals}\label{sec:future-recoding-signals}

This work explored three different kinds of recoding signals. While these were not able to produce any substantial improvements compared to the baseline, this does not have to be the case with other choices. Chapter \ref{sec:role-error-signals} discussed the shortcomings of the explored options (surprisal, predictive entropy). Other error signals could be tested on a toy data set or proven mathematically to improve the objective function.\footnote{This could e.g.\ be achieved by proving that there is a constant linear relationship between the error signals and the loss function - therefore a reduction of the error signal like derived in theorem \ref{theorem:error-reduction} will linearly decrease the loss by the same factor.} But it is also imaginable to use recoding to maintain certain desirable properties in the hidden activations - after all, the loss function only serves as proxy for the \emph{real objective}, similar to how Neural Machine Translation models might be trained using cross-entropy loss but are actually supposed to optimize BLEU or another, similar metric. Recoding could help here to retain certain linguistic features that are not considered during optimization but that improve some property of the translation. Apart from additional supervision, there is also a large space of unsupervised signals to explore, although my experiments have shown that this can turn out to be challenging.

\subsection{Step size}\label{sec:future-step-size}

Other research work could focus on the issue of step size by exploring other choices of parameterization and perform more expressive studies about the issue. In this work, one choice tried to predict the step size based on the current hidden activations. Other models could try to find a better solution by basing the prediction on the output activations or completely different features.\\

The step size could also be kept constant relative to the properties to the model. The improvement of the error signal is only guaranteed by the choice of the smallest Lipschitz constant $L$ of the error function that produces the signal. This function maps the hidden activations to some signal and is parameterized by the decoder layer weights. Although the Lipschitz value of this function changes every time the parameters are updated, it is conceivable to compute this quantity to avoid misguided updates. Albeit recent work has shown that computing the exact Lipschitz constant is a NP-hard problem even in shallow neural networks, multiple other efforts have been  dedicated to approximating it \citep{virmaux2018lipschitz, oberman2018lipschitz, gouk2018regularisation}. In this way, improvement would be guaranteed (up to the precision of the approximation) and the magnitude of the correction solely dependent on the recoding gradient.\\

\subsection{Efficiency}\label{sec:future-efficiency}

One major drawback of the proposed method seems to lie in its efficiency. In table \ref{table:exp-num-samples-results} it is noted that the models using predictive entropy as their recoding signal process inputs dramatically slower than the baseline. The disparity is further exacerbated when increasing the sample / ensemble size of the models. But even for the surprisal recoding model, a significant slow-down is observed based on the additional backward passes: Even though automatic differentiation engines like used in \verb|PyTorch| are heavily optimized, the gradient calculation takes up a large portion of the training time. The recoder enlarges that portion, where the increase grows linearly with sequence length and network depth.\\

Based on the increasing resource requirements for state-of-the-art models, a recent position paper \citep{schwartz2019green} has advocated to make efficiency an additional evaluation criterion of models. From this point of view, future research should evaluate whether the current model formulation could be
improved by e.g.\ approximating the recoding gradient with the finite difference method. This has already successfully been applied in other approaches with gradient-heavy computations like differentiable architecture search \citep{liu2018darts}.\\

\subsection{Training Procedure \& Architecture}\label{sec:future-trainining}

In the tested models, the recoder was combined with the \gls{LSTM} Language Model and jointly trained in the most simple way. To reduce the computational load, it could be sensible to integrate the recoder as some sort of special gate that is only used in specific circumstances. In this thesis, only modifications of the current hidden state were discussed; yet the recoder could also be extended to recode activations further back in the sequence, although this likely to result in additional computational costs. Inspiration for a solution could possibly be drawn from the work from \cite{ke2018sparse}, who use a self-attention mechanism for sparse credit assignment instead of regular backpropagation through time to direct the gradient flow.\\

Given that joint training also seemed to produce suboptimal weights in the case of predictive entropy recoding, it could also be imagined to train the model using a variation of \emph{teacher forcing} like commonly used in Neural Machine Translation, where the recoder is only used in a certain percentage of batches or time steps. Another possibility is to train the base model independently in some pretraining phase and add the recoder afterwards or exclusively during inference (the ablation study in chapter \ref{sec:ablation} showed that adding the surprisal recoder to the baseline during inference resulted in small improvements).

\section{Outlook}\label{sec:outlook}

This work is only a small piece in an incomprehensibly big
landscape of Artificial Intelligence and Machine Learning research. As such, I would like to dedicate the last part of this thesis to embed it into a larger frame of reference, which points out possible research directions happening on a larger scale.

\subsection{Innotivation over Scale}

Taking the position of a devil's advocat, someone might argue: ``What is the value of models that do not outperform
the current state-of-the-art?'' or ''Why try x when model y already produces very good results on this task?''.
Indeed, especially in Natural Language Processing a line of recent works based on the transformer architecture by \cite{vaswani2017attention} has established new state-of-the-art performances on Language Modeling.\\

These include Google's BERT \citep{devlin2019bert}, OpenAI's GPT-2 \citep{radford2019language} and XLNet \citep{yang2019xlnet}. On the flip side, these models require huge amounts of data and computional resources. The training of XLNet has been estimated to have cost about than 61.000 \$.\footnote{Source: \url{https://syncedreview.com/2019/06/27/the-staggering-cost-of-training-sota-ai-models/} (04.08.19).} This has several, worrying implications:
First, with these resource requirements, scientific papers become hard to reproduce. These costs only allow training of these models in the context of well-funded insitutions, namely top-tier universities and affluent tech giants. Secondly, the reliance on large-scale hardware produces a high electricity consumption along with a worrysome carbon footprint, which bears a certain irony: These models try to (loosely) imitate the human brain, a biological computer that is actually very energy efficient \citep{schwartz2019green}. Lastly, scaling up data sets and the number of parameters does not necessarily increase the semblance to human cognition: A recent paper found that BERT's almost human-level performance in an argument reasoning Comprehension task was due to spurious statistical features in the data set \citep{niven2019probing}.\\

All these reasons are reiterated to make the following point: While these models have tremendously propelled research and helped established new cutting-edge results on different tasks, squeezing out a few more performance points by scaling up should not be the singular objective of research. Instead, there is merit in experimenting with completely different architectures - like attempted in this work. This might produce innovations to help make models process data better and use their parameters more efficiently, which in turn can thus produce more biologically plausible and energy-efficient models. And to get there, we can follow a model that already displays the desired behavior.

\subsection{Inspiration from Human Cognition}

The human brain is a marvelous object. It comprises 86 billion neurons \citep{azevedo2009equal}, is able to perform all its complex computations while only consuming around 12.6 watts of energy per day.\footnote{Source: \url{https://www.scientificamerican.com/article/thinking-hard-calories/} (04.08.19).} While scientists understand more and more about its modus operandi, their insights have provided a continuous stream of inspiration for AI research and vice versa, making the two fields historically entangled: From the beginnings of Deep and Reinforcement Learning to many current ideas build on these insights, an overview of which is given in \cite{hassabis2017neuroscience}. Given the line of reasoning induced in the previous section, I would like to end my thesis on the following note: Notwithstanding the rapid development of and impressive milestones achieved by Artificial Intelligence in recent years, the field has still a long way to go to actually achieve human-level intelligence. It was advocated to bridge this gap by equipping models with some of the innate cognitive properties of humans \citep{lake2017building}. And although it is undeniable that moving away from the biological example will always yield important results for research as well, this work should be contextualized as one small attempt to make progress by drawing inspiration from the one thing that we know proven to be a working example of general intelligence: Ourselves.

{%
\setstretch{1.1}
\renewcommand{\bibfont}{\normalfont\small}
\setlength{\biblabelsep}{0pt}
\setlength{\bibitemsep}{0.5\baselineskip plus 0.5\baselineskip}
\printbibliography

@inproceedings{giulianelli2018under,
  title={Under the Hood: Using Diagnostic Classifiers to Investigate and Improve how Language Models Track Agreement Information},
  author={Giulianelli, Mario and Harding, Jack and Mohnert, Florian and Hupkes, Dieuwke and Zuidema, Willem},
  booktitle={Proceedings of the 2018 EMNLP Workshop BlackboxNLP: Analyzing and Interpreting Neural Networks for NLP},
  pages={240--248},
  year={2018}
}

@article{rumelhart1988learning,
  title={Learning representations by back-propagating errors},
  author={Rumelhart, David E and Hinton, Geoffrey E and Williams, Ronald J and others},
  journal={Cognitive modeling},
  volume={5},
  number={3},
  pages={1},
  year={1988}
}

@article{fernandez2009novel,
  title={A novel connectionist system for improved unconstrained handwriting recognition},
  author={Fernandez, A Graves M Liwicki S and Bunke, R Bertolami H and Schmiduber, J},
  journal={IEEE Transactions on Pattern Analysis and Machine Intelligence},
  volume={31},
  number={5},
  year={2009}
}

@inproceedings{niven2019probing,
  title={Probing Neural Network Comprehension of Natural Language Arguments},
  author={Niven, Timothy and Kao, Hung-Yu},
  booktitle={Proceedings of the 57th Conference of the Association for Computational Linguistics},
  pages={4658--4664},
  year={2019}
}

@inproceedings{liu2018darts,
  title={{DARTS}: Differentiable Architecture Search},
  author={Hanxiao Liu and Karen Simonyan and Yiming Yang},
  booktitle={International Conference on Learning Representations},
  year={2019},
  url={https://openreview.net/forum?id=S1eYHoC5FX},
}

@article{hassabis2017neuroscience,
  title={Neuroscience-inspired artificial intelligence},
  author={Hassabis, Demis and Kumaran, Dharshan and Summerfield, Christopher and Botvinick, Matthew},
  journal={Neuron},
  volume={95},
  number={2},
  pages={245--258},
  year={2017},
  publisher={Elsevier}
}

@inproceedings{garipov2018loss,
  title={Loss surfaces, mode connectivity, and fast ensembling of dnns},
  author={Garipov, Timur and Izmailov, Pavel and Podoprikhin, Dmitrii and Vetrov, Dmitry P and Wilson, Andrew G},
  booktitle={Advances in Neural Information Processing Systems},
  pages={8789--8798},
  year={2018}
}

@article{oberman2018lipschitz,
  author    = {Adam M. Oberman and
               Jeff Calder},
  title     = {Lipschitz regularized Deep Neural Networks converge and generalize},
  journal   = {CoRR},
  volume    = {abs/1808.09540},
  year      = {2018},
  url       = {http://arxiv.org/abs/1808.09540},
  archivePrefix = {arXiv},
  eprint    = {1808.09540},
  bibsource = {dblp computer science bibliography, https://dblp.org}
}

@article{gouk2018regularisation,
  title={Regularisation of Neural Networks by Enforcing Lipschitz Continuity},
  author={Gouk, Henry and Frank, Eibe and Pfahringer, Bernhard and Cree, Michael J},
  journal={stat},
  volume={1050},
  pages={14},
  year={2018}
}

@inproceedings{ke2018sparse,
  title={Sparse attentive backtracking: Temporal credit assignment through reminding},
  author={Ke, Nan Rosemary and GOYAL, Anirudh Goyal ALIAS PARTH and Bilaniuk, Olexa and Binas, Jonathan and Mozer, Michael C and Pal, Chris and Bengio, Yoshua},
  booktitle={Advances in Neural Information Processing Systems},
  pages={7640--7651},
  year={2018}
}

@inproceedings{merrill-etal-2019-finding,
    title = "Finding Hierarchical Structure in Neural Stacks Using Unsupervised Parsing",
    author = "Merrill, William  and
      Khazan, Lenny  and
      Amsel, Noah  and
      Hao, Yiding  and
      Mendelsohn, Simon  and
      Frank, Robert",
    booktitle = "Proceedings of the 2019 ACL Workshop BlackboxNLP: Analyzing and Interpreting Neural Networks for NLP",
    month = aug,
    year = "2019",
    address = "Florence, Italy",
    publisher = "Association for Computational Linguistics",
    url = "https://www.aclweb.org/anthology/W19-4823",
    pages = "224--232",
}

@article{sturt1999structural,
  title={Structural change and reanalysis difficulty in language comprehension},
  author={Sturt, Patrick and Pickering, Martin J and Crocker, Matthew W},
  journal={Journal of Memory and Language},
  volume={40},
  number={1},
  pages={136--150},
  year={1999},
  publisher={Elsevier}
}

@inproceedings{van2018modeling,
  title={Modeling garden path effects without explicit hierarchical syntax.},
  author={Van Schijndel, Marten and Linzen, Tal},
  booktitle={CogSci},
  year={2018}
}

@article{lake2017building,
  title={Building machines that learn and think like people},
  author={Lake, Brenden M and Ullman, Tomer D and Tenenbaum, Joshua B and Gershman, Samuel J},
  journal={Behavioral and brain sciences},
  volume={40},
  year={2017},
  publisher={Cambridge University Press}
}

@article{azevedo2009equal,
  title={Equal numbers of neuronal and nonneuronal cells make the human brain an isometrically scaled-up primate brain},
  author={Azevedo, Frederico AC and Carvalho, Ludmila RB and Grinberg, Lea T and Farfel, Jos{\'e} Marcelo and Ferretti, Renata EL and Leite, Renata EP and Filho, Wilson Jacob and Lent, Roberto and Herculano-Houzel, Suzana},
  journal={Journal of Comparative Neurology},
  volume={513},
  number={5},
  pages={532--541},
  year={2009},
  publisher={Wiley Online Library}
}

@misc{schwartz2019green,
    title={Green AI},
    author={Roy Schwartz and Jesse Dodge and Noah A. Smith and Oren Etzioni},
    year={2019},
    eprint={1907.10597},
    archivePrefix={arXiv},
    primaryClass={cs.CY}
}

@inproceedings{sak2014long,
  title={Long short-term memory recurrent neural network architectures for large scale acoustic modeling},
  author={Sak, Ha{\c{s}}im and Senior, Andrew and Beaufays, Fran{\c{c}}oise},
  booktitle={Fifteenth annual conference of the international speech communication association},
  year={2014}
}

@article{linzen2016assessing,
  title={Assessing the ability of LSTMs to learn syntax-sensitive dependencies},
  author={Linzen, Tal and Dupoux, Emmanuel and Goldberg, Yoav},
  journal={Transactions of the Association for Computational Linguistics},
  volume={4},
  pages={521--535},
  year={2016},
  publisher={MIT Press}
}

@inproceedings{gulordava2018colorless,
  author    = {Kristina Gulordava and
               Piotr Bojanowski and
               Edouard Grave and
               Tal Linzen and
               Marco Baroni},
  title     = {Colorless Green Recurrent Networks Dream Hierarchically},
  booktitle = {Proceedings of the 2018 Conference of the North American Chapter of
               the Association for Computational Linguistics: Human Language Technologies,
               {NAACL-HLT} 2018, New Orleans, Louisiana, USA, June 1-6, 2018, Volume
               1 (Long Papers)},
  pages     = {1195--1205},
  year      = {2018},
  crossref  = {DBLP:conf/naacl/2018-1},
  url       = {https://aclanthology.info/papers/N18-1108/n18-1108},
  bibsource = {dblp computer science bibliography, https://dblp.org}
}

@article{bergstra2012random,
  title={Random search for hyper-parameter optimization},
  author={Bergstra, James and Bengio, Yoshua},
  journal={Journal of Machine Learning Research},
  volume={13},
  number={Feb},
  pages={281--305},
  year={2012}
}

@article{bernardy2018can,
  title={Can Recurrent Neural Networks Learn Nested Recursion?},
  author={Bernardy, Jean-Philippe},
  journal={LiLT (Linguistic Issues in Language Technology)},
  volume={16},
  number={1},
  year={2018}
}

@article{srivastava2014dropout,
  title={Dropout: a simple way to prevent neural networks from overfitting},
  author={Srivastava, Nitish and Hinton, Geoffrey and Krizhevsky, Alex and Sutskever, Ilya and Salakhutdinov, Ruslan},
  journal={The Journal of Machine Learning Research},
  volume={15},
  number={1},
  pages={1929--1958},
  year={2014},
  publisher={JMLR. org}
}

@inproceedings{zhu2017deep,
  title={Deep and confident prediction for time series at uber},
  author={Zhu, Lingxue and Laptev, Nikolay},
  booktitle={2017 IEEE International Conference on Data Mining Workshops (ICDMW)},
  pages={103--110},
  year={2017},
  organization={IEEE}
}

@inproceedings{gal2016theoretically,
  title={A theoretically grounded application of dropout in recurrent neural networks},
  author={Gal, Yarin and Ghahramani, Zoubin},
  booktitle={Advances in neural information processing systems},
  pages={1019--1027},
  year={2016}
}

@article{pearce2018uncertainty,
  title={Uncertainty in Neural Networks: Bayesian Ensembling},
  author={Pearce, Tim and Zaki, Mohamed and Brintrup, Alexandra and Neely, Andy},
  journal={stat},
  volume={1050},
  pages={12},
  year={2018}
}

@inproceedings{nagabandi2018learning,
  author    = {Anusha Nagabandi and
               Ignasi Clavera and
               Simin Liu and
               Ronald S. Fearing and
               Pieter Abbeel and
               Sergey Levine and
               Chelsea Finn},
  title     = {Learning to Adapt in Dynamic, Real-World Environments through Meta-Reinforcement
               Learning},
  booktitle = {7th International Conference on Learning Representations, {ICLR} 2019,
               New Orleans, LA, USA, May 6-9, 2019},
  year      = {2019},
  crossref  = {DBLP:conf/iclr/2019},
  url       = {https://openreview.net/forum?id=HyztsoC5Y7},
  bibsource = {dblp computer science bibliography, https://dblp.org}
}

@inproceedings{dai2019transformerxl,
  author    = {Zihang Dai and
               Zhilin Yang and
               Yiming Yang and
               Jaime G. Carbonell and
               Quoc Viet Le and
               Ruslan Salakhutdinov},
  title     = {Transformer-XL: Attentive Language Models beyond a Fixed-Length Context},
  booktitle = {Proceedings of the 57th Conference of the Association for Computational
               Linguistics, {ACL} 2019, Florence, Italy, July 28- August 2, 2019,
               Volume 1: Long Papers},
  pages     = {2978--2988},
  year      = {2019},
  crossref  = {DBLP:conf/acl/2019-1},
  url       = {https://www.aclweb.org/anthology/P19-1285/},
  bibsource = {dblp computer science bibliography, https://dblp.org}
}

@inproceedings{bahdanau2014neural,
  author    = {Dzmitry Bahdanau and
               Kyunghyun Cho and
               Yoshua Bengio},
  title     = {Neural Machine Translation by Jointly Learning to Align and Translate},
  booktitle = {3rd International Conference on Learning Representations, {ICLR} 2015,
               San Diego, CA, USA, May 7-9, 2015, Conference Track Proceedings},
  year      = {2015},
  crossref  = {DBLP:conf/iclr/2015},
  url       = {http://arxiv.org/abs/1409.0473},
  bibsource = {dblp computer science bibliography, https://dblp.org}
}

@inproceedings{blundell2015weight,
  title={Weight Uncertainty in Neural Network},
  author={Blundell, Charles and Cornebise, Julien and Kavukcuoglu, Koray and Wierstra, Daan},
  booktitle={International Conference on Machine Learning},
  pages={1613--1622},
  year={2015}
}

@article{rosenblatt1958perceptron,
  title={The perceptron: a probabilistic model for information storage and organization in the brain.},
  author={Rosenblatt, Frank},
  journal={Psychological review},
  volume={65},
  number={6},
  pages={386},
  year={1958},
  publisher={American Psychological Association}
}

@book{jurafsky2000speech,
  title={Speech \& language processing},
  author={Jurafsky, Dan},
  year={2000},
  publisher={Pearson Education India}
}

@article{bengio2003neural,
  title={A neural probabilistic language model},
  author={Bengio, Yoshua and Ducharme, R{\'e}jean and Vincent, Pascal and Jauvin, Christian},
  journal={Journal of machine learning research},
  volume={3},
  number={Feb},
  pages={1137--1155},
  year={2003}
}

@article{buchanan2005very,
  title={A (very) brief history of artificial intelligence},
  author={Buchanan, Bruce G},
  journal={Ai Magazine},
  volume={26},
  number={4},
  pages={53--53},
  year={2005}
}

@misc{minsky1969perceptrons,
  title={Perceptrons. An Introduction to Computational Geometry. 1969, Expanded},
  author={Minsky, ML and Papert, SA},
  year={1969},
  publisher={Cambridge, MA: MIT Press}
}

@article{csaji2001approximation,
  title={Approximation with artificial neural networks},
  author={Cs{\'a}ji, Bal{\'a}zs Csan{\'a}d},
  journal={Faculty of Sciences, Etvs Lornd University, Hungary},
  volume={24},
  pages={48},
  year={2001},
  publisher={Citeseer}
}

@inproceedings{edunov2018understanding,
  author    = {Sergey Edunov and
               Myle Ott and
               Michael Auli and
               David Grangier},
  title     = {Understanding Back-Translation at Scale},
  booktitle = {Proceedings of the 2018 Conference on Empirical Methods in Natural
               Language Processing, Brussels, Belgium, October 31 - November 4, 2018},
  pages     = {489--500},
  year      = {2018},
  crossref  = {DBLP:conf/emnlp/2018},
  url       = {https://aclanthology.info/papers/D18-1045/d18-1045},
  bibsource = {dblp computer science bibliography, https://dblp.org}
}

@article{van2016wavenet,
  title={WaveNet: A generative model for raw audio.},
  author={Van Den Oord, A{\"a}ron and Dieleman, Sander and Zen, Heiga and Simonyan, Karen and Vinyals, Oriol and Graves, Alex and Kalchbrenner, Nal and Senior, Andrew W and Kavukcuoglu, Koray},
  journal={SSW},
  volume={125},
  year={2016}
}

@article{evans2018novo,
  title={De novo structure prediction with deeplearning based scoring},
  author={Evans, R and Jumper, J and Kirkpatrick, J and Sifre, L and Green, TFG and Qin, C and Zidek, A and Nelson, A and Bridgland, A and Penedones, H and others},
  journal={Annu Rev Biochem},
  volume={77},
  pages={363--382},
  year={2018}
}

@article{silver2016mastering,
  title={Mastering the game of Go with deep neural networks and tree search},
  author={Silver, David and Huang, Aja and Maddison, Chris J and Guez, Arthur and Sifre, Laurent and Van Den Driessche, George and Schrittwieser, Julian and Antonoglou, Ioannis and Panneershelvam, Veda and Lanctot, Marc and others},
  journal={nature},
  volume={529},
  number={7587},
  pages={484},
  year={2016},
  publisher={Nature Publishing Group}
}

@inproceedings{krause2018dynamic,
  title={Dynamic Evaluation of Neural Sequence Models},
  author={Krause, Ben and Kahembwe, Emmanuel and Murray, Iain and Renals, Steve},
  booktitle={International Conference on Machine Learning},
  pages={2771--2780},
  year={2018}
}

@article{gers2001lstm,
  title={LSTM recurrent networks learn simple context-free and context-sensitive languages},
  author={Gers, Felix A and Schmidhuber, E},
  journal={IEEE Transactions on Neural Networks},
  volume={12},
  number={6},
  pages={1333--1340},
  year={2001},
  publisher={IEEE}
}

@inproceedings{lakshminarayanan2017simple,
  title={Simple and scalable predictive uncertainty estimation using deep ensembles},
  author={Lakshminarayanan, Balaji and Pritzel, Alexander and Blundell, Charles},
  booktitle={Advances in Neural Information Processing Systems},
  pages={6402--6413},
  year={2017}
}

@book{mackay2003information,
  title={Information theory, inference and learning algorithms},
  author={MacKay, David JC and Mac Kay, David JC},
  year={2003},
  publisher={Cambridge university press}
}

@techreport{widrow1960adaptive,
  title={Adaptive switching circuits},
  author={Widrow, Bernard and Hoff, Marcian E},
  year={1960},
  institution={Stanford Univ Ca Stanford Electronics Labs}
}

@article{hochreiter1997long,
  title={Long short-term memory},
  author={Hochreiter, Sepp and Schmidhuber, J{\"u}rgen},
  journal={Neural computation},
  volume={9},
  number={8},
  pages={1735--1780},
  year={1997},
  publisher={MIT Press}
}

@inproceedings{gal2016dropout,
  title={Dropout as a Bayesian approximation: Representing model uncertainty in deep learning},
  author={Gal, Yarin and Ghahramani, Zoubin},
  booktitle={International Conference on Machine Learning},
  pages={1050--1059},
  year={2016}
}

@article{cheng2018polynomial,
  author    = {Xi Cheng and
               Bohdan Khomtchouk and
               Norman Matloff and
               Pete Mohanty},
  title     = {Polynomial Regression As an Alternative to Neural Nets},
  journal   = {CoRR},
  volume    = {abs/1806.06850},
  year      = {2018},
  url       = {http://arxiv.org/abs/1806.06850},
  archivePrefix = {arXiv},
  eprint    = {1806.06850},
  bibsource = {dblp computer science bibliography, https://dblp.org}
}

@article{polyak1964some,
  title={Some methods of speeding up the convergence of iteration methods},
  author={Polyak, Boris T},
  journal={USSR Computational Mathematics and Mathematical Physics},
  volume={4},
  number={5},
  pages={1--17},
  year={1964},
  publisher={Elsevier}
}

@article{grice1975logic,
  title={Logic and conversation},
  author={Grice, H Paul and Cole, Peter and Morgan, Jerry L and others},
  journal={1975},
  pages={41--58},
  year={1975}
}

@article{derose1988grammatical,
  title={Grammatical category disambiguation by statistical optimization},
  author={DeRose, Steven J},
  journal={Computational linguistics},
  volume={14},
  number={1},
  pages={31--39},
  year={1988},
  publisher={MIT Press}
}

@book{ginsburg1966mathematical,
  title={The Mathematical Theory of Context Free Languages.[Mit Fig.]},
  author={Ginsburg, Seymour},
  year={1966},
  publisher={McGraw-Hill Book Company}
}

@inproceedings{church1989stochastic,
  title={A stochastic parts program and noun phrase parser for unrestricted text},
  author={Church, Kenneth Ward},
  booktitle={International Conference on Acoustics, Speech, and Signal Processing,},
  pages={695--698},
  year={1989},
  organization={IEEE}
}

@book{abney2007semisupervised,
  title={Semisupervised learning for computational linguistics},
  author={Abney, Steven},
  year={2007},
  publisher={Chapman and Hall/CRC}
}

@book{sapir1921introduction,
  title={An introduction to the study of speech},
  author={Sapir, Edward},
  year={1921},
  publisher={Citeseer}
}

@article{chomsky1956three,
  title={Three models for the description of language},
  author={Chomsky, Noam},
  journal={IRE Transactions on information theory},
  volume={2},
  number={3},
  pages={113--124},
  year={1956},
  publisher={IEEE}
}

@article{evans2009myth,
  title={The myth of language universals: Language diversity and its importance for cognitive science},
  author={Evans, Nicholas and Levinson, Stephen C},
  journal={Behavioral and brain sciences},
  volume={32},
  number={5},
  pages={429--448},
  year={2009},
  publisher={Cambridge University Press}
}

@article{hauser2002faculty,
  title={The faculty of language: what is it, who has it, and how did it evolve?},
  author={Hauser, Marc D and Chomsky, Noam and Fitch, W Tecumseh},
  journal={science},
  volume={298},
  number={5598},
  pages={1569--1579},
  year={2002},
  publisher={American Association for the Advancement of Science}
}

@article{ruder2016overview,
  author    = {Sebastian Ruder},
  title     = {An overview of gradient descent optimization algorithms},
  journal   = {CoRR},
  volume    = {abs/1609.04747},
  year      = {2016},
  url       = {http://arxiv.org/abs/1609.04747},
  archivePrefix = {arXiv},
  eprint    = {1609.04747},
  bibsource = {dblp computer science bibliography, https://dblp.org}
}

@article{bengio2015deep,
  title={Deep learning},
  author={Bengio, Yoshua and Goodfellow, Ian J and Courville, Aaron},
  journal={Nature},
  volume={521},
  number={7553},
  pages={436--444},
  year={2015},
  publisher={Citeseer}
}

@article{everett20042005,
  title={L. 2005. Cultural constraints on grammar and cognition in Pirah{\~a}},
  author={Everett, Daniel and Berlin, B and Goncalves, MA and Kay, P and Levinson, SC and Pawley, A},
  journal={Current Anthropology},
  volume={46},
  number={4},
  year={2004}
}

@inproceedings{liu2018lstms,
  author    = {Nelson F. Liu and
               Omer Levy and
               Roy Schwartz and
               Chenhao Tan and
               Noah A. Smith},
  title     = {LSTMs Exploit Linguistic Attributes of Data},
  booktitle = {Proceedings of The Third Workshop on Representation Learning for NLP,
               Rep4NLP@ACL 2018, Melbourne, Australia, July 20, 2018},
  pages     = {180--186},
  year      = {2018},
  crossref  = {DBLP:conf/rep4nlp/2018},
  url       = {https://aclanthology.info/papers/W18-3024/w18-3024},
  bibsource = {dblp computer science bibliography, https://dblp.org}
}

@inproceedings{jumelet2018language,
  author    = {Jaap Jumelet and
               Dieuwke Hupkes},
  title     = {Do Language Models Understand Anything? On the Ability of LSTMs to
               Understand Negative Polarity Items},
  booktitle = {Proceedings of the Workshop: Analyzing and Interpreting Neural Networks
               for NLP, BlackboxNLP@EMNLP 2018, Brussels, Belgium, November 1, 2018},
  pages     = {222--231},
  year      = {2018},
  crossref  = {DBLP:conf/emnlp/2018blackbox},
  url       = {https://aclanthology.info/papers/W18-5424/w18-5424},
  bibsource = {dblp computer science bibliography, https://dblp.org}
}

@article{hupkes2018visualisation,
  title={Visualisation and'diagnostic classifiers' reveal how recurrent and recursive neural networks process hierarchical structure},
  author={Hupkes, Dieuwke and Veldhoen, Sara and Zuidema, Willem},
  journal={Journal of Artificial Intelligence Research},
  volume={61},
  pages={907--926},
  year={2018}
}

@article{marcus1993building,
  title={Building a large annotated corpus of English: The Penn Treebank},
  author={Marcus, Mitchell and Santorini, Beatrice and Marcinkiewicz, Mary Ann},
  year={1993}
}

@article{radford2019language,
  title={Language models are unsupervised multitask learners},
  author={Radford, Alec and Wu, Jeffrey and Child, Rewon and Luan, David and Amodei, Dario and Sutskever, Ilya},
  journal={OpenAI Blog},
  volume={1},
  pages={8},
  year={2019}
}

@article{yang2019xlnet,
  author    = {Zhilin Yang and
               Zihang Dai and
               Yiming Yang and
               Jaime G. Carbonell and
               Ruslan Salakhutdinov and
               Quoc V. Le},
  title     = {XLNet: Generalized Autoregressive Pretraining for Language Understanding},
  journal   = {CoRR},
  volume    = {abs/1906.08237},
  year      = {2019},
  url       = {http://arxiv.org/abs/1906.08237},
  archivePrefix = {arXiv},
  eprint    = {1906.08237},
  bibsource = {dblp computer science bibliography, https://dblp.org}
}

@article{kirkpatrick2017overcoming,
  title={Overcoming catastrophic forgetting in neural networks},
  author={Kirkpatrick, James and Pascanu, Razvan and Rabinowitz, Neil and Veness, Joel and Desjardins, Guillaume and Rusu, Andrei A and Milan, Kieran and Quan, John and Ramalho, Tiago and Grabska-Barwinska, Agnieszka and others},
  journal={Proceedings of the national academy of sciences},
  volume={114},
  number={13},
  pages={3521--3526},
  year={2017},
  publisher={National Acad Sciences}
}

@article{milne1982predicting,
  title={Predicting garden path sentences},
  author={Milne, Robert William},
  journal={Cognitive science},
  volume={6},
  number={4},
  pages={349--373},
  year={1982},
  publisher={Elsevier}
}

@article{koenig2017robot,
  title={Robot life-long task learning from human demonstrations: a Bayesian approach},
  author={Koenig, Nathan and Matari{\'c}, Maja J},
  journal={Autonomous Robots},
  volume={41},
  number={5},
  pages={1173--1188},
  year={2017},
  publisher={Springer}
}

@inproceedings{achille2018life,
  title={Life-long disentangled representation learning with cross-domain latent homologies},
  author={Achille, Alessandro and Eccles, Tom and Matthey, Loic and Burgess, Chris and Watters, Nicholas and Lerchner, Alexander and Higgins, Irina},
  booktitle={Advances in Neural Information Processing Systems},
  pages={9873--9883},
  year={2018}
}

@article{kaiser2017learning,
  author    = {Lukasz Kaiser and
               Ofir Nachum and
               Aurko Roy and
               Samy Bengio},
  title     = {Learning to Remember Rare Events},
  journal   = {CoRR},
  volume    = {abs/1703.03129},
  year      = {2017},
  url       = {http://arxiv.org/abs/1703.03129},
  archivePrefix = {arXiv},
  eprint    = {1703.03129},
  bibsource = {dblp computer science bibliography, https://dblp.org}
}

@inproceedings{sukhbaatar2015end,
  title={End-to-end memory networks},
  author={Sukhbaatar, Sainbayar and Weston, Jason and Fergus, Rob and others},
  booktitle={Advances in neural information processing systems},
  pages={2440--2448},
  year={2015}
}

@article{brundage2018malicious,
  author    = {Miles Brundage and
               Shahar Avin and
               Jack Clark and
               Helen Toner and
               Peter Eckersley and
               Ben Garfinkel and
               Allan Dafoe and
               Paul Scharre and
               Thomas Zeitzoff and
               Bobby Filar and
               Hyrum S. Anderson and
               Heather Roff and
               Gregory C. Allen and
               Jacob Steinhardt and
               Carrick Flynn and
               Se{\'{a}}n {\'{O}} h{\'{E}}igeartaigh and
               Simon Beard and
               Haydn Belfield and
               Sebastian Farquhar and
               Clare Lyle and
               Rebecca Crootof and
               Owain Evans and
               Michael Page and
               Joanna Bryson and
               Roman Yampolskiy and
               Dario Amodei},
  title     = {The Malicious Use of Artificial Intelligence: Forecasting, Prevention,
               and Mitigation},
  journal   = {CoRR},
  volume    = {abs/1802.07228},
  year      = {2018},
  url       = {http://arxiv.org/abs/1802.07228},
  archivePrefix = {arXiv},
  eprint    = {1802.07228},
  bibsource = {dblp computer science bibliography, https://dblp.org}
}

@article{hornik1989multilayer,
  title={Multilayer feedforward networks are universal approximators},
  author={Hornik, Kurt and Stinchcombe, Maxwell and White, Halbert},
  journal={Neural networks},
  volume={2},
  number={5},
  pages={359--366},
  year={1989},
  publisher={Elsevier}
}

@inproceedings{dalvi2019one,
  author    = {Fahim Dalvi and
               Nadir Durrani and
               Hassan Sajjad and
               Stephan Vogel},
  title     = {Incremental Decoding and Training Methods for Simultaneous Translation
               in Neural Machine Translation},
  booktitle = {Proceedings of the 2018 Conference of the North American Chapter of
               the Association for Computational Linguistics: Human Language Technologies,
               NAACL-HLT, New Orleans, Louisiana, USA, June 1-6, 2018, Volume 2 (Short
               Papers)},
  pages     = {493--499},
  year      = {2018},
  crossref  = {DBLP:conf/naacl/2018-2},
  url       = {https://aclanthology.info/papers/N18-2079/n18-2079},
  bibsource = {dblp computer science bibliography, https://dblp.org}
}

@inproceedings{conneau2018you,
  author    = {Alexis Conneau and
               Germ{\'{a}}n Kruszewski and
               Guillaume Lample and
               Lo{\"{\i}}c Barrault and
               Marco Baroni},
  title     = {What you can cram into a single {\textbackslash}{\textdollar}{\&}!{\#}*
               vector: Probing sentence embeddings for linguistic properties},
  booktitle = {Proceedings of the 56th Annual Meeting of the Association for Computational
               Linguistics, {ACL} 2018, Melbourne, Australia, July 15-20, 2018, Volume
               1: Long Papers},
  pages     = {2126--2136},
  year      = {2018},
  crossref  = {DBLP:conf/acl/2018-1},
  url       = {https://aclanthology.info/papers/P18-1198/p18-1198},
  bibsource = {dblp computer science bibliography, https://dblp.org}
}

@article{christiansen2016now,
  title={The Now-or-Never bottleneck: A fundamental constraint on language},
  author={Christiansen, Morten H and Chater, Nick},
  journal={Behavioral and Brain Sciences},
  volume={39},
  year={2016},
  publisher={Cambridge University Press}
}

@article{tessler2019incremental,
  title={Incremental understanding of conjunctive generic sentences},
  author={Tessler, Michael Henry and Gu, Karen and Levy, Roger Philip},
  year={2019},
  booktitle={In Proceedings of the 41st Annual Meeting of the Cognitive Science Society}
}

@inproceedings{vaswani2017attention,
  title={Attention is all you need},
  author={Vaswani, Ashish and Shazeer, Noam and Parmar, Niki and Uszkoreit, Jakob and Jones, Llion and Gomez, Aidan N and Kaiser, {\L}ukasz and Polosukhin, Illia},
  booktitle={Advances in neural information processing systems},
  pages={5998--6008},
  year={2017}
}

@inproceedings{devlin2019bert,
  title={BERT: Pre-training of Deep Bidirectional Transformers for Language Understanding},
  author={Devlin, Jacob and Chang, Ming-Wei and Lee, Kenton and Toutanova, Kristina},
  booktitle={Proceedings of the 2019 Conference of the North American Chapter of the Association for Computational Linguistics: Human Language Technologies, Volume 1 (Long and Short Papers)},
  pages={4171--4186},
  year={2019}
}

@article{radford2018improving,
  title={Improving language understanding by generative pre-training},
  author={Radford, Alec and Narasimhan, Karthik and Salimans, Tim and Sutskever, Ilya},
  journal={URL https://s3-us-west-2. amazonaws. com/openai-assets/research-covers/languageunsupervised/language understanding paper. pdf},
  year={2018}
}

@inproceedings{gong2018frage,
  title={FRAGE: frequency-agnostic word representation},
  author={Gong, Chengyue and He, Di and Tan, Xu and Qin, Tao and Wang, Liwei and Liu, Tie-Yan},
  booktitle={Advances in Neural Information Processing Systems},
  pages={1334--1345},
  year={2018}
}

@article{battaglia2018relational,
  author    = {Peter W. Battaglia and
               Jessica B. Hamrick and
               Victor Bapst and
               Alvaro Sanchez{-}Gonzalez and
               Vin{\'{\i}}cius Flores Zambaldi and
               Mateusz Malinowski and
               Andrea Tacchetti and
               David Raposo and
               Adam Santoro and
               Ryan Faulkner and
               {\c{C}}aglar G{\"{u}}l{\c{c}}ehre and
               H. Francis Song and
               Andrew J. Ballard and
               Justin Gilmer and
               George E. Dahl and
               Ashish Vaswani and
               Kelsey R. Allen and
               Charles Nash and
               Victoria Langston and
               Chris Dyer and
               Nicolas Heess and
               Daan Wierstra and
               Pushmeet Kohli and
               Matthew Botvinick and
               Oriol Vinyals and
               Yujia Li and
               Razvan Pascanu},
  title     = {Relational inductive biases, deep learning, and graph networks},
  journal   = {CoRR},
  volume    = {abs/1806.01261},
  year      = {2018},
  url       = {http://arxiv.org/abs/1806.01261},
  archivePrefix = {arXiv},
  eprint    = {1806.01261},
  bibsource = {dblp computer science bibliography, https://dblp.org}
}

@inproceedings{howard2018universal,
  author    = {Jeremy Howard and
               Sebastian Ruder},
  title     = {Universal Language Model Fine-tuning for Text Classification},
  booktitle = {Proceedings of the 56th Annual Meeting of the Association for Computational
               Linguistics, {ACL} 2018, Melbourne, Australia, July 15-20, 2018, Volume
               1: Long Papers},
  pages     = {328--339},
  year      = {2018},
  crossref  = {DBLP:conf/acl/2018-1},
  url       = {https://www.aclweb.org/anthology/P18-1031/},
  doi       = {10.18653/v1/P18-1031},
  bibsource = {dblp computer science bibliography, https://dblp.org}
}

@misc{hochreiter2001gradient,
  title={Gradient flow in recurrent nets: the difficulty of learning long-term dependencies},
  author={Hochreiter, Sepp and Bengio, Yoshua and Frasconi, Paolo and Schmidhuber, J{\"u}rgen and others},
  year={2001},
  publisher={A field guide to dynamical recurrent neural networks. IEEE Press}
}

@inproceedings{virmaux2018lipschitz,
  title={Lipschitz regularity of deep neural networks: analysis and efficient estimation},
  author={Virmaux, Aladin and Scaman, Kevin},
  booktitle={Advances in Neural Information Processing Systems},
  pages={3835--3844},
  year={2018}
}

@phdthesis{gal2016uncertainty,
  title={Uncertainty in deep learning},
  author={Gal, Yarin},
  year={2016},
  school={PhD thesis, University of Cambridge}
}

@inproceedings{li2018visualizing,
  title={Visualizing the loss landscape of neural nets},
  author={Li, Hao and Xu, Zheng and Taylor, Gavin and Studer, Christoph and Goldstein, Tom},
  booktitle={Advances in Neural Information Processing Systems},
  pages={6389--6399},
  year={2018}
}

@book{bishop2006pattern,
  title={Pattern recognition and machine learning},
  author={Bishop, Christopher M},
  year={2006},
  publisher={springer}
}

@inproceedings{dauphin2014identifying,
  title={Identifying and attacking the saddle point problem in high-dimensional non-convex optimization},
  author={Dauphin, Yann N and Pascanu, Razvan and Gulcehre, Caglar and Cho, Kyunghyun and Ganguli, Surya and Bengio, Yoshua},
  booktitle={Advances in neural information processing systems},
  pages={2933--2941},
  year={2014}
}

@inproceedings{futrell2019neural,
  title={Neural language models as psycholinguistic subjects: Representations of syntactic state},
  author={Futrell, Richard and Wilcox, Ethan and Morita, Takashi and Qian, Peng and Ballesteros, Miguel and Levy, Roger},
  booktitle={Proceedings of the 2019 Conference of the North American Chapter of the Association for Computational Linguistics: Human Language Technologies, Volume 1 (Long and Short Papers)},
  pages={32--42},
  year={2019}
}

@book{nesterov2018lectures,
  title={Lectures on Convex Optimization: Second Edition},
  author={Nesterov, Yurii},
  volume={137},
  year={2018},
  publisher={Springer Science \& Business Media}
}
}

\appendix
\chapter{Lemmata, Theorems \& Derivations}\label{appendix:derivations}

This part of the appendix contains many supplementary derivations and proofs that are not essential
to the contents of earlier chapters but worth including or simply too long. Section \ref{appendix:elbo}
derives the evidence lower bound for variational inference and gives some intuition about its implications.
Section \ref{appendix:pred-entropy} shows how a measure for model confidence in classification, predictive entropy,
can be approximated using the methods described in chapters \ref{sec:confidence-dropout} and \ref{sec:confidence-ensembles}. Section \ref{appendix:lipschitz-gradients} gives a short introduction to the concept of Lipschitz continuity and derives an important lemma.
Sections
\ref{appendix-surprisal-recoding-grad} and \ref{appendix-pred-entropy-recoding-grad} derive the corresponding
recoding gradients for surprisal and predictive entropy.

\section{Derivation of the Evidence lower bound}\label{appendix:elbo}

This section described the derivation of the evidence lower bound (\gls{ELBO}) or variational free energy. In variational inference, we try
to approximate the infeasible posterior $p(\btheta|\mathcal{D})$ by an easier distribution $q_\bphi(\btheta)$ parameterized by parameters $\bphi$.
We can cast this into the form of an optimization problem by trying to find the set of parameters $\bphi^*$ that appromates the true posterior best

\begin{equation}
    \bphi^* = \argmin_\bphi \text{KL}\big[q_\bphi(\btheta)||p(\btheta|\mathcal{D})\big]
\end{equation}

where $\text{KL}[\cdot||\cdot]$ is the Kullback-Leibler divergence defined as

\begin{equation}
    \text{KL}\big[q_\bphi(\btheta)||p(\btheta|\mathcal{D})\big] = \int q_\bphi(\btheta) \log \Big(\frac{q_\bphi(\btheta)}{p(\btheta|\mathcal{D})}\Big) d\btheta
\end{equation}

Because $p(\btheta|\mathcal{D})$ is still not computable, we expand it in the following way

\begin{derivation}
  \text{KL}[q_\bphi(\btheta)||p(\btheta|\mathcal{D})] & = & {\displaystyle \int} q_\bphi(\btheta) \log \Big(\frac{q_\bphi(\btheta)}{p(\btheta|\mathcal{D})}\Big) d\btheta \wnl
  & = & \int q_\bphi(\btheta) \log \Big(\frac{q_\bphi(\btheta)p(\mathcal{D})}{p(\btheta, \mathcal{D})}\Big) d\btheta \wnl
  & = & \int q_\bphi(\btheta) \log \Big(\frac{q_\bphi(\btheta)}{p(\btheta, \mathcal{D})}\Big) d\btheta + \int q_\bphi(\btheta) \log p(\mathcal{D}) d\btheta \wnl
  & = & \int q_\bphi(\btheta) \log q_\bphi(\btheta) d\btheta - \int q_\bphi(\btheta) \log p(\btheta, \mathcal{D}) d\btheta  +  \log p(\mathcal{D}) \wnl
  & = & \mathbb{E}_{q_\bphi(\btheta) }[\log p(\btheta, \mathcal{D})] + \mathbb{H}[q_\bphi(\btheta) ] +  \log p(\mathcal{D}) \wnl
  & \ge & \mathbb{E}_{q_\bphi(\btheta) }[\log p(\btheta, \mathcal{D})] + \mathbb{H}[q_\bphi(\btheta) ] = \text{ELBO}[q_\bphi(\btheta)] \wnl
\end{derivation}

where we omit the log evidence $\log p(\mathcal{D})$ and the quantity therefore becomes the lower bound of $\text{KL}[q_\bphi(\btheta)||p(\btheta|\mathcal{D})]$. It can also be rewritten into the form given in an earlier chapter:

\begin{derivation}
  \text{ELBO}[q_\bphi(\btheta)] & = & \int q_\bphi(\btheta) \log q_\bphi(\btheta) d\btheta - \int q_\bphi(\btheta) \log p(\btheta, \mathcal{D}) d\btheta \wnl
  & = & \int q_\bphi(\btheta) \log \Big( \frac{q_\bphi(\btheta)}{p(\btheta, \mathcal{D}) } \Big) d\btheta \wnl
  & = & \int q_\bphi(\btheta) \log \Big( \frac{q_\bphi(\btheta)p(\btheta)}{p(\btheta, \mathcal{D})p(\btheta)} \Big) d\btheta \wnl
  & = & \int q_\bphi(\btheta) \log \Big( \frac{q_\bphi(\btheta)}{p(\btheta)} \Big) d\btheta - \int q_\bphi(\btheta) \log \Big( \frac{p(\btheta, \mathcal{D})}{p(\btheta)} \Big) d\btheta  \wnl
  & = & \text{KL}\big[q_\bphi(\btheta)||p(\btheta)\big] - \mathbb{E}_{q_\bphi(\btheta)}[\log p(\mathcal{D}|\btheta)] \wnl
\end{derivation}

This reformulation gives us more intution about the \gls{ELBO}: The first term encourages densities that are close to the prior distribution $p(\btheta)$, the second term is the expected likelihood that rises when the latent variables or model parameters $\btheta$ explain the data well.

\section{Approximating Predictive Entropy}\label{appendix:pred-entropy}

In this section I re-state the proof given in \cite{gal2016uncertainty} showing that we can approximate the predictive entropy of a model using MC Dropout as well explicitly transferring the same proof to the work of \cite{pearce2018uncertainty}. The problem lies in the fact that the predictive entropy of a model producing a discrete probability distribution defined as

\begin{equation}\label{eq:pred-entropy}
    \mathbb{H}[y|\bx, \mathcal{D}_{\text{train}}] = - {\displaystyle \sum_c} p(y=c|\bx, \mathcal{D}_{\text{train}})\log p(y=c|\bx, \mathcal{D}_{\text{train}}) \wnl
\end{equation}

contains the term $p(y=c|\bx, \mathcal{D}_{\text{train}})$, which we cannot evaluate. However, both MC Dropout \citep{gal2016dropout} and Bayesian Anchored Ensembling \citep{pearce2018uncertainty} are able to approximate this quantity. In the former case, we can expand it to include the true posterior over the model weights

\begin{equation}\label{eq:entropy-post-mcd}
    p(y=c|\bx, \mathcal{D}_{\text{train}}) = \int p(y=c|\bx, \bomega)p(\bomega|\mathcal{D}_{\text{train}})d\bomega \wnl
\end{equation}

 which can be approximated by using the variational distribution placed on the model's weights, which is obtained by Monte Carlo estimates of $K$ model forward passes using different dropout masks (see chapter \ref{sec:confidence-dropout} for more details):

\begin{derivation}
    & & \lim\limits_{k \to \infty} \frac{1}{K}\sum_{k=1}^K p(y=c|\bx, \bomega^{(k)}) \wnl
    & = & \int p(y=c|\bx, \bomega) q^*(\bomega)d\bomega \wnl
    & \approx &  \int p(y=c|\bx, \bomega)p(\bomega| \mathcal{D}_{\text{train}}) d\bomega \wnl
\end{derivation}

Which we can now insert into eq. \ref{eq:pred-entropy}. In the latter case, we similarly have to consider the true posterior like in eq. \ref{eq:entropy-post-mcd}, but this time over all the model parameters $\btheta$ instead of just all model weights $\bomega$:

\begin{equation}
    p(y=c|\bx, \mathcal{D}_{\text{train}}) = \int p(y=c|\bx, \btheta)p(\btheta|\mathcal{D}_{\text{train}})d\btheta \wnl
\end{equation}

We saw in chapter \ref{sec:confidence-ensembles} that the posterior distribution
can be approximated by the \gls{MAP}-solution generating function $f_{\text{MAP}}(\cdot)$.
Let us the denote the probability of a class predicted by the $k$-th member of the ensemble trained
using the anchor parameters $\btheta_0^{(k)} \sim \mathcal{N}(\bmu_0, \bSigma_{\text{prior}})$ as $p(y = c | \bx, \btheta_0^{(k)})$. Then it can be shown that

\begin{derivation}\label{eq:bae-post-approx}
    & & \lim\limits_{\substack{k \to \infty\\ \uparrow \text{corr.}}} \frac{1}{K}\sum_{k=1}^K p(y=c|\bx, \btheta_0^{(k)})p(\btheta_0^{(k)}) \wnl
    & = & \int p(y=c|\bx, \btheta)p(\btheta|\mathcal{D}_{\text{train}})d\btheta \wnl
\end{derivation}

where $\uparrow \text{corr.}$ stands for increasing correlation between the parameters $\btheta^{(k)}$, which is known to occur in deep neural networks during training (see \cite{cheng2018polynomial} \S 7.1, \cite{srivastava2014dropout}). Therefore, this method is equally compatible to approximate the predictive entropy in eq. \ref{eq:pred-entropy}. The whole proof of insight is too elaborate to be included here, which is why I refer the reader to appendix A of \cite{pearce2018uncertainty} for the full derivation.

\section{Lipschitz Continuity}\label{appendix:lipschitz-gradients}

\begin{definition}[Lipschitz continuity]
  A function $f: \mathbb{R}^n \mapsto \mathbb{R}$ is called Lipschitz continuous
  on $\mathbb{R}^n$ with a non-negative constant $L$ if

  \begin{equation}
    |f(\bx) - f(\by)| \le L||\bx-\by||_{(\infty)}\ \forall \bx, \by \in \mathbb{R}^n
  \end{equation}

  where the $l_\infty$ norm is defined as

  \begin{equation}
    ||\bx||_{(\infty)} = \max_{1 \le i \le n}|\bx_i|
  \end{equation}
\end{definition}

Intuitively, a Lipschitz continuous function is limited in its slope $|f(\bx) - f(\by)|$
by some Lipschitz constant $L$. The set of Lipschitz continuous functions also is
a superset of all continuously differentiable functions. There also exists a class
of functions $f(\cdot)$ with a Lipschitz continuous gradient that follows a similar definition:

\begin{definition}[Lipschitz continuous gradients]
  The first derivative of a function $f: \mathbb{R}^n \mapsto \mathbb{R}$ is called Lipschitz continuous
  on $\mathbb{R}^n$ if

  \begin{equation}
    ||\nabla f(\bx) - \nabla f(\by)|| \le L||\bx-\by||\ \forall \bx, \by \in \mathbb{R}^n
  \end{equation}
\end{definition}

The following lemma is adapted from \cite{nesterov2018lectures} \S 1.2.2:

\begin{lemma}\label{lemma:lipschitz-gradients}
  For every function $f: \mathbb{R}^n \mapsto \mathbb{R}$ with a Lipschitz continuous first derivative it holds that

  \begin{equation}
    |f(\by) - f(\bx) - \langle\nabla f(\bx), \by - \bx\rangle| \le \frac{L}{2}||\by - \bx||^2
  \end{equation}
\end{lemma}

\emph{Proof.} We start off be realizing that the value of a function $f$ at a point $\by$ can be reached by starting at a point $\bx$ and adding all the slopes of $\nabla f$ on the $\by-\bx$ line to $f(\bx)$:

\begin{derivation}
  f(\by) & = & f(\bx) + \int_0^1 \langle \nabla f(\bx + \tau(\by-\bx)), \by - \bx\rangle d\tau \wnl
   & = & f(\bx) + \int_0^1 \langle \nabla f(\bx + \tau(\by-\bx)) - \nabla f(\bx) + \nabla f(\bx), \by - \bx\rangle d\tau \wnl
   & = & f(\bx) + \langle\nabla f(\bx), \by - \bx \rangle + \int_0^1 \langle \nabla f(\bx + \tau(\by-\bx)) - \nabla f(\bx), \by - \bx\rangle d\tau \\
\end{derivation}
\begin{derivation}
  |f(\by) - f(\bx) - \langle\nabla f(\bx), \by - \bx \rangle | & = & |\int_0^1 \langle \nabla f(\bx + \tau(\by-\bx)) - \nabla f(\bx), \by - \bx\rangle d\tau| \wnl
\end{derivation}
Now, due to the fact that $-|f(\bx)| \le f(\bx) \le |f(\bx)| \rightarrow -\int|f(\bx)|d\bx \le \int f(\bx)d\bx \le \int|f(\bx)|d\bx \rightarrow |\int f(\bx)d\bx| \le \int|f(\bx)|d\bx$ (first line) and using the Cauchy-Schwartz inequality (second line) alongside the definition of Lipschitz continuous gradients above (third line) we can write

\begin{derivation}
  |f(\by) - f(\bx) - \langle\nabla f(\bx), \by - \bx \rangle | & \le & \int_0^1 |\langle \nabla f(\bx + \tau(\by-\bx)) - \nabla f(\bx), \by - \bx\rangle| d\tau \wnl
  & \le & \int_0^1 ||\nabla f(\bx + \tau(\by-\bx)) - \nabla f(\bx)||\cdot||\by - \bx|| d\tau \wnl
  & \le & \int_0^1 L||\bx + \tau(\by-\bx) - \bx||\cdot||\by - \bx|| d\tau \wnl
  & = & \int_0^1 \tau L|| \by-\bx ||^2 \wnl
  & = & \Big[\frac{1}{2}\tau^2L|| \by-\bx ||^2\Big]_0^1 \wnl
  & = & \frac{L}{2}|| \by-\bx ||^2 \wnl
\end{derivation}
\proofend

\section{Theoretical guarantees of Recoding}\label{appendix:theoretical-guarantees}

In the following I will show that recoding indeed reduces the error signal $\delta_t$ produced by some hidden activations $\bh_t$, which parallels a proof described in \cite{nesterov2018lectures}.

\begin{theorem}[Encoded error reduction]\label{theorem:error-reduction}
  For some error signal $\delta_t$ based on some hidden activations $\bh_t$ and some error signal $\delta_t^\prime$ based on the recoded hidden activations $\bh_t^\prime$ obtained using the same error function $e(\cdot)$, it holds that

   \[\delta_t^\prime \le \delta_t \]
\end{theorem}

\emph{Proof.} Let us denote the function producing a time-dependent error signal $\delta_t$ based on hidden activations $\bh_t$ as $e(\cdot)$. We can then apply the property for functions with Lipschitz continous gradients described in appendix \ref{appendix:lipschitz-gradients}, lemma \ref{lemma:lipschitz-gradients}:

\begin{equation}
    |e(\bh_t^\prime) - e(\bh_t) - \langle\nabla e(\bh_t), \bh_t^\prime -\bh_t\rangle| \le \frac{L}{2}|| \bh_t^\prime -\bh_t||^2
\end{equation}

Given the recoding update rule described in equation \ref{eq:recoding-update}, we can rewrite this as

\begin{derivation}
  |e(\bh_t^\prime) - e(\bh_t) - \langle \nabla e(\bh_t), \bh_t^\prime -\bh_t \rangle| & \le & \frac{L}{2}|| \bh_t^\prime -\bh_t||^2 \wnl
  |e(\bh_t^\prime) - e(\bh_t) - \langle \nabla e(\bh_t), -\alpha\nabla e(\bh_t)\rangle| & \le & \frac{L}{2}||-\alpha\nabla e(\bh_t)||^2 \wnl
  |e(\bh_t^\prime) - e(\bh_t) + \alpha||\nabla e(\bh_t)||^2| & \le & \frac{\alpha^2L}{2}||\nabla e(\bh_t)||^2 \wnl
  e(\bh_t^\prime) & \le & e(\bh_t) - \alpha||\nabla e(\bh_t)||^2 + \frac{\alpha^2L}{2}||\nabla e(\bh_t)||^2 \wnl
   & = & e(\bh_t) -\alpha(1 - \frac{\alpha L}{2})||\nabla e(\bh_t)||^2 \wnl
\end{derivation}

In order to find the optimal step size $\alpha$, we can identify the extremum of $-\alpha(1 - \frac{\alpha L}{2})$ as $\alpha^* = \frac{1}{L}$ by setting it to $0$ and solving for $\alpha$. Resubstituting into the result above yields

\begin{derivation}
  e(\bh_t^\prime) & \le & e(\bh_t) -\alpha(1 - \frac{\alpha L}{2})||\nabla e(\bh_t)||^2 \wnl
  & = & e(\bh_t) -\frac{1}{L}(1 - \frac{L}{2L})||\nabla e(\bh_t)||^2 \wnl
  e(\bh_t) - e(\bh_t^\prime) & \ge & \frac{1}{2L}||\nabla e(\bh_t)||^2 \wnl
  \delta_t - \delta_t^\prime & \ge & \frac{1}{2L}||\nabla_{\bh_t} \delta_t||^2 \wnl
\end{derivation}

This implies that the improvement in terms of the error signal after recoding is at least $\frac{1}{2L}||\nabla_{\bh_t} \delta_t||^2$. Assuming the Lipschitz constant is $L > 0$, this implies that
a theoretical new error signal $\delta_t^\prime$ based on the recoded hidden activations $\bh_t^\prime$ has at least the same value as the original error signal $\delta_t$ or lower. \proofend

In practice, this also entails that obtaining $L$ can help us determine the optimal step size at every time step so that we obtain a guaranteed improvement. This means that setting a suboptimal step size still can result in a higher error signal encoded in the hidden activations, contrary to the implication of this theorem.

\begin{theorem}[Encoded error reduction through time]\label{theorem:error-reduction-tt}
  Given a sequence of time steps from $t$ to $t+k$ where intermediate hidden activations $\bh_{t+1}^\prime, \ldots, \bh_{t+k}^\prime$ are being recoded, we denote $\delta_{t+k}$ the error signal at time step $t+k$ where only the activations $\bh_{t}$ were not recoded and $\delta_{t+k}^*$ the error signal resulting from all the activations from $t$ to $t+k$, including $\bh_{t}$, being recoded. It then holds that

   \[\delta_{t+k}^* \le \delta_{t+k} \]

\end{theorem}

\emph{Proof.} Here I show how recoding activations at a time step $t$ even influences the error signal produced at a later time step $\delta_{t+k}$. To show this, we evaluate the difference to the case where recoding is not performed at time step $t$. Let us denote $g_\btheta^{(t)}(\cdot)$ the recurrent function\footnote{This is the same recurrent function, the superscript is simply used to keep track of the number of times the function was applied.} from equation \ref{eq:recurrency} and $r^{(t)}(\cdot)$ the recoding function used at time step $t$. We furthermore denote the hidden activations produced by the recurrent function based on recoded hidden activations as $\bh_{t+1}^\prime = g_\btheta^{(t+1)}(\bx_{t+1}, \bh_t^\prime) = g_\btheta^{(t+1)}(\bx_{t+1}, r^{(t)}(\bh_t))$.
We can thus expand the error signal terms as follows

\begin{derivation}
  \delta_{t+k}^* & = & e(\bh_{t+k}^\prime) = e(g_\btheta^{(t+k)}(\bx_{t+k}, \bh_{t+k-1}^\prime)) \wnl
  & = & e(g_\btheta^{(t+k)}(\bx_{t+k},\ \ldots\ r^{(t+1)}(g_\btheta^{(t+1)}(\bx_{t+1}, \varcol{\bh_{t}^\prime})))) \wnl
  \delta_{t+k} & = & e(g_\btheta^{(t+k)}(\bx_{t+k},\ \ldots\ r^{(t+1)}(g_\btheta^{(t+1)}(\bx_{t+1}, \varcol{\bh_{t}})))) \wnl
\end{derivation}

where the \varcol{only difference} is that for $\delta_{t+k}^*$, the recoding of $\bh_t$ produced $\bh_t^\prime$, while for $\delta_{t+k}$, these activations where left untouched. We now define a composite function $e_{t}^{t+k}(\cdot)$ as

\begin{derivation}
  e_{t}^{t+k} & = & e \circ g_\btheta^{(t+k)} \circ r^{(t+k-1)} \circ g_\btheta^{(t+k-1)}   \circ \ldots \circ r^{(t+1)} \circ g_\btheta^{(t+1)} \wnl
\end{derivation}

where we omit the intermediate inputs $\bx_{t+1}, \ldots, \bx_{t+k}$ in favor of a more compact notation for $e_{t}^{t+k}(\cdot)$ as they are not relevant for the recoding gradient\footnote{This is an admitted abuse of notation, as composition for multivariate functions would be denoted here as $g_\btheta^{(t+1)}|_{\bh_t = r(\bh_t)}(\bx_{t+1}, r(\bh_t))$ The notation above was therefore used to increase readability. The proof still holds as we are only interested in the gradient of the composite function w.r.t some $\bh_t$ and not $\bx_t$ and the derivative $\pder{\bh_{t+1}}{\bh_t^\prime}$ does not involve $\bx_t$.}.
Using this definition, we can follow the same proof as in the last theorem: As it applies to any black-box function with Lipschitz continuous gradients and because the composition of functions with this property also possesses Lipschitz continuous gradients, we can show that

\begin{derivation}
  e_{t}^{t+k}(\bh_t) - e_{t}^{t+k}(\bh_t^\prime) & \ge & \frac{1}{2L}||\nabla e_{t}^{t+k}(\bh_t)||^2 \wnl
  \delta_{t+k} - \delta_{t+k}^* & \ge & \frac{1}{2L}||\nabla_{\bh_t} \delta_{t+k}||^2 \wnl
\end{derivation}

and therefore analagously $\delta_{t+k}^* \le \delta_{t+k}$.\proofend

This shows that the recoding even improves the error signal of later time steps in a recurrent context given a sensible choice of the step size $\alpha$. However, it should be noted that this improvement might be very small, as $\nabla_{\bh_t} \delta_{t+k}$ requires one to evaluate the chain of partial derivatives

\begin{derivation}
    \nabla_{\bh_t} \delta_{t+k} & = & \pder{\delta_{t+k}}{\bh_{t+k}}\pder{\bh_{t+k}}{\bh_{t+k-1}^\prime}\pder{\bh_{t+k-1}^\prime}{\bh_{t+k-1}}\ldots\pder{\bh_{t+1}}{\bh_t^\prime}\pder{\bh_t^\prime}{\bh_t} \wnl
\end{derivation}

which is reminiscent of the vanishing gradient problem.

\section{Derivation of the Surprisal recoding gradient}\label{appendix-surprisal-recoding-grad}

This section describes the full derivation of the recoding gradient when using surprisal as the error signal $\bdelta_t$. Let us start by rewriting the definition of surprisal given in eq. \ref{eq:surprisal-error-signal}  into a form that is easier to take a derivative of:

\begin{derivation}
    \text{Z}(y_t|\bh_t, \btheta) & = & \sum_{c}\indicator{y_t = c}\bo_{tc}^{-\bo_{tc}} - 1 \wnl
    & = & \sum_{c}\indicator{y_t = c}\exp\Big(-\bo_{tc}\log(\bo_{tc})\Big) - 1 \wnl
    & = & \exp\Big(-\mathbf{A}\bo_{t}^T\log(\bo_{t})\Big) - 1 \wnl
\end{derivation}

where $\mathbf{A}$ denotes a square $C \times C$ matrixe where $C$ is the number of classes\footnote{In the language modelling case, this corresponds to the cardinality of the vocabularity $|\mathcal{V}|$.}, which only contains zeros except for a single entry on the diagonal corresponding to the label $y_t$:

\begin{derivation}
  \mathbf{A} & = &  \begin{bmatrix}
                    \indicator{A_{11} = y_t}  & 0                         & \cdots  & 0 \\
                    0                         & \indicator{A_{22} = y_t}  & \cdots  & 0  \\
                    \vdots                    & \vdots                    & \ddots  & \vdots   \\
                    0                         & 0                         & \cdots  & \indicator{A_{CC} = y_t} \\
                    \end{bmatrix}
\end{derivation}

We can then write the recoding gradient in terms of its partial derivatives:

\begin{derivation}
  \nabla_{\bh_t}\text{Z}(y_t|\bh_t, \btheta) & = & \pder{\text{Z}(y_t|\bh_t, \btheta)}{\bo_t}\pder{\bo_t}{\botilde_t}\pder{\botilde_t}{\bh_t} \wnl
\end{derivation}

This way, we can evaluate every partial derivative individually:

\begin{derivation}
    \pder{\text{Z}(y_t|\bh_t, \btheta)}{\bo_t} & = & \pder{}{\bo_t}\exp\Big(-\mathbf{A}\bo_{t}^T\log(\bo_{t})\Big) - 1 \wnl
    & = & -\mathbf{A}\pder{}{\bo_t}\Big(\bo_{t}^T\log(\bo_{t})\Big)\exp\Big(-\mathbf{A}\bo_{t}^T\log(\bo_{t})\Big)\wnl
    & = & -\mathbf{A}\Big(\log(\bo_t) + \bo_{t}^T\Big(\frac{1}{\bo_t}\Big)\Big)\exp\Big(-\mathbf{A}\bo_{t}^T\log(\bo_{t})\Big)\wnl
    & = & -\mathbf{A}\Big(\log(\bo_t) + \mathbf{1}\Big)\exp\Big(-\mathbf{A}\bo_{t}^T\log(\bo_{t})\Big)\wnl
\end{derivation}

For the derivative of the softmax layer, we cannot take $\pder{\bo_t}{\botilde_t}$ directly and therefore revert to its derivative on a scalar-level first using the quotient rule:

\begin{derivation}
    \pder{\bo_{ti}}{\botilde_{tj}} & = & \pder{}{\botilde_{tj}}\frac{\exp(\botilde_{ti})}{\sum_{j=1}^J\exp(\botilde_{tj})} \wnl
    & = & \frac{\pder{}{\botilde_{tj}}\Big(\exp(\botilde_{ti})\Big)\Big(\sum_{j=1}^J\exp(\botilde_{tj}) \Big)}{\Big(\sum_{j=1}^J\exp(\botilde_{tj})\Big)^2} - \frac{\exp(\botilde_{ti})\pder{}{\botilde_{tj}}\Big(\sum_{j=1}^J\exp(\botilde_{tj})\Big)}{\Big(\sum_{j=1}^J\exp(\botilde_{tj})\Big)^2} \\[0.6cm]
    & = & \frac{\indicator{i = j}\exp(\botilde_{ti})}{\sum_{j=1}^J\exp(\botilde_{tj})} - \frac{\exp(\botilde_{ti})\exp(\botilde_{tj})}{\Big(\sum_{j=1}^J\exp(\botilde_{tj})\Big)^2} \wnl
    & = & \frac{\indicator{i = j}\exp(\botilde_{ti})}{\sum_{j=1}^J\exp(\botilde_{tj})} - \frac{\exp(\botilde_{ti})}{\sum_{j=1}^J\exp(\botilde_{tj})} \cdot \frac{\exp(\botilde_{tj})}{\sum_{j=1}^J\exp(\botilde_{tj})} \wnl
    & = & \indicator{i = j}\bo_{ti} - \bo_{ti}\bo_{tj} = \bo_{ti}(\indicator{i = j} - \bo_{tj}) \wnl
\end{derivation}

where the last step is performed by using the original definition of the softmax function. Finally, we can convert this result back into vector form by noticing that $\indicator{i = j}\bo_{ti}$ corresponds to a diagonal matrix whose elements are taken from the vector $\bo_{ti}$ and that the value $\bo_{ti}\bo_{tj}$ corresponds to an entry in the matrix produced by multipling $\bo_{t}$ with its transpose:

\begin{derivation}
  \pder{\bo_{t}}{\botilde_{t}} & = & \text{diag}(\bo_{t}) - \bo_{t}\bo_{t}^T \wnl
\end{derivation}

Finally, by deriving the last part

\begin{derivation}
    \pder{\botilde_t}{\bh_t} & = & \pder{}{\bh_t}\Big(\bW_{ho}\bh_t + \bb_{ho}\Big) = \bW_{ho} \wnl
\end{derivation}

we can identify the full recoding gradient as

\begin{derivation}
  \nabla_{\bh_t}\text{Z}(y_t|\bh_t, \btheta) & = & -\mathbf{A}\Big(\log(\bo_t) + \mathbf{1}\Big)\exp\Big(-\mathbf{A}\bo_{t}^T\log(\bo_{t})\Big)\Big(\text{diag}(\bo_{t}) - \bo_{t}\bo_{t}^T\Big)\bW_{ho} \wnl
\end{derivation}

\proofend

\section{Derivations fo the Predictive Entropy recoding gradient}\label{appendix-pred-entropy-recoding-grad}

In the following I provide a full derivation of the recoding gradient when using predictive entropy as the error signal $\bdelta_t$. The predictive entropy is very helpful when judging the certainty of a model in a classification setting as it quantifies the average amount of information in the model's predictive distribution. It's maximum value is reached when the distribution is uniform, the entropy is $0$ when only class has probability one. The predictive entropy is defined as

\begin{derivation}
    \mathbb{H}[y_t|\bx_t, \mathcal{D}_{\text{train}}] & = & - {\displaystyle \sum_c} p(y=c|\bx, \mathcal{D}_{\text{train}})\log p(y=c|\bx, \mathcal{D}_{\text{train}})
\end{derivation}

Evaluating $p(y=c|\bx, \mathcal{D}_{\text{train}})$ is not feasible, we can therefore estimate this quantity using a Monte Carlo estimate combining multiple predictive distributions $\bo_t^{(k)}$ obtained under different circumstances (different sets of dropout masks for variational or dropout recoding or different ensemble members for Anchored Bayesian Ensembling). Section \ref{appendix:pred-entropy} shows in his work how this approximates the real predictive entropy in the limit. We thus obtain

\begin{derivation}
   \bohat_t & = & \frac{1}{K}{\displaystyle\sum_{k=1}^K} \bo_t^{(k)} \wnl
   p(y=c|\bx, \mathcal{D}_{\text{train}}) & \approx & \bohat_{tc} \wnl
   \tilde{\mathbb{H}}[y_t|\bh_t, \mathcal{D}_{\text{train}}] & = & - {\displaystyle\sum_c} \bohat_{tc}\log(\bohat_{tc}) =-\bohat_t^T\log(\bohat_t)
\end{derivation}

where condition on the hidden state $\bh_t$ due to the recurrent setting.
Next, let us write the gradient $\nabla_{\bh_t} \tilde{\mathbb{H}}[y_t|\bh_t, \mathcal{D}_{\text{train}}]$ in terms of its partial derivatives:

\begin{derivation}
  \nabla_{\bh_t} \tilde{\mathbb{H}}[y_t|\bh_t, \mathcal{D}_{\text{train}}] & = & \pder{\tilde{\mathbb{H}}[y_t|\bh_t, \mathcal{D}_{\text{train}}]}{\bohat_t}{\displaystyle \sum_{k=1}^K}\pder{\bohat_t}{\bo_t^{(k)}}\pder{\bo_t^{(k)}}{\botilde_t^{(k)}}\pder{\botilde_t^{(k)}}{\bh_t}
\end{derivation}

where we can now evaluate the parts separately:

\begin{derivation}
    \pder{\tilde{\mathbb{H}}[y_t|\bh_t, \mathcal{D}_{\text{train}}]}{\bohat_t} & = & \pder{}{\bohat_t}\Big(-\bohat_t^T\log(\bohat_t)\Big) \wnl
    & = & - \Big(\pder{}{\bohat_t} \bohat_t\Big)^T \log(\bohat_t) - \bohat_t^T\Big(\pder{}{\bohat_t}\log(\bohat_t)\Big) \wnl
    & = & - \log(\bohat_t) - \bohat_t^T\Big(\frac{1}{\bohat_t}\Big) \wnl
    & = & - \Big(\log(\bohat_t) + \mathbf{1}\Big) \wnl
    \pder{\bohat_t}{\bo_t^{(j)}} & = & \pder{}{\bo_t^{(j)}}\Big(\frac{1}{K}{\displaystyle\sum_{k=1}^K} \bo_t^{(k)}\Big) = \frac{1}{K} \wnl
\end{derivation}

because we evaluated $\pder{\bo_t^{(k)}}{\botilde_t^{(k)}}$ and $\pder{\botilde_t^{(k)}}{\bh_t}$ already in the previous section for a single $\bo_t$, $\botilde_t$, we obtain the final result as

\begin{derivation}
    \nabla_{\bh_t} \tilde{\mathbb{H}}[y_t|\bh_t, \mathcal{D}_{\text{train}}] & = & - \Big(\log(\bohat_t) + \mathbf{1}\Big)\frac{1}{K}{\displaystyle \sum_{k=1}^K}(\text{diag}(\bo_t^{(k)}) - \bo_t^{(k)}(\bo_t^{(k)})^T)\bW_{ho}^{(k)} \wnl
\end{derivation}

\proofend

\chapter{Experimental Details \& Supplementary Results}\label{appendix:additional}

In constrast to the previous appendix, this chapter supplies the reader with additional, practical
information about the models and experiments in this thesis. To this end, section \ref{appendix:pseudocode}
lists pseudocode for the different models. Section \ref{appendix:practical-consid} ponders on practical issues that
have to be considered when implementing the recoding procedure. Section \ref{sec:hyperparameter-search} gives more detail to the hyperparameter search that was employed to train the models in chapter \ref{chapter:experiments} and
section \ref{appendix:additional-plots} gives additional plots from chapters \ref{chapter:experiments} and \ref{chapter:qualitative}.

\section{Pseudocode}\label{appendix:pseudocode}

This section contains the pseudocode for the main models used in this thesis.
Fig. \ref{fig:code-surp-recoding} refers to the surprisal-based recoding approach
described in chapter \ref{subsec:surprisal}. The code that describes the implementation
of MC Dropout recoding in chapter \ref{subsec:mc-dropout} and Bayesian Anchored Ensemble recoding in chapter \ref{subsec:anchored-ensembling}
is given in figs. \ref{fig:code-mcd-recoding} and \ref{fig:code-bae-recoding}, respectively.

\begin{figure}[H]
  \begin{algorithm}[H]
    \DontPrintSemicolon
    \SetKwComment{Comment}{}{}
    \ForAll{$\bx_t \in \mathcal{X}$}{
        $\bh_t = g_\btheta(\bx_t, \bh_{t-1}^\prime)$\;
        $\botilde_t = \bW_{ho}\bh_t + \bb_{ho}$\;
        $\bo_t = \text{softmax}(\botilde_t)$\;
        \BlankLine

        \Comment{/* Recoding}
        Set $\delta_t = \text{Z}(y_t|\bx_t, \btheta)$ based on gold label $y_t$\;
        Calculate $\nabla_{\bh_t}\delta_t$\;
        $\bh_t^\prime = \bh_t - \alpha\nabla_{\bh_t}\delta_t$\;
    }
  \end{algorithm}
  \caption[Pseucode for Recoding with Surprisal]{Pseudocode for the forward pass of a model using surprisal as the recoding error signal $\delta_t$.}\label{fig:code-surp-recoding}
\end{figure}

\begin{figure}[H]
  \begin{algorithm}[H]
  \DontPrintSemicolon
  \SetKwComment{Comment}{}{}

\Comment{/* Forward pass}
  \ForAll{$\bx_t \in \mathcal{X}$}{
      $\bh_t = g_\btheta(\bx_t, \bh_{t-1}^\prime)$;
      \BlankLine
      \Comment{/* Sample k different dropout masks and predict}
      \For{$k \in 1 \ldots K$}{
          $\bZ_{ij}^{(k)} \sim \text{Bernoulli}(p)\ \forall i = 1, \ldots, N\ \ \forall j= 1, \ldots, |V|$\;
          $\bW_{ho}^{(k)} = \bW_{ho}\odot\bZ^{(k)}$\;
          $\botilde_t^{(k)} = \bW_{ho}^{(k)}\bh_t + \bb_{ho}$\;
          $\bo_t^{(k)} = \text{softmax}(\botilde_t^{(k)})$\;
      }

      \BlankLine
      \Comment{/* Recoding}
      Calculate predictive entropy $\delta_t \equiv \tilde{\mathbb{H}}[y|\bh_t, \mathcal{D}_{\text{train}}]$\;
      Calculate $\nabla_{\bh_t}\delta_t$\;
      $\bh_t^\prime = \bh_t - \alpha_t\nabla_{\bh_t}\delta_t$\;
      \BlankLine

      \Comment{/* Recompute output distribution for loss}
      $\botilde_t^\prime = \bW_{ho}\bh_t^\prime + \bb_{ho}$\;
      $\bo_t^\prime = \text{softmax}(\botilde_t^\prime)$\;
  }
  \end{algorithm}
  \caption[Pseudocode for Recoding with MC Dropout]{Pseudocode for the forward pass of a model using its predictive variance estimated with MC Dropout as the recoding error signal $\delta_t$.}\label{fig:code-mcd-recoding}
\end{figure}

\begin{figure}[h]
  \begin{algorithm}[H]
  \DontPrintSemicolon
  \SetKwComment{Comment}{}{}

  \Comment{/* Before training}
  \Comment{/* Initialize weights for every member of the ensemble by}
  \For{$k \in 1 \ldots K$}{
     Sample weights $\bW_{ho}^{(k)} \sim \mathcal{N}(0, \mathbf{\Sigma}_{\text{prior}})$\;
     Sample anchor points $\bW_0^{(k)} \sim \mathcal{N}(0, \mathbf{\Sigma}_{\text{prior}})$\;
  }

  \BlankLine
  \Comment{/* Forward pass}
    \ForAll{$\bx_t \in \mathcal{X}$}{
        $\bh_t = g_\btheta(\bx_t, \bh_{t-1}^\prime)$;
        \BlankLine
        \Comment{/* Use k different ensemble members to make predictions}
        \For{$k \in 1 \ldots K$}{
            $\botilde_t^{(k)} = \bW_{ho}^{(k)}\bh_t + \bb_{ho}$\;
            $\bo_t^{(k)} = \text{softmax}(\botilde_t^{(k)})$\;
        }

        \BlankLine
        \Comment{/* Recoding}
        Calculate predictive entropy $\delta_t \equiv \tilde{\mathbb{H}}[y|\bh_t, \mathcal{D}_{\text{train}}]$\;
        Calculate $\nabla_{\bh_t}\delta_t$\;
        $\bh_t^\prime = \bh_t - \alpha_t\nabla_{\bh_t}\delta_t$\;
        \BlankLine

        \Comment{/* Recompute output distribution for loss}
        $\botilde_t^\prime = \bW_{ho}\bh_t^\prime + \bb_{ho}$\;
        $\bo_t^\prime = \text{softmax}(\botilde_t^\prime)$\;
    }

  \BlankLine
  \Comment{/* Loss}
  \For{$k \in 1 \ldots K$}{
      Calculate anchored weight decay loss: $\mathcal{L}_{\text{anchor}}^{(k)} = \frac{1}{N}||\bGamma^{1/2}\bW_{ho}^{(k)} - \bW_0^{(k)}||$\;
  }
  \Comment{/* Amortize Loss}
  $\mathcal{L}_{\text{total}} = \frac{1}{K}\sum_{k=1}^K \big( \mathcal{L}^{(k)}_\text{CE} + \mathcal{L}_{\text{anchor}}^{(k)}\big)$\;

  \end{algorithm}
  \caption[Pseudocode for Recoding with Bayesian Anchored Ensembles]{Pseudocode for the forward pass of a model using its predictive variance estimated with Bayesian Anchored ensembling as the recoding error signal $\delta_t$.}\label{fig:code-bae-recoding}
\end{figure}

\section{Practical Considerations}\label{appendix:practical-consid}

Lastly I elaborate on some practical issues when implementing recoding. Although we derived the recoding gradients $\nabla_{\bh_t}\delta_t$ by hand in the previous appendix, this is luckily not necessary in modern Deep Learning frameworks like \verb|PyTorch| or \verb|Tensorflow|, as those come equipped with an automatic differentiation engine, which builds a computational graph on the network and is therefore able to efficiently compute partial derivatives. This gives rise to the question whether all the operations involved in the recoding should be added to the graph as well. In preliminary experiments this did not seem to make a significant difference in terms of performance. However when using \verb|PyTorch|, including them in the graph sometimes lead to memory leakages in the automatic differentiation engine, the circumstances of which were rather inconsistent. It is possible to violate common practice and perform the recoding on the hidden activations' \verb|.data| attribute. This way avoids any memory issues, but does not allow gradients to backpropagate far enough when using the learned or predicted step size. In the latter case, it should also be made sure that the hidden states that are used as the predictor's input are detached from the graph to avoid memory errors (gradient will otherwise be backpropagated through them as well). As a final point, I found it useful to use the \verb|torch.autograd.grad| function to compute recoding gradients instead of calling \verb|.backward()| on the error signal tensor, as hidden activations do not have to be explicitly declared as variables that the autograd engine has to keep track off this way.

\section{Hyperparameter Search}\label{sec:hyperparameter-search}

In this section I provide more detail about the hyperparameter search performed on the
Penn Treebank Corpous using the different model variants described in this thesis. As described in chapter \ref{sec:exp-training}, many basic parameters are taken from the work of \cite{gulordava2018colorless}, who performed extensive grid search over $68$ different combinations in the Wikitext-2 data set.\footnote{For more details refer to the supplementary material of \cite{gulordava2018colorless} under \url{https://github.com/facebookresearch/colorlessgreenRNNs/blob/master/supplementary_material.pdf}.} The resulting parameters are listed in table \ref{subfig:all-hyperparams}.\\

\begin{table}
    \begin{subfigure}[b]{0.3\columnwidth}
    \centering
    \setlength{\tabcolsep}{4pt}
    \def\arraystretch{1.5}
    \resizebox{.9\columnwidth}{!}{
    \begin{tabular}{@{}ll@{}}
         \toprule[1.5pt]
         Hyperparameter & Value \\
         \toprule[1.15pt]
         Number of layers & 2 \\
         Embedding size & 650 \\
         Hidden size & 650 \\
         Batch size & 64 \\
         Learning rate & 20 \\
         Gradient clipping & 0.25 \\
         Sequence Length & 35 \\
         \bottomrule[1.15pt]
    \end{tabular}}
    \vspace{0.2cm}
    \caption{Hyperparameters shared by all models.}\label{subfig:all-hyperparams}
    \end{subfigure}\hfill%
    \begin{subfigure}[b]{0.7\columnwidth}
    \centering
    \setlength{\tabcolsep}{4pt}
    \def\arraystretch{1.5}
    \resizebox{.85\columnwidth}{!}{
    \begin{tabular}{@{}lll@{}}
         \toprule[1.5pt]
         Hyperparameter & Recoding Models \\
         \toprule[1.15pt]
         Step size & All \\
         \# Samples & MC Dropout, BAE, Variational \\
         Prior scale & MC Dropout, BAE, Variational \\
         Weight decay & MC Dropout, Variational \\
         MC Dropout & MC Dropout, Variational \\
         Learning rate & Surprisal \\
         \bottomrule[1.15pt]
    \end{tabular}}
    \vspace{0.2cm}
    \caption{Hyperparameters shared by all or some recoding-based models.}\label{subfig:recoding-hyperparams}
    \end{subfigure}
    \caption[Training Hyperparameters]{List of core hyperparameters based in the work of \cite{gulordava2018colorless} as well as an attribution of additional parameters to the recoding models.}\label{table:hyperparams}
\end{table}

Secondly, we have to consider the hyperparameters used by recoding models, which are listed in table \ref{table:hyperparams}. For this purpose, I sample $64$ combinations of hyperparameters for surprisal recoding and recoding with predictive entropy, respectively, following the idea of \cite{bergstra2012random}, from either uniform or exponential distributions, which are listed in fig. \ref{table:hyperparameter-sample-ranges}. These ranges and corresponding distributions are based on preliminary, manual experiments. Following the creation of a set of hyperparameters, a model is trained for $4$ epochs on the training set. This point in time was chosen because it was also observed that models start to diverge during this training phase which is indicative of their final (relative) performance. Lastly, the validation perplexity is recorded and used to select the best found parameters, which are given in table \ref{table:hyperparameter-best}.

\begin{table}
  \begin{subfigure}[b]{0.6\columnwidth}
  \centering
  \setlength{\tabcolsep}{4pt}
  \def\arraystretch{1.5}
  \resizebox{.95\columnwidth}{!}{
  \begin{tabular}{@{}rll@{}}
       \toprule[1.5pt]
       Hyperparameter & Distribution & Range \\
       \toprule[1.15pt]
       Step size & Exponential & $[0.5, 75]$ \\
       \# Samples & Uniform & $[2, 50]$ \\
       Prior scale & Uniform & $[0.1, 0.4]$ \\
       Weight decay & Exponential & $[0.0001, 0.01]$ \\
       Dropout & Uniform & $[0.1, 0.6]$ \\
       MC Dropout & Uniform & $[0.1, 0.6]$ \\
       \bottomrule[1.15pt]
  \end{tabular}}
  \vspace{0.2cm}
  \caption{Sample ranges.}\label{table:hyperparameter-sample-ranges}
  \end{subfigure}\hfill%
  \begin{subfigure}[b]{0.4\columnwidth}
    \centering
    \setlength{\tabcolsep}{4pt}
    \def\arraystretch{1.5}
    \resizebox{.875\columnwidth}{!}{
    \begin{tabular}{@{}rl@{}}
         \toprule[1.5pt]
         Hyperparameter & Value \\
         \toprule[1.15pt]
         Step size & $10.19$ (Surp.) \\
          & $0.001$ (rest) \\
         \# Samples &  $15$ \\
         Prior scale & $0.29$ \\
         Weight decay & $4.82 \cdot 10^{-5}$ \\
         Dropout & $0.15$ \\
         MC Dropout & $0.42$ \\
         \bottomrule[1.15pt]
    \end{tabular}}
    \vspace{0.2cm}
    \caption{Best values found.}\label{table:hyperparameter-best}
    \end{subfigure}
    \caption[Sample ranges for hyperparameter search]{Left: Ranges from which listed hyperparameters are sampled. In case they are sampled an exponential distribution, the mean value is set to $\beta = (\text{upper} - \text{lower})/4$. Right: Best values found for recoding hyperparameters, rounded to two decimal points.}
\end{table}

\section{Additional plots}\label{appendix:additional-plots}

This section simply puts additional plots on display that were not essential to the conclusions drawn in the corresponding sections.

\begin{figure}[h]
  \centering
  \begin{tabular}{cc}
    \includegraphics[width=0.495\textwidth]{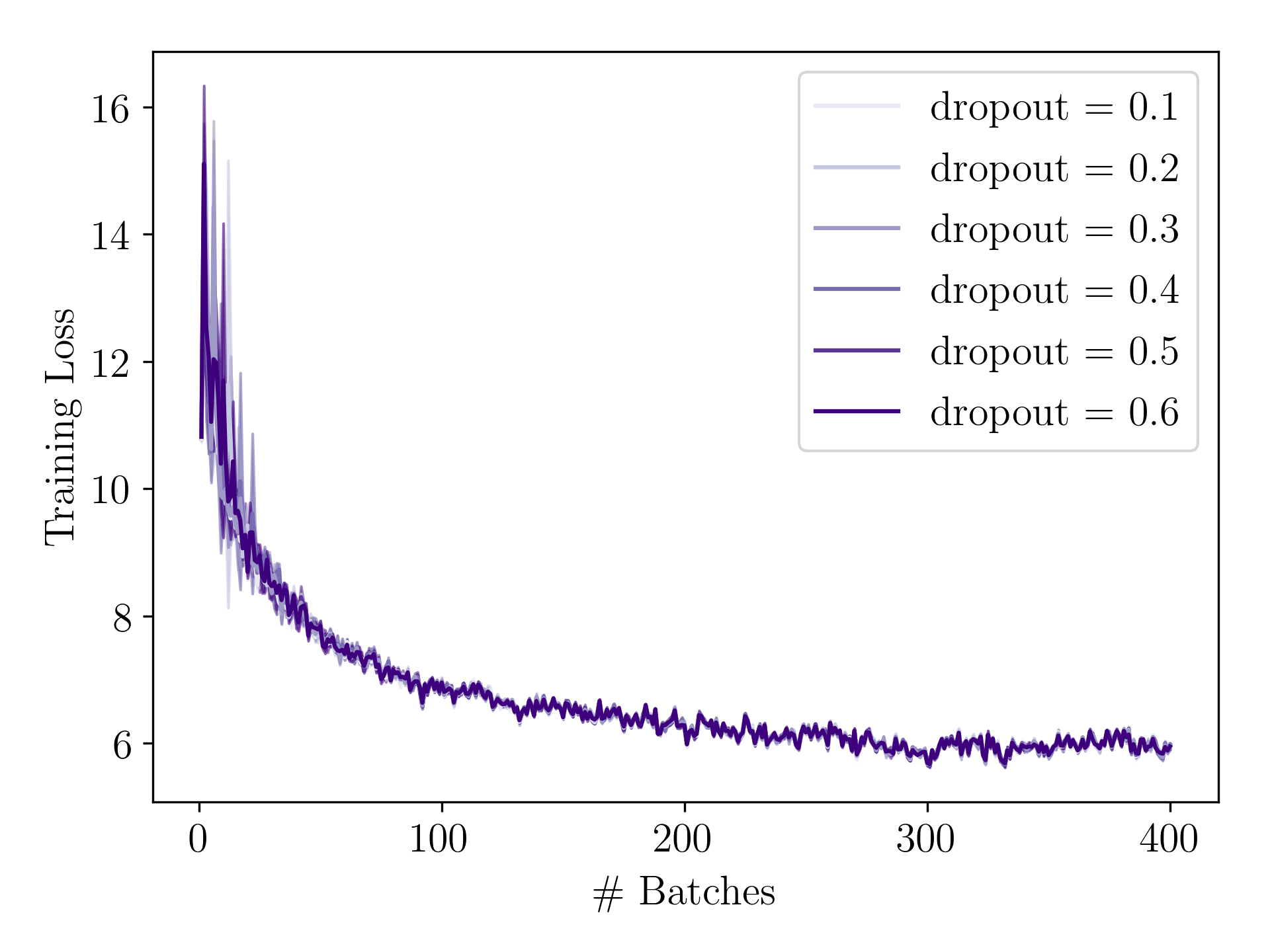} &
    \includegraphics[width=0.495\textwidth]{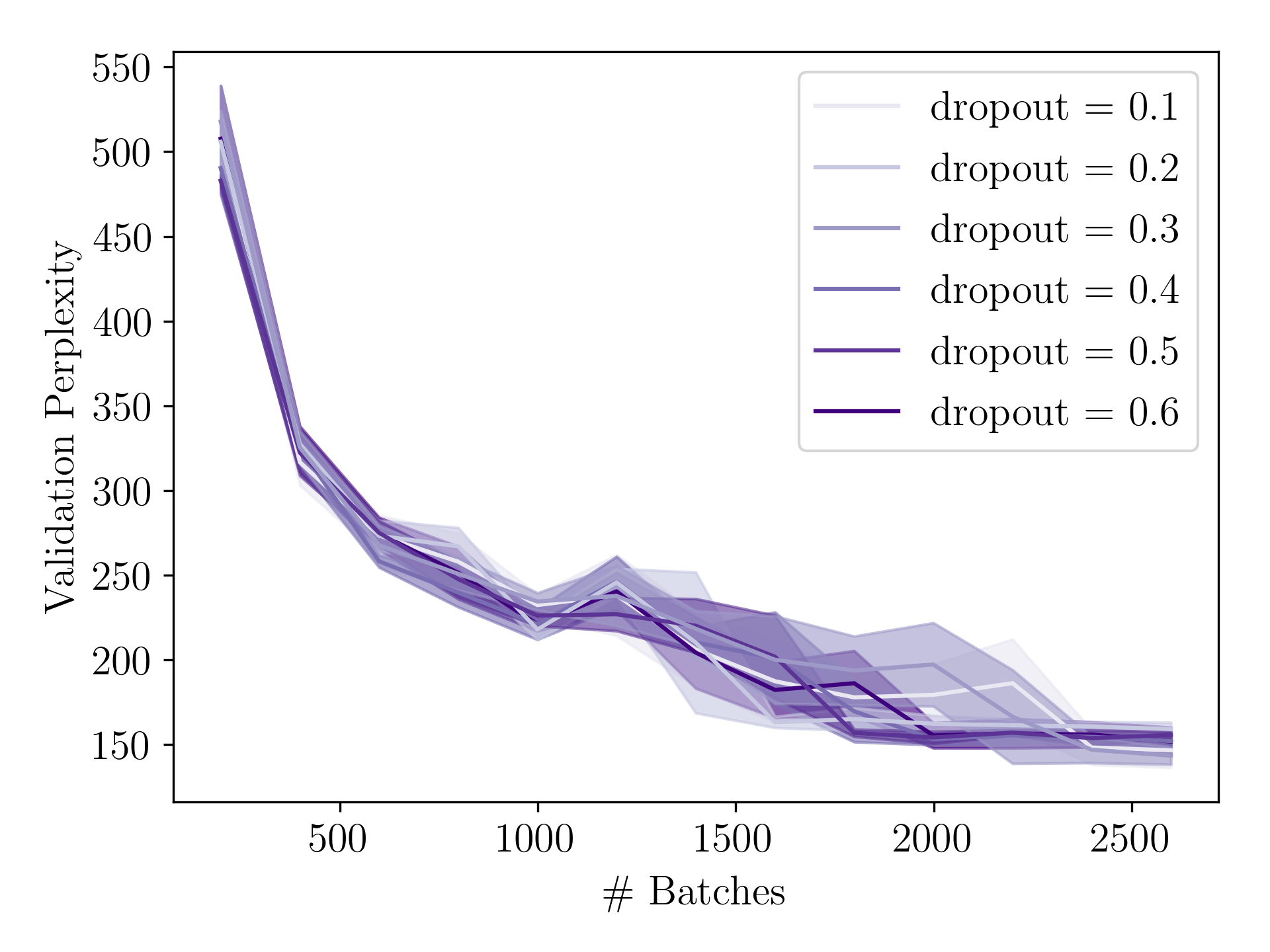} \\
  \end{tabular}
  \caption[Training loss for different dropout rates when using \gls{MCD} recoding]{Training loss (left column) and validation perplexities (right column) on \gls{PTB} using a different dropout rate to approximate predictive uncertainty using \mcdcol{MC Dropout} recoding. Curves denote the mean over $n=4$ runs, with intervals signifying the standard deviation. Best viewed in color.}\label{fig:exp-dropout-curves}
\end{figure}

\begin{figure}[h]
  \centering
  \captionsetup[subfigure]{justification=centering}
  \begin{subfigure}[b]{0.5\columnwidth}
    \centering
    \includegraphics[width=0.95\textwidth]{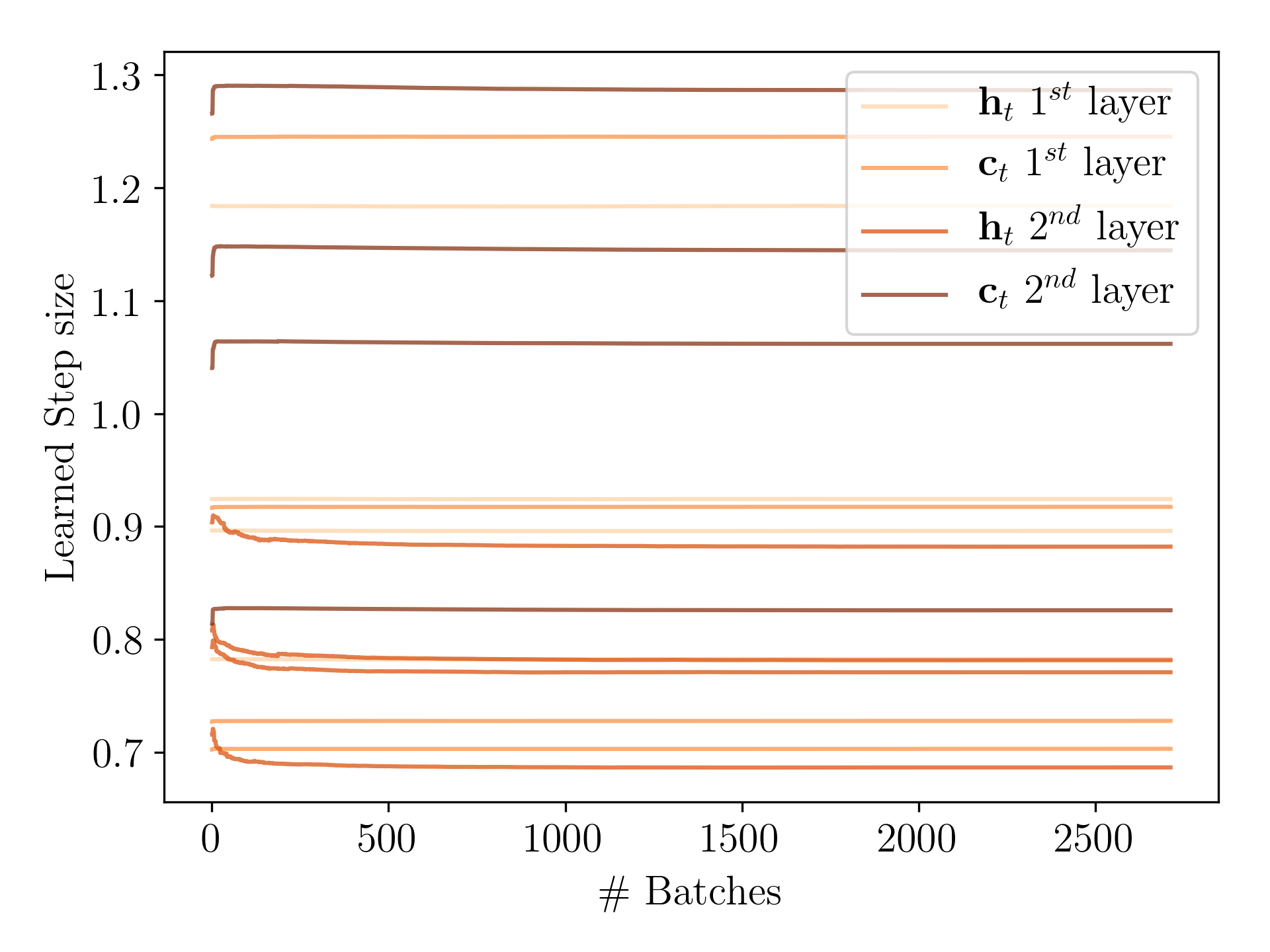}
    \caption{Learned steps for $n=4$ surp. models.}
  \end{subfigure}\hfill%
  \begin{subfigure}[b]{0.5\columnwidth}
    \centering
    \includegraphics[width=0.95\textwidth]{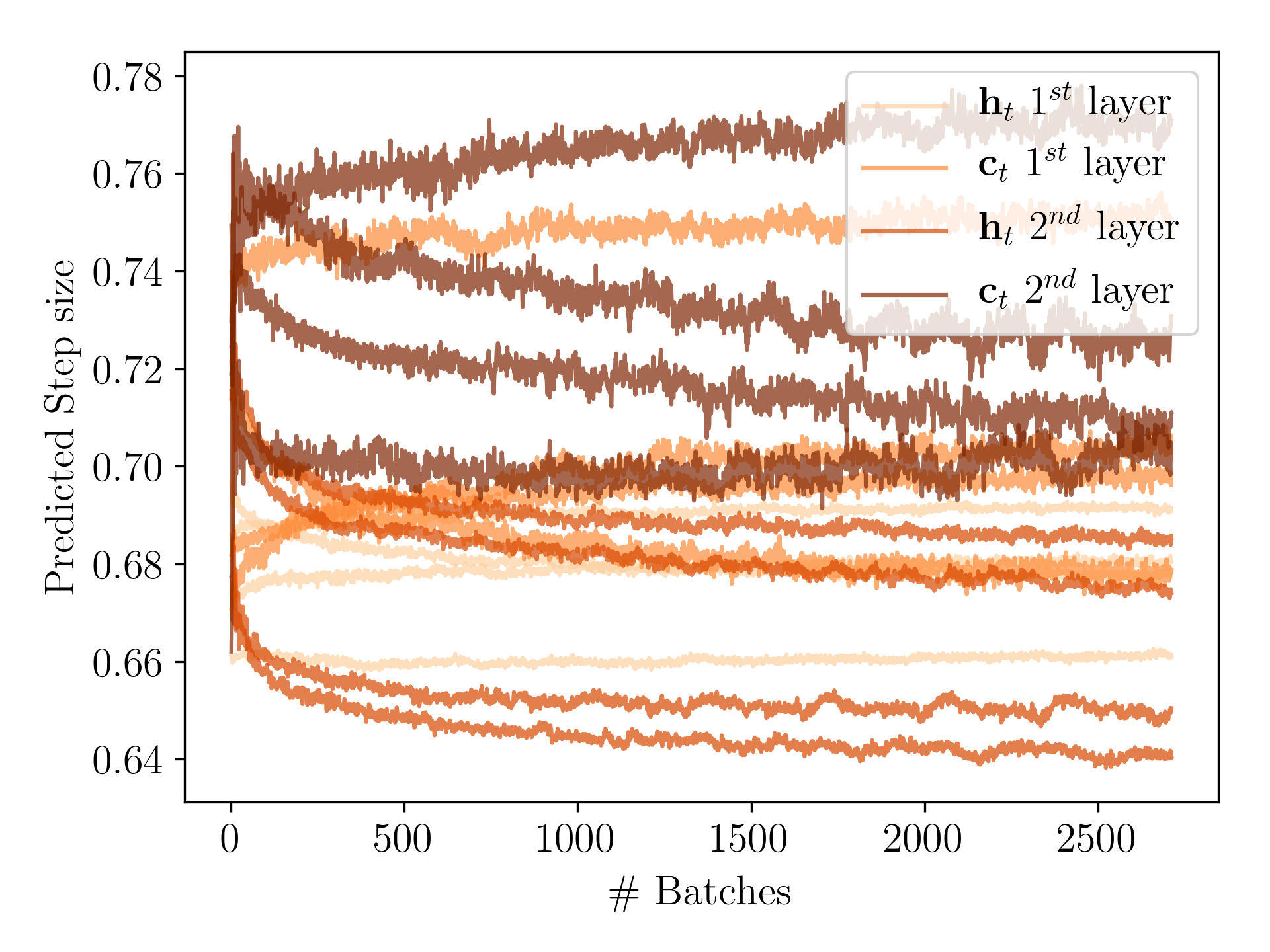}
    \caption{Predicted steps for $n=4$ surp. models.}
  \end{subfigure}\\
  \begin{subfigure}[b]{0.5\columnwidth}
    \centering
    \includegraphics[width=0.95\textwidth]{./img/learned/mcd_learned_steps.png}
    \caption{Learned steps for $n=4$ \gls{MCD} models.}
  \end{subfigure}\hfill%
  \begin{subfigure}[b]{0.5\columnwidth}
    \centering
    \includegraphics[width=0.95\textwidth]{./img/learned/ensemble_predicted_steps.png}
    \caption{Predicted steps for $n=4$ \gls{BAE} models.}
  \end{subfigure}
  \caption[Additional plots for dynamic step size models.]{Additional plots for dynamic step size models showing the development of the average learned / predicted step size per batch over the course of the training.}\label{fig:dynamic-steps-additional-results}
\end{figure}

\begin{figure}[h]
  \centering
  \begin{tabular}{cc}
    \includegraphics[width=0.495\textwidth]{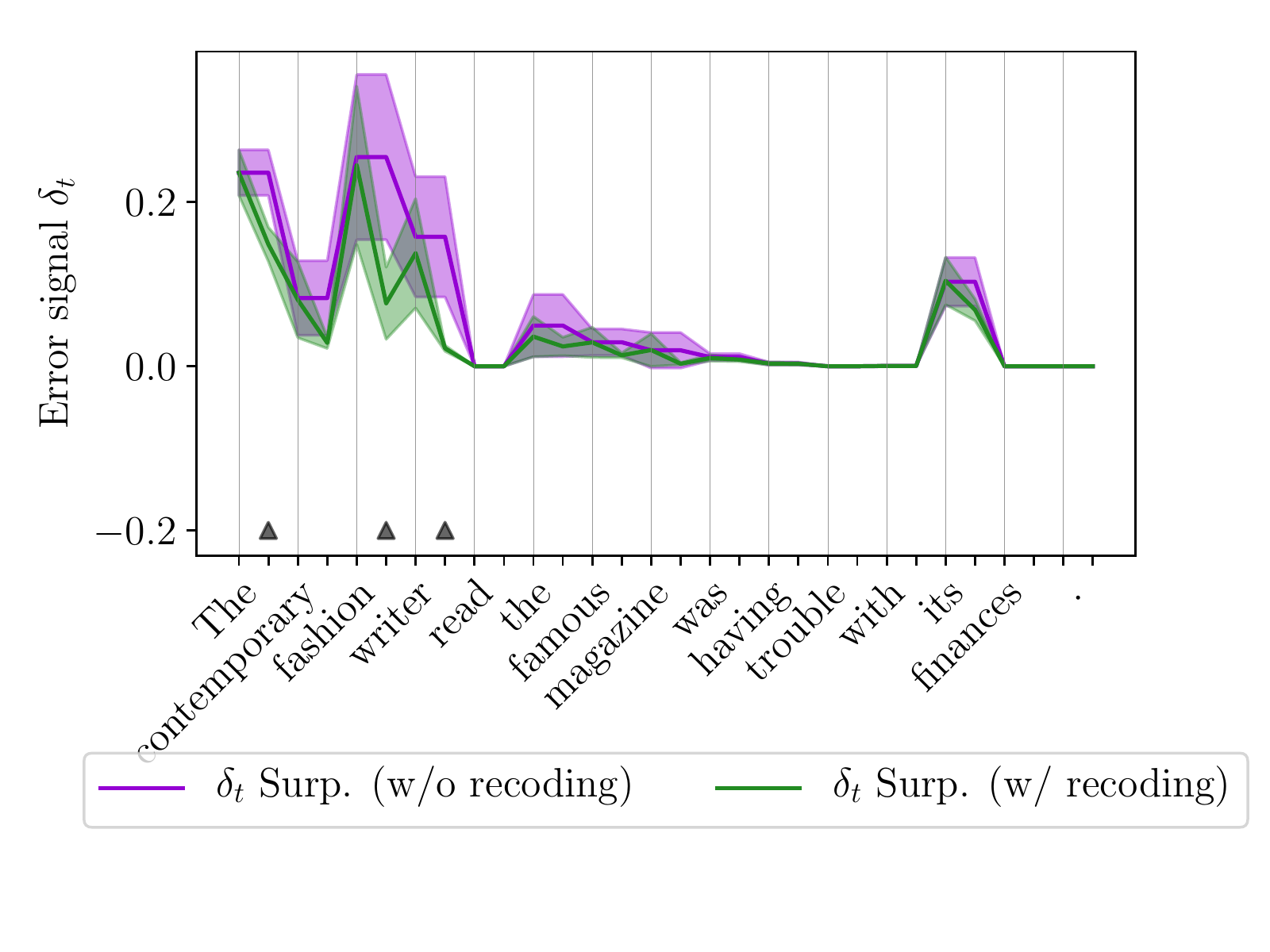} &
    \includegraphics[width=0.495\textwidth]{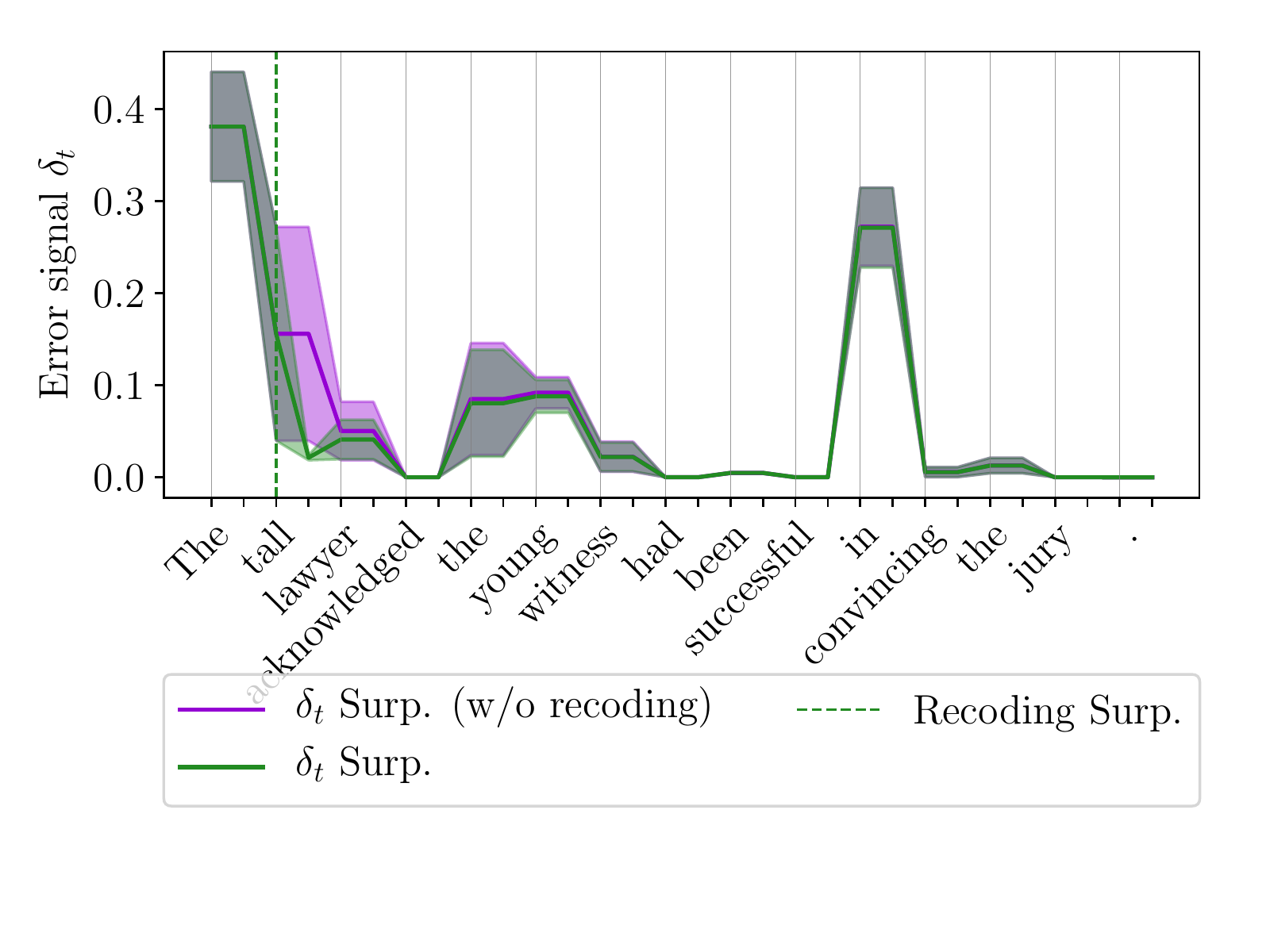} \\
    \includegraphics[width=0.495\textwidth]{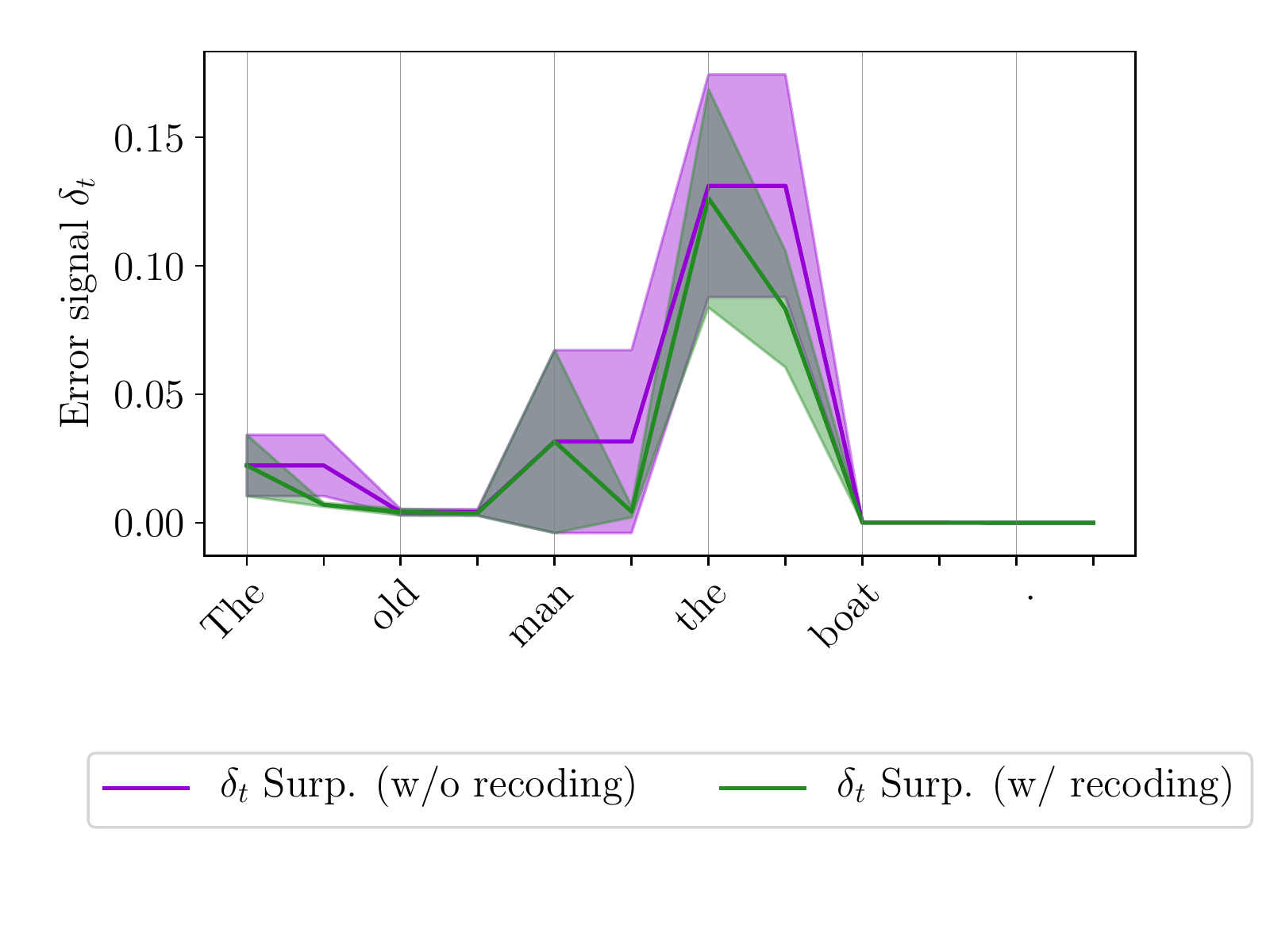} &
    \includegraphics[width=0.495\textwidth]{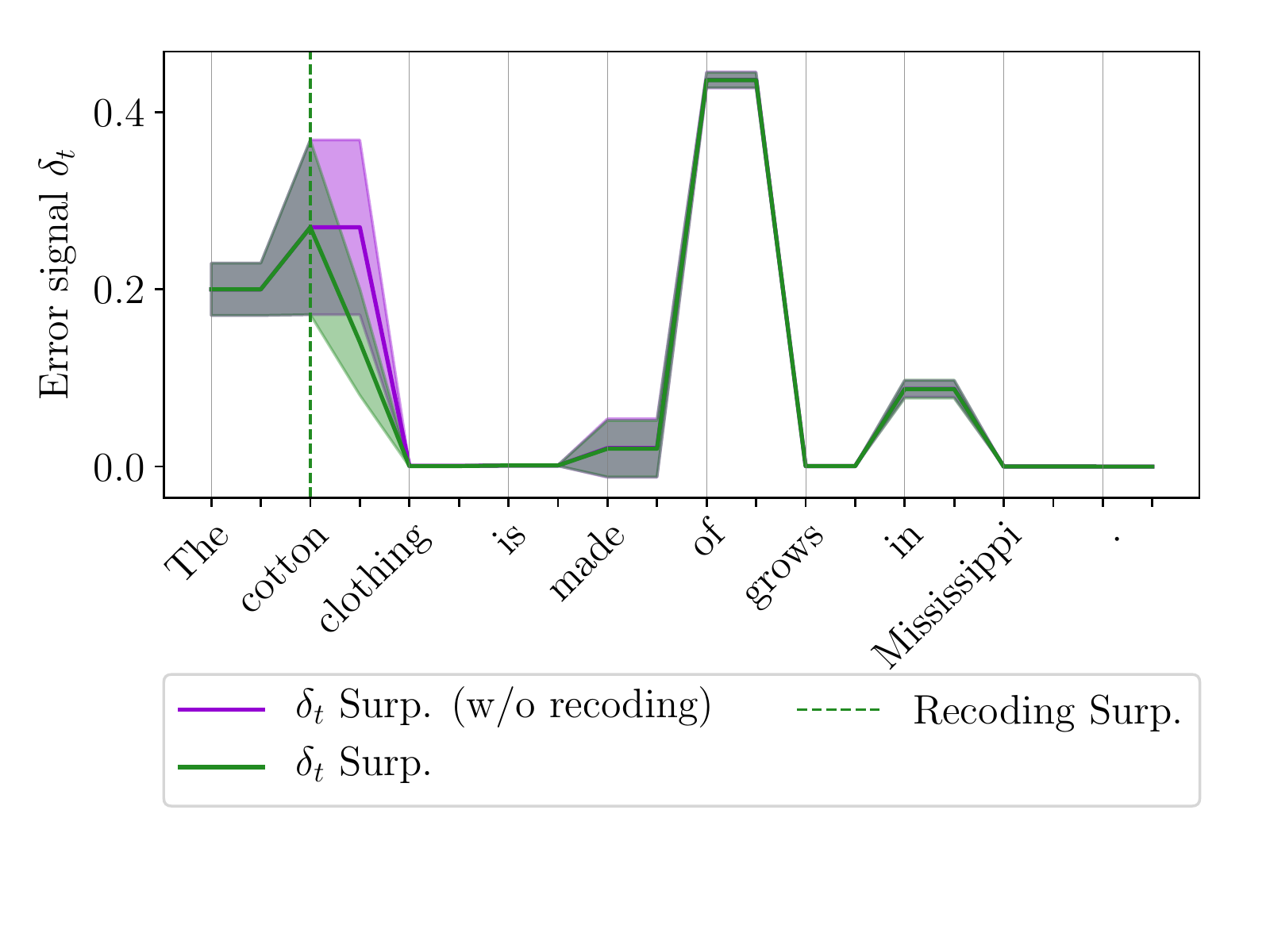} \\
  \end{tabular}
  \caption[Additional plots for error reduction through recoding on garden path sentences]{Additional plots  for error reduction through recoding on garden path sentence for the surprisal based recoding model with fixed step size. In the bottom row, recoding at all (left) or a single time step (right) has no statistically significant effect on the error signal scores. Gray markers denote statistically significant differences (two-tailed Student's t-test, $p < 0.05$).}\label{fig:qualtitative-additional-error-reduction}
\end{figure}

\end{document}